\documentclass{midl} 

\makeatletter
\renewcommand*{\subfigure}[1][]{%
  \bgroup
  \def\@subfigcap{#1}%
  \@subfigure
}

\renewcommand*{\@subfigure}[2][b]{%
  \advance\c@figure by 1\relax
  \refstepcounter{subfigure}%
  \sbox\@subfloatcapbox{%
    \ifx\@subfigcap\@empty
    \else
      \@subfigcap
    \fi
  }%
  \sbox\@subfloatcontsbox{#2}%
  \begin{tabular}[#1]{@{}c@{}}%
    \usebox\@subfloatcontsbox%
    \ifdim\wd\@subfloatcapbox>0pt
      \\ \usebox\@subfloatcapbox
    \fi
  \end{tabular}%
  \egroup
}
\makeatother

\usepackage{mwe} 
\usepackage{algorithm}
\usepackage{algorithmic}
\usepackage{multirow}
\usepackage{tablefootnote}
\usepackage{placeins}
\usepackage{amsmath}
\usepackage{amssymb}
\usepackage{xcolor}

\jmlrvolume{-- 37}
\jmlryear{2026}
\jmlrworkshop{Full Paper -- MIDL 2026}
\editors{Accepted for publication at MIDL 2026}

\title[Impact of Reconstruction on Downstream AI Fairness and Performance]{Evaluating the Impact of Medical Image Reconstruction on Downstream AI Fairness and Performance}

 \midlauthor{\Name{Matteo Wohlrapp\nametag{$^{1,2}$}} \Email{matteo.wohlrapp@cdtm.de}\\
  \Name{Niklas Bubeck\nametag{$^{2,3}$}} \Email{niklas.bubeck@tum.de}\\
  \Name{Daniel Rueckert\nametag{$^{2,3,4}$}} \Email{daniel.rueckert@tum.de}\\
  \Name{William Lotter\nametag{$^{1,5}$}} \Email{lotterb@ds.dfci.harvard.edu}\\
  \addr $^{1}$Department of Data Science, Dana-Farber Cancer Institute, Boston, MA, USA\\
  \addr $^{2}$AI in Medicine, Technical University of Munich, Munich, Germany\\
  \addr $^{3}$Munich Center for Machine Learning (MCML), Munich, Germany\\
  \addr $^{4}$Department of Computing, Imperial College London, London, UK\\
   \addr $^{5}$Harvard Medical School, Boston, MA, USA
  }

\begin{document}

\maketitle

\begin{abstract}
AI-based image reconstruction models are increasingly deployed in clinical workflows to improve image quality from noisy data, such as low-dose X-rays or accelerated MRI scans. However, these models are typically evaluated using pixel-level metrics like PSNR, leaving their impact on downstream diagnostic performance and fairness unclear. We introduce a scalable evaluation framework that applies reconstruction and diagnostic AI models in tandem, which we apply to two tasks (classification, segmentation), three reconstruction approaches (U-Net, GAN, diffusion), and two data types (X-ray, MRI) to assess the potential downstream implications of reconstruction. We find that conventional reconstruction metrics poorly track task performance, where diagnostic accuracy remains largely stable even as reconstruction PSNR declines with increasing image noise. Fairness metrics exhibit greater variability, with reconstruction sometimes amplifying demographic biases, particularly regarding patient sex. However, the overall magnitude of this additional bias is modest compared to the inherent biases already present in diagnostic models. To explore potential bias mitigation, we adapt two strategies from classification literature to the reconstruction setting, but observe limited efficacy. Overall, our findings emphasize the importance of holistic performance and fairness assessments throughout the entire medical imaging workflow, especially as generative reconstruction models are increasingly deployed. 

\end{abstract}

\begin{keywords}
Fairness, Image Reconstruction, GANs, Diffusion Models
\end{keywords}

\begin{figure*}[h]
\centering
\includegraphics[width=\linewidth]{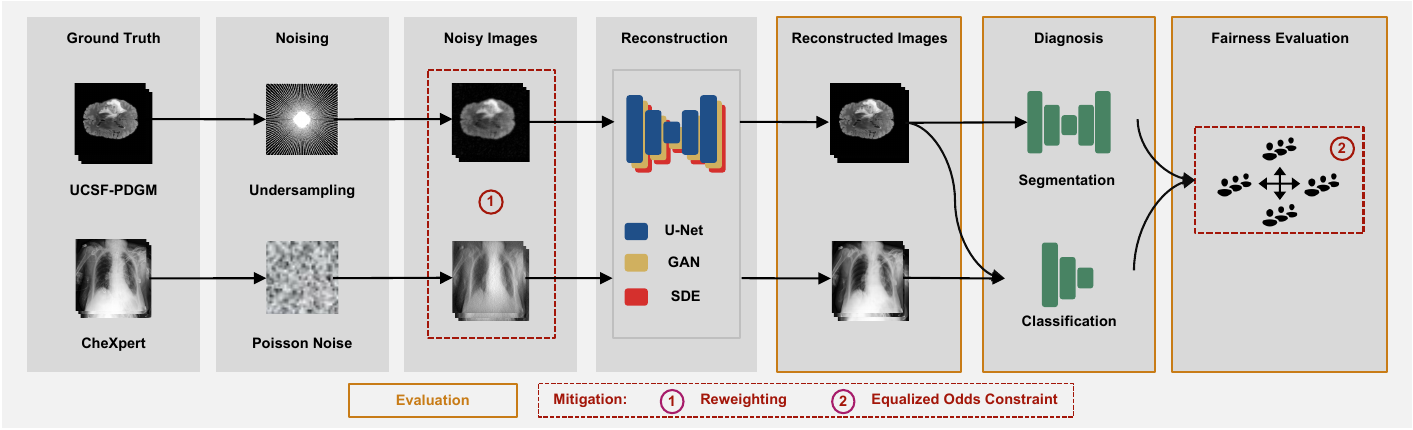}
\vspace{-20pt}
\caption{Combined pipeline for downstream bias evaluation and mitigation in medical image reconstruction. MRI and X-ray images undergo realistic simulated degradation and are subsequently reconstructed with three approaches before serving as input to downstream prediction models. Reconstruction quality, downstream performance, and fairness are evaluated. Subsequently, two bias mitigation strategies are applied exclusively during reconstruction fine-tuning.}
\label{fig:method}
\end{figure*}

\section{Introduction}
AI-based image reconstruction is an increasingly integral component of clinical workflows. These approaches are designed to enhance the quality of noisy medical images such as low-dose X-rays or faster-sampled MRIs, ultimately generating new medical images by imputing patterns learned from the training datasets \cite{AHISHAKIYE2021118}. Notably, there are now over 80 FDA-cleared devices based on this approach \cite{Singh2025.03.13.25323924}, whose generated images are ultimately interpreted by clinicians. 

Traditionally, reconstruction model performance has been evaluated using pixel-level image metrics such as PSNR. However, these metrics provide an incomplete picture, as they do not reflect the impact of reconstructed images on subsequent clinical tasks. This gap raises a key unresolved question: \textit{How does AI-based reconstruction influence downstream clinical performance and, in particular, fairness?} The latter is especially important to assess given the risk of generative models in encoding biases \cite{saumure2025humor, ruggeri-nozza-2023-multi, luccioni2023stablebiasanalyzingsocietal, mehrabi2022surveybiasfairnessmachine}. While some smaller-scale studies have involved clinician review of AI-reconstructed images \citep{Feuerriegel2023-rt, Lee2024-ic}, this approach is not scalable, especially when investigating nuanced performance differences across subgroups. 

In this work, we assess the downstream implications of AI-based reconstruction through an evaluation framework that leverages reconstruction and classification/segmentation AI models applied in tandem. The framework provides a scalable approach to understand how reconstruction errors propagate, while also simulating a realistic clinical scenario as both reconstruction and diagnostic models are increasingly deployed in medical workflows. We apply this framework across three reconstruction approaches (U-Net, GAN, diffusion), two imaging domains (MRI, X-ray), and two tasks (classification and segmentation). We additionally propose and evaluate bias mitigation techniques tailored to reconstruction models. Our findings highlight differences in trends between image metrics and diagnostic accuracy, and the potential of reconstruction models to shift demographic biases.
\section{Related Work}

\paragraph{Reconstruction Models in Medical Imaging:}
Medical image reconstruction is a popular AI application due to its promise in increasing image quality while facilitating lower radiation doses and faster scanning times \cite{AHISHAKIYE2021118, Diab2025-fm}. Given pairs of noisy (i.e., undersampled/lower dose) and original images, these models are trained to reconstruct the original from the noisy image. Recently, unsupervised methods have also been developed that do not require paired original and noisy images \cite{Chen2023-vn, Sultan2025-co, CHEN2026102015}. Variations of the U-Net \cite{Unet} are commonly used as the neural network architecture for reconstruction models. In addition to standard losses like mean-squared error (MSE), GAN and diffusion-based approaches are common in the field \cite{bousse2024review, Heckel2024}.

\paragraph{Fairness Analysis in Medical Imaging:}
Research on bias in AI-driven healthcare spans various medical domains, with medical imaging receiving considerable attention. In classification tasks, biases are typically revealed by comparing performance across subgroups. Studies cover various imaging modalities, including brain MRI \cite{Stanley2022FairnessrelatedPA, ioannou2022studydemographicbiascnnbased}, chest X-rays \cite{Kalantari2021UnderdiagnosisCheX, Glocker2023AlgorithmicEO, Yang2024-dm, Lotter2024-sk}, dermatology images \cite{CHIU2024103188, groh2021evaluating}, and retinal images \cite{Burlina2021Retinal}. They address sensitive attributes such as sex \cite{Stanley2022FairnessrelatedPA}, age \cite{Kalantari2021UnderdiagnosisCheX}, race \cite{Kalantari2021UnderdiagnosisCheX}, and skin tone \cite{Kinyanyui2020Dermatology}, evaluating disparities using performance metrics such as Area Under the Curve (AUC) \cite{Kalantari2021UnderdiagnosisCheX}, or more dedicated fairness criteria \cite{jamanetworkopen.2023.42203}. In segmentation, studies have assessed segmentation performance under varying demographic distributions, such as by race and sex representation in training datasets \cite{ioannou2022studydemographicbiascnnbased,lee2022systematicstudyracesex,puyol2022fairness}.

\paragraph{Fairness Analysis of Reconstruction Models:}
Reconstruction model performance is typically measured using image quality metrics such as Peak Signal-to-Noise Ratio (PSNR) and Structural Similarity Index Measure (SSIM). Recent studies assessing subgroup biases primarily rely on these metrics, examining how image quality varies across demographic subgroups. For instance, \citet{du2023unveilingfairnessbiasesdeep} investigated fairness in deep learning-based brain MRI reconstruction, highlighting disparities in image reconstruction quality across different demographic groups using PSNR and SSIM. Similarly, \citet{Sheg24reconbias} explored fairness challenges and potential solutions in ultrasound computed tomography, identifying significant disparities in reconstruction performance linked to subgroup attributes. With limited available literature, bias evaluation in reconstruction models is an emerging area of research for which there is a need to study the implications of image reconstruction on downstream tasks.

\paragraph{Bias Mitigation:}
In classification, substantial efforts have focused on developing bias mitigation strategies. Data-centric approaches directly modify training datasets, employing methods such as data redistribution \cite{Oguguo23}, differentiable resampling techniques \cite{repair}, harmonization of datasets \cite{bissoto2019deconstructingbiasskinlesion}, and synthetic generation of diverse samples \cite{WANG2024105047}. Additionally, methods like Just Train Twice (JTT) target misclassified instances to implicitly mitigate subgroup biases without explicit annotations \cite{JTT}.

Representation-level strategies aim to learn unbiased feature representations through explicit disentanglement. Techniques include variational autoencoders \cite{Creager2019FlexiblyFR}, orthogonal disentanglement methods enforcing independence between sensitive attributes and task-specific features \cite{Sarhan,WenlongOrtho,CHIU2024103188,FairDisCo}, and group-adaptive architectures employing demographic-specific attention mechanisms \cite{Gong}.

Optimization-level methods integrate fairness constraints into model training via adversarial learning, fairness-specific loss functions, or specialized training regimens. Adversarial methods discourage encoding protected attributes \cite{Zhang,Adeli2019RepresentationLW,KimKimKimKimKim,Wang}, distributionally robust optimization (Group DRO) targets worst-case subgroup performance \cite{DRO}, and fairness-specific constraints can be incorporated directly into training \cite{marcinkevičs2022debiasingdeepchestxray}. Post-processing methods adjust model outputs after training, employing techniques such as calibration and pruning \cite{wu2022fairpruneachievingfairnesspruning}.

While prior studies have focused mainly on bias mitigation in classification tasks, there remains a critical need to assess analogous strategies for image reconstruction.
\section{Methods}

Our framework, visualized in Figure~\ref{fig:method}, encompasses image denoising, downstream task evaluation, fairness assessment, and bias mitigation for medical image reconstruction. The framework uses classification and segmentation models to estimate the effect of reconstruction on downstream task performance and fairness. Additionally, mitigation strategies are applied exclusively at the reconstruction stage to determine their ability to reduce downstream biases without retraining diagnostic models. 

\subsection{Datasets}
We apply our framework to public datasets from two distinct imaging domains:

\paragraph{MRI:} UCSF-PDGM includes 501 pre-operative glioma MRI exams from patients with diffuse glioma, along with tumor masks and labels for subtype and grade~\cite{Calabrese_2022}. We use the T2-weighted FLAIR volumes for all analyses. 

\paragraph{X-Ray:} CheXpert comprises 224,316 radiographs from 65,240 patients annotated for 14 thoracic findings~\cite{CheXpert}, of which we use 12 (excluding ``Support Devices'' and ``No Findings'' to focus on disease pathologies).  

We use a 70/10/20 train/validation/test split stratified by patient for both datasets. For CheXpert, the training set is further divided into non-overlapping sets for reconstruction and classification model training, with percentages of 70/30, respectively. For UCSF-PDGM, the same training data is for both tasks given smaller sample size. 
Group-wise fairness is assessed for age (dichotomized at the dataset median), sex, and self-reported race (unavailable for UCSF-PDGM). Detailed attribute distributions are reported in Tables~\ref{tab:chex_dataset} and~\ref{tab:ucsf_dataset} in the Appendix.

\subsection{Noising Process}
We simulate realistic acquisition degradations as follows:
\paragraph{MRI:} $k$-space data is masked with radial undersampling patterns~\cite{FengRadial} at acceleration factors 4, 8, and 16, where higher acceleration means greater undersampling (see Appendix \ref{app:mri}). 
\paragraph{X-Ray:} Standard-dose images are Radon-projected to sinogram space, bow-tie filtered, and corrupted with Poisson noise parameterized by photon count ($100{,}000$, $10{,}000$, $3{,}000$), with lower photon count yielding more noise ~\cite{Gibson2023APM}. 

These ranges approximate realistic acquisition conditions, with examples in the Appendix (Figures \ref{fig:noise-chex}, \ref{fig:noise-ucsf}, \ref{fig:chex-example}, and \ref{fig:ucsf-example}).

\subsection{Models}
We employ three reconstruction models alongside task-specific diagnostic models. Additional information on the compute infrastructure and model hyperparameters can be found in the Appendix.

\paragraph{Reconstruction:}
To cover deterministic, adversarial, and diffusion regimes, we train from scratch a standard U-Net~\cite{Unet} with MSE loss, a Pix2Pix GAN~\cite{pix2pix2017}, and a Stochastic Differential Equations (SDE)-based diffusion model~\cite{sde} for each dataset. We note that the GAN and diffusion models also use a U-Net as the model architecture, but are based on a different training paradigm.

\paragraph{Diagnostic:}
For classification on UCSF-PDGM, an ImageNet-initialized ResNet50~\cite{resnet} was trained separately to predict WHO grade and tumor type. The model is trained at the slice-level, and at testing, volume-level predictions are performed individually on each slice and then aggregated using the median. For CheXpert classification, a single ImageNet-initialized DenseNet model~\cite{densenet} was trained to jointly predict the 12 findings following~\citet{Cohen2021TorchXRayVisionAL}. For segmentation on UCSF-PDGM, we use an ImageNet-initialized U-Net. Segmentation is not evaluated on CheXpert due to the absence of masks. All downstream models are trained on the original, non-degraded images.

\subsection{Performance and Fairness Evaluation}
Reconstruction quality is measured by PSNR. Downstream performance uses AUROC for classification and Dice for segmentation. For classification fairness, we report the worst-case Equalized Odds (EODD) \cite{DBLP:journals/corr/HardtPS16} difference between groups:
\begin{align*}
max_{i,j}|P&(\hat{Y} = 1 \mid Y = y, A = a_i) \\&- P(\hat{Y} = 1 \mid Y = y, A = a_j)|,  \quad \forall y \in \{0,1\}, \\& \quad 
\forall \quad \text{attribute A} \in \mathcal{A}, \quad \text{subgroups} \quad a_i.
\end{align*} To compute this metric, model predictions are binarized using a balanced threshold selected to achieve approximately equal sensitivity and specificity in the validation split. Equality of Opportunity (EOP) results are also reported in the Appendix (Figures \ref{fig:bias_chexpert_eop} and \ref{fig:bias_ucsf_class_eop}).

For segmentation fairness, we adapt the Skewed-Error Ratio (SER)~\cite{SiddiquiFairSeg} to Dice:
\[
SER_A=\frac{\max_{i}(1-\text{Dice}_{a_i})}{\min_{j}(1-\text{Dice}_{a_j})}, \quad a_i \in A, \quad \text{A} \in \mathcal{A}
\]
Results using an unnormalized Dice difference are also provided in the Appendix (Figure \ref{fig:bias_ucsf_class_eop}).

Statistical comparisons of subgroup fairness differences were performed using bootstrapped estimates with 1,000 iterations. Bootstrap-derived p-values were used to determine statistical significance with a two-sided \textit{p} $<$ 0.05.

\subsection{Bias Mitigation}
We adapt two bias mitigation strategies that were originally proposed for classification models. Each approach involves fine-tuning only the reconstruction models after the original training described above. The differentiable equalized-odds approach also relies on using the reconstruction and classification models applied in tandem, but the classification network is frozen to exclusively assess the potential for bias mitigation at the reconstruction stage. 

\paragraph{Sample Reweighting:}
A weighted sampler draws each example with inverse joint subgroup frequency during fine-tuning, ensuring that each subgroup (and combination thereof across attributes) is represented with the same frequency. The reconstruction model is fine-tuned using the corresponding original reconstruction loss. 

\paragraph{Differentiable Equalized-Odds:} For reconstruction output \(\hat x=f(x)\) and classifier output \(\hat y=g(\hat x)\) we minimize:  
$
\mathcal{L}_{\mathrm{EODD}}
=\ell_{\mathrm{rec}}(\hat{x})
+\lambda_{\mathrm{fair}}\,
\mathrm{EMA}\!\bigl(\ell_{\mathrm{BCE}}(\hat{y})+\mathrm{EODD}^{2}\bigr),
$
where $\ell_{\mathrm{rec}}$ represents the original reconstruction loss for the model, $\ell_{\mathrm{BCE}}$ represents the binary cross-entropy loss for the frozen classifier, $\text{EMA}$ represents an exponential moving average, and EODD represents a differentiable Equalized Odds constraint inspired by~\citet{marcinkevičs2022debiasingdeepchestxray}. Specifically, we use the maximum EODD difference of any subgroup as defined above and compute it via soft predictions:
$
\tilde y = \sigma\!\bigl((\hat{y})-\tau)/T\bigr),
$
where the threshold \(\tau\) and temperature \(T\) are set at 0.5 and 0.3, respectively. One loss is computed across all sensitive attributes (i.e., the max EODD over age, sex, and race). In the Appendix, we show that minimizing $\mathrm{EODD}^2$ between subgroups corresponds to minimizing their covariance.

\paragraph{Code:} Available at \url{https://github.com/lotterlab/reconstruction_evaluation}

\section{Results}
We first evaluate the impact of reconstruction on downstream task performance before analyzing fairness and the effectiveness of mitigation techniques.

\begin{figure}[ht]
    \centering

    \includegraphics[width=0.95\linewidth]{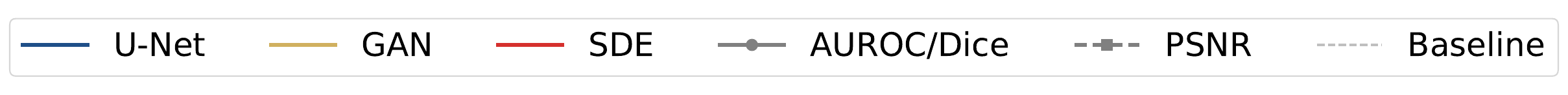}
    \vspace{1em}

    \begin{tabular}{c@{\hspace{0.5cm}}c}
        \subfigure[]{\includegraphics[width=0.49\linewidth]{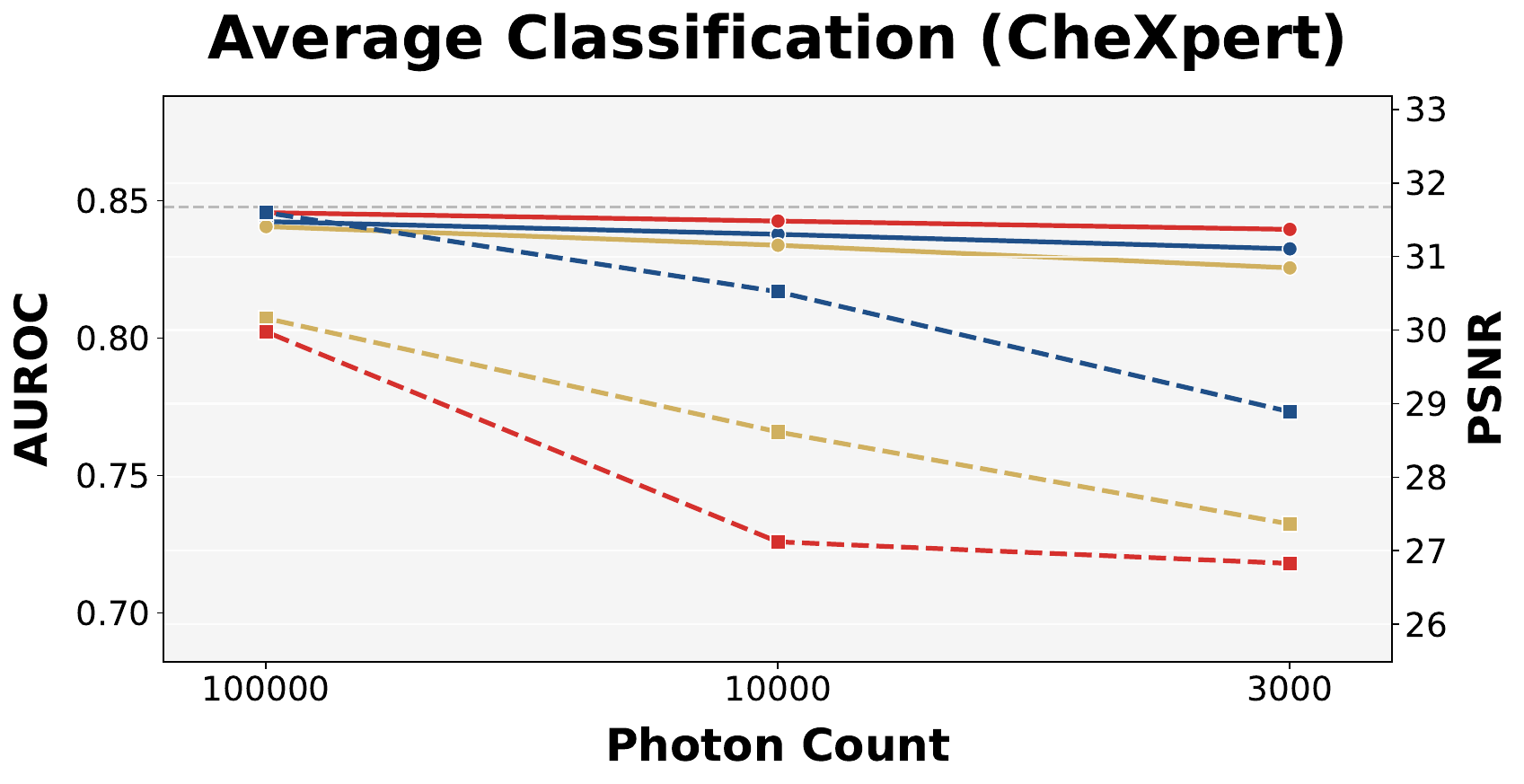}} &
        \subfigure[]{\includegraphics[width=0.49\linewidth]{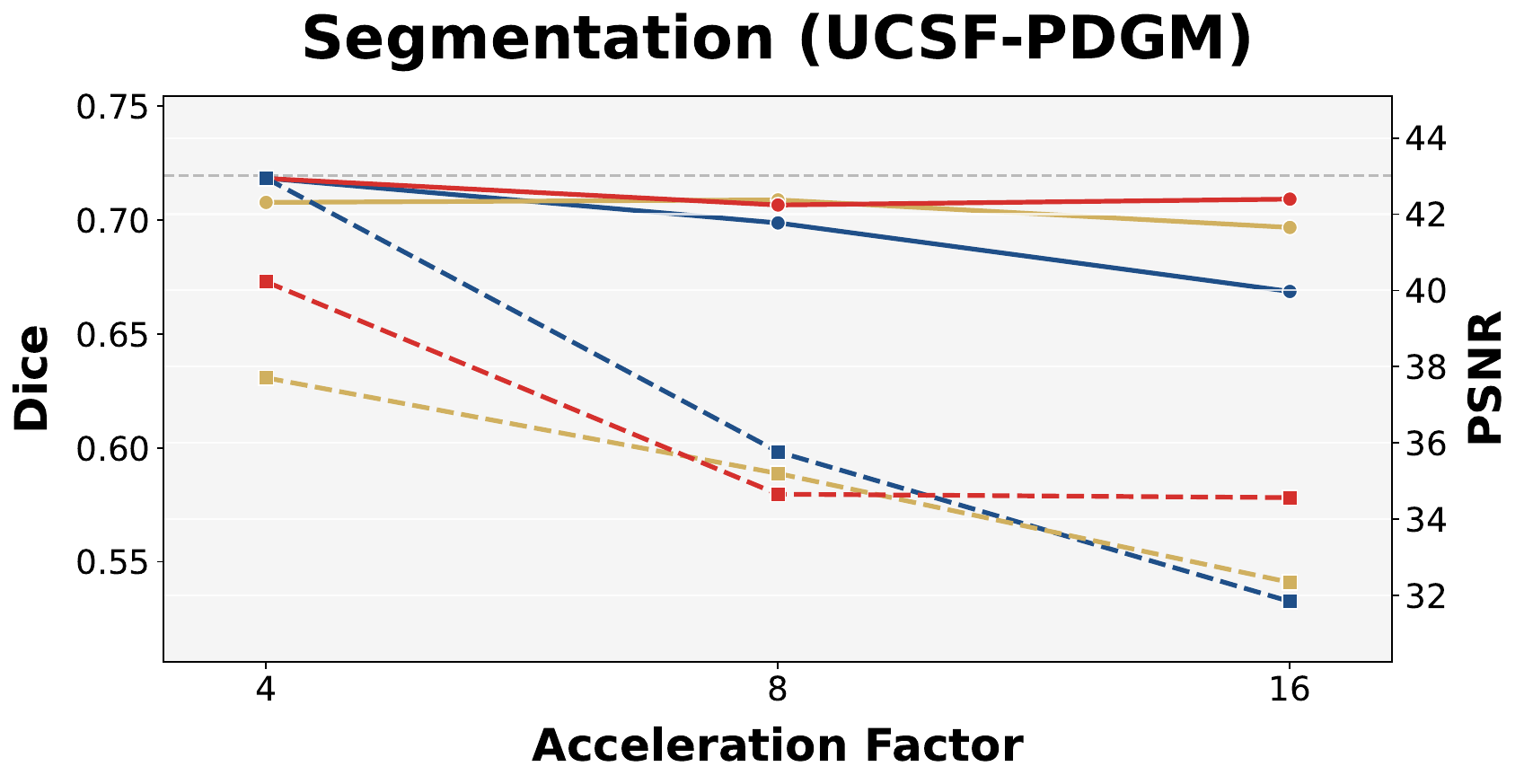}}
    \end{tabular}
    \vspace{-10pt}
    \caption{Downstream performance and PSNR at varying noise levels. Axes for PSNR and task performance are scaled to comparable percentage ranges. Although PSNR declines as noise increases, task performance remains stable. Baseline indicates performance on original images.}
    \label{fig:performance}
\end{figure}

\subsection{Impact of Reconstruction on Task Performance}

Figure~\ref{fig:performance} summarizes downstream performance as a function of reconstruction noise. We report segmentation Dice for UCSF-PDGM and the mean AUROC across the 12 CheXpert pathologies. For clarity, the y-axes for PSNR and the task metrics are normalized to the same percentage range. Across all experiments, diagnostic performance remains largely unchanged, even though PSNR decreases substantially with increasing noise. Specifically, the Dice score for UCSF-PDGM segmentation varies by no more than~$\sim\!3\,\%$ across noise conditions, and the mean CheXpert AUROC fluctuates by only~$1\,\%$. In contrast, PSNR decreases by over $10$ dB ($26\,\%$) for UCSF-PDGM and by~$\sim\!3$ dB ($9\,\%$) for CheXpert. Analogous results for UCSF-PDGM classification are presented in the Appendix (Figure~\ref{fig:performance_ucsf}), where the same pattern—substantial PSNR loss but minimal impact on task performance—holds for all three reconstruction models. Using SSIM instead of PSNR as the reconstruction metric also shows similar trends (Appendix Figure \ref{fig:performance_ssim1}).

A closer look at CheXpert reveals a mild dependence on baseline task difficulty: pathologies with lower initial AUROC show slightly larger declines. For example, consolidation remains stable with U-Net reconstruction AUROC at~0.91, whereas lung lesion drops from~0.79 to~0.77 as noise increases (see Appendix Table~\ref{tab:chex_perf} for more details).

\subsection{Impact of Reconstruction on Fairness}
Although aggregate task performance is largely unaffected, reconstruction could still alter relative performance across demographic subgroups. To test this possibility, we evaluated fairness on the downstream models using acceleration factor 8 for UCSF-PDGM and a photon count of 10,000 for CheXpert, representing the middle noise levels.

 \begin{figure}[t!]
 \centering
 \includegraphics[width=0.7\linewidth]{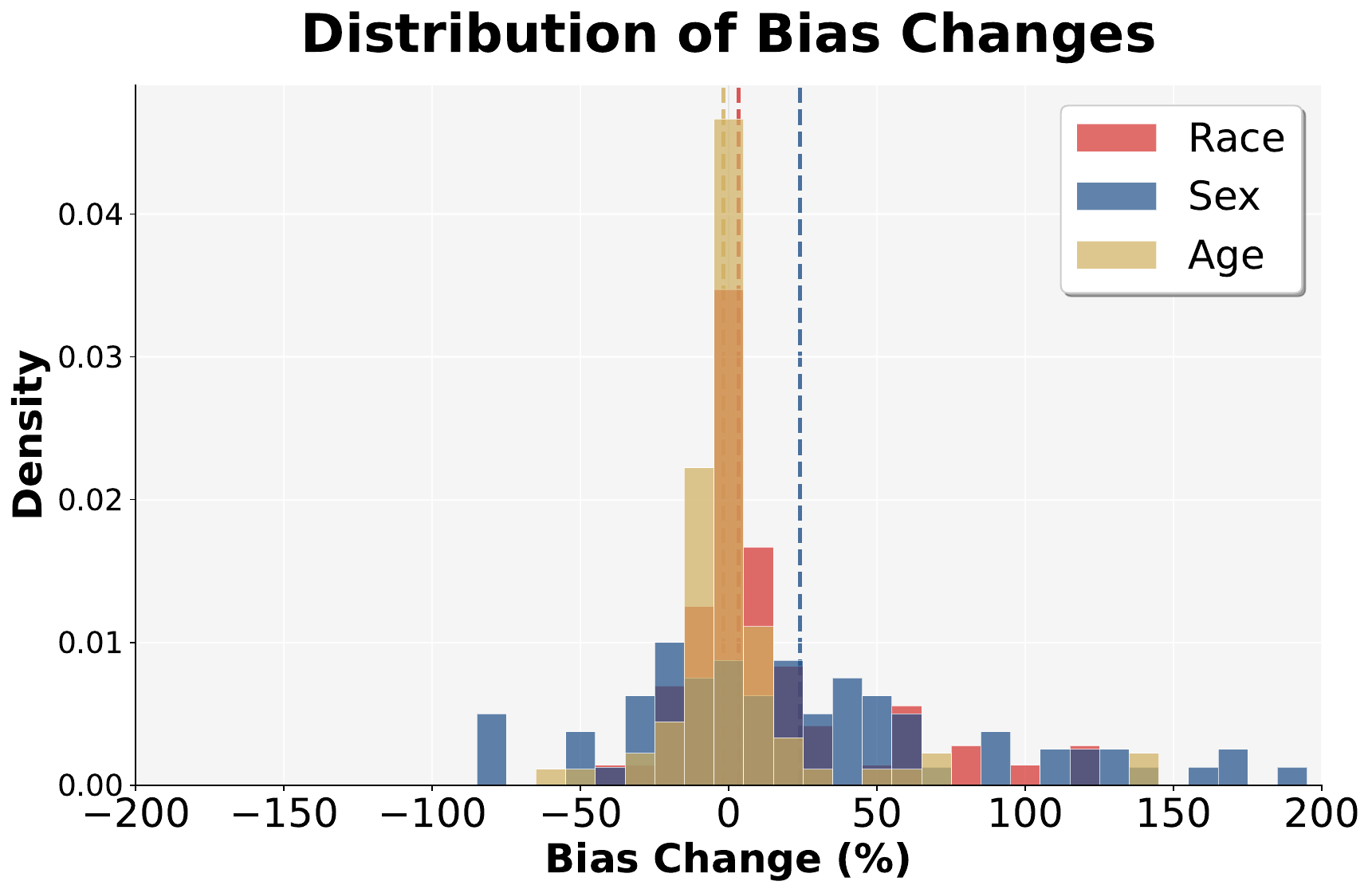}
 \vspace{-10pt}
 \caption{Distribution of bias changes (percent change compared to original images) across all reconstruction models, datasets, and tasks, stratified by sensitive attribute. The vertical lines mark the medians. Most shifts cluster near zero, but sex shows a broader positive tail. Separate plots by sensitive attribute are contained in Fig. \ref{fig:alt_bias_histogram}.}
    \label{fig:histogram}
\end{figure}

\begin{figure*}[!t]
\centering

\includegraphics[width=\textwidth]{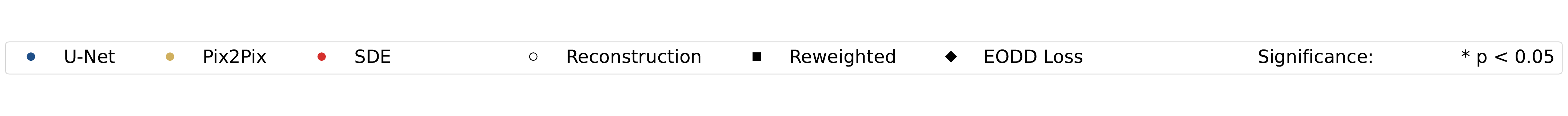}

\begin{tabular}{c@{\hspace{0.2cm}}c}

\subfigure[]{%
\begin{minipage}[t]{0.49\textwidth}
\centering
\includegraphics[width=\linewidth]{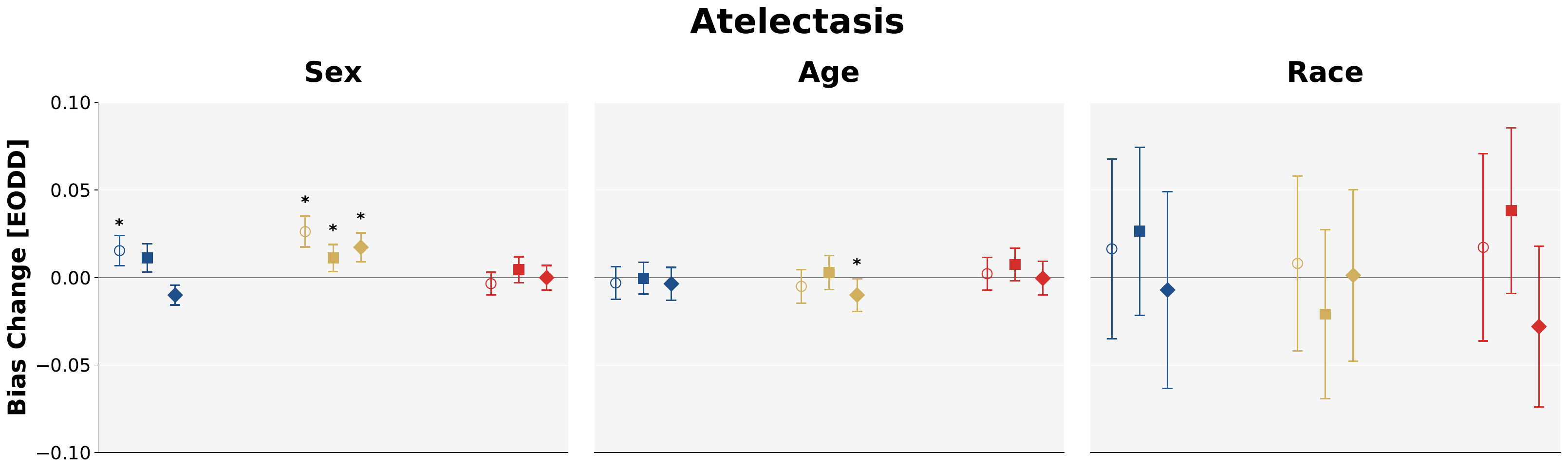}
\end{minipage}
}
&
\subfigure[]{%
\begin{minipage}[t]{0.49\textwidth}
\centering
\includegraphics[width=\linewidth]{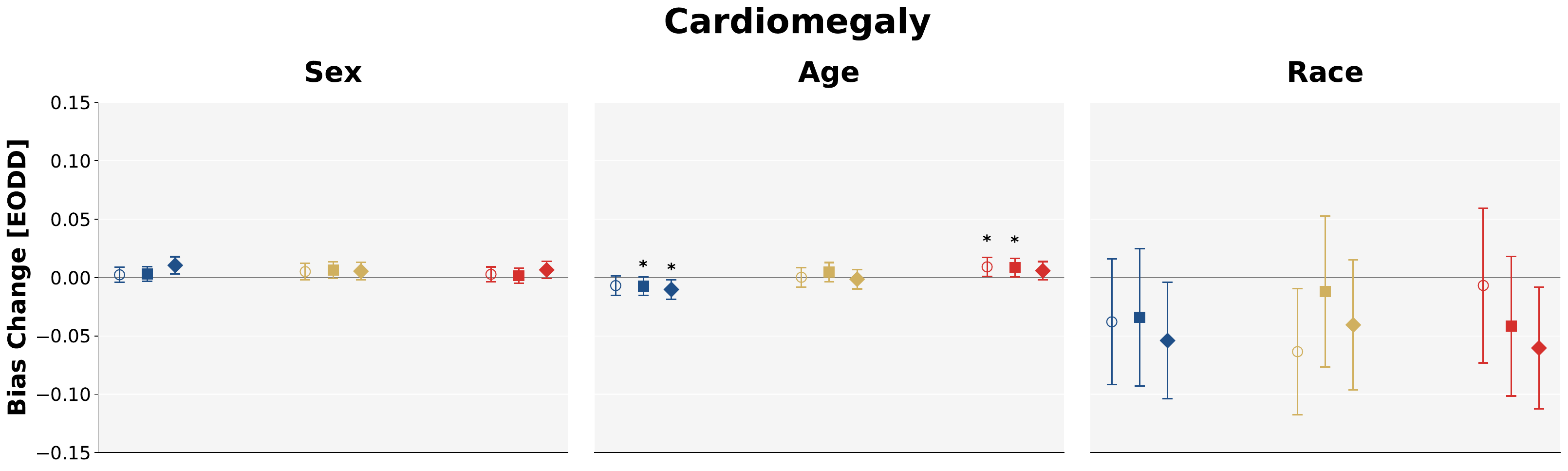}
\end{minipage}
}
\\[0.3cm]

\subfigure[]{%
\begin{minipage}[t]{0.49\textwidth}
\centering
\includegraphics[width=\linewidth]{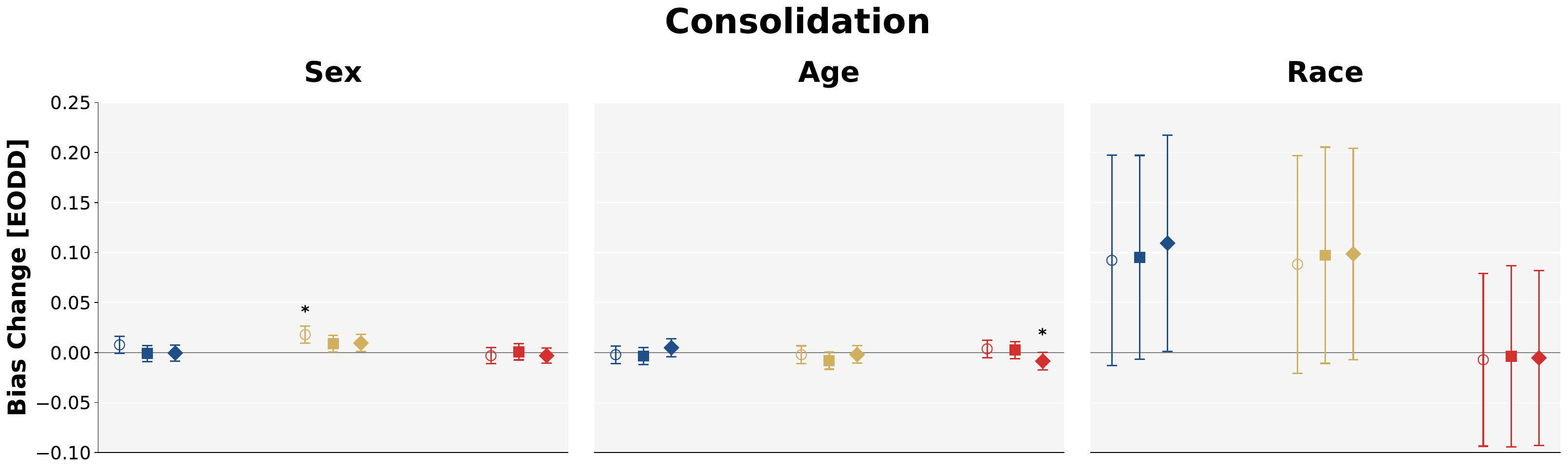}
\end{minipage}
}
&
\subfigure[]{%
\begin{minipage}[t]{0.49\textwidth}
\centering
\includegraphics[width=\linewidth]{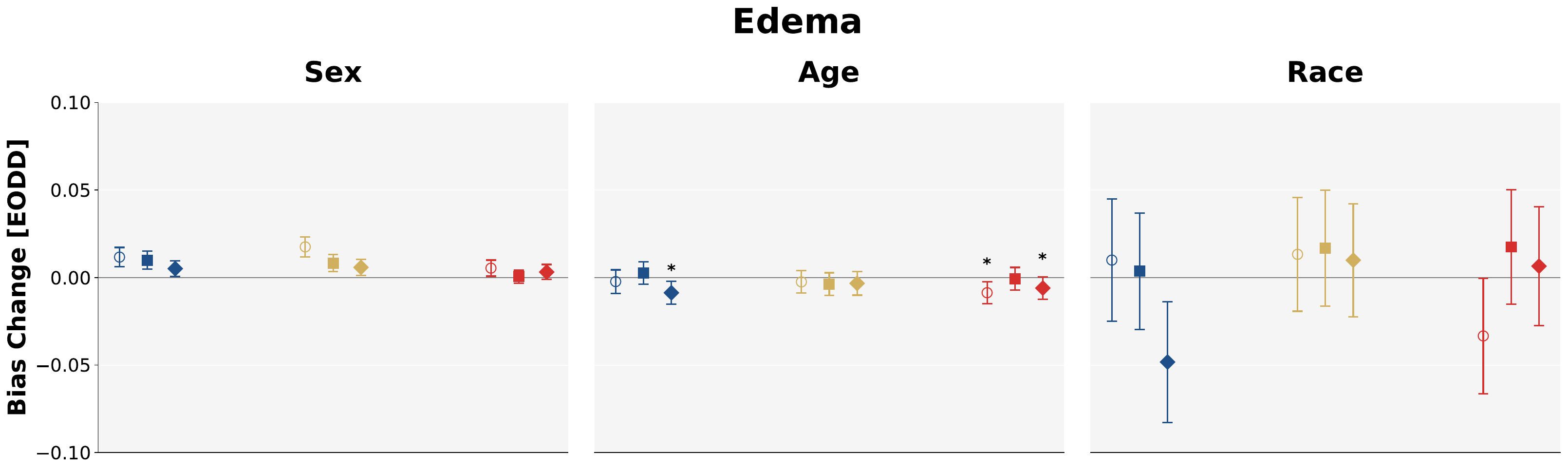}
\end{minipage}
}
\\[0.3cm]

\subfigure[]{%
\begin{minipage}[t]{0.49\textwidth}
\centering
\includegraphics[width=\linewidth]{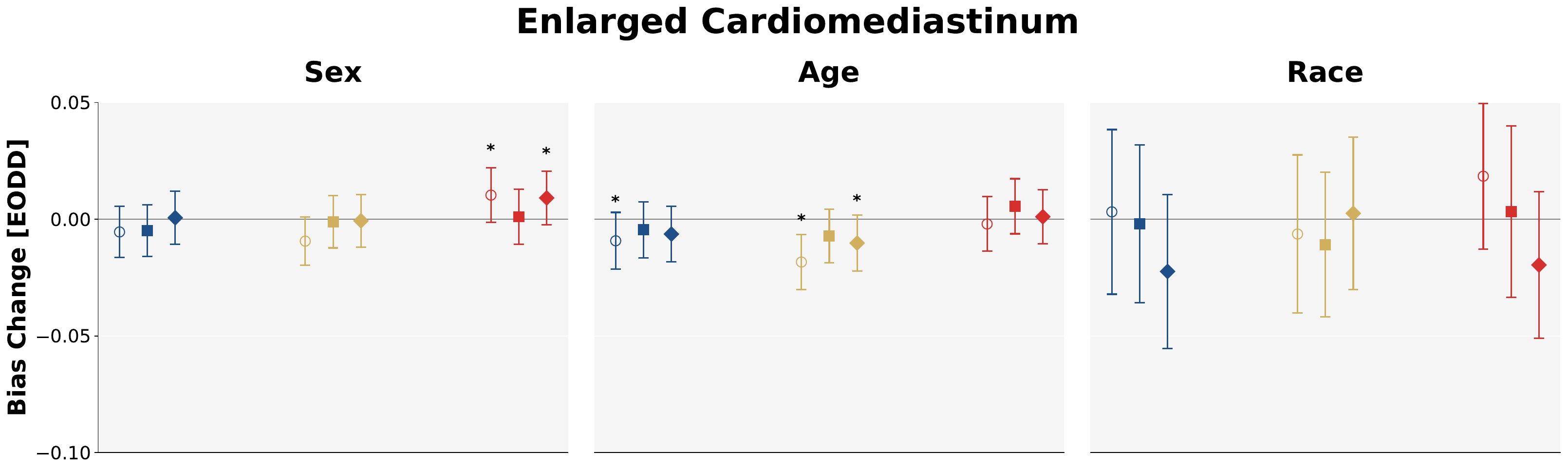}
\end{minipage}
}
&
\subfigure[]{%
\begin{minipage}[t]{0.49\textwidth}
\centering
\includegraphics[width=\linewidth]{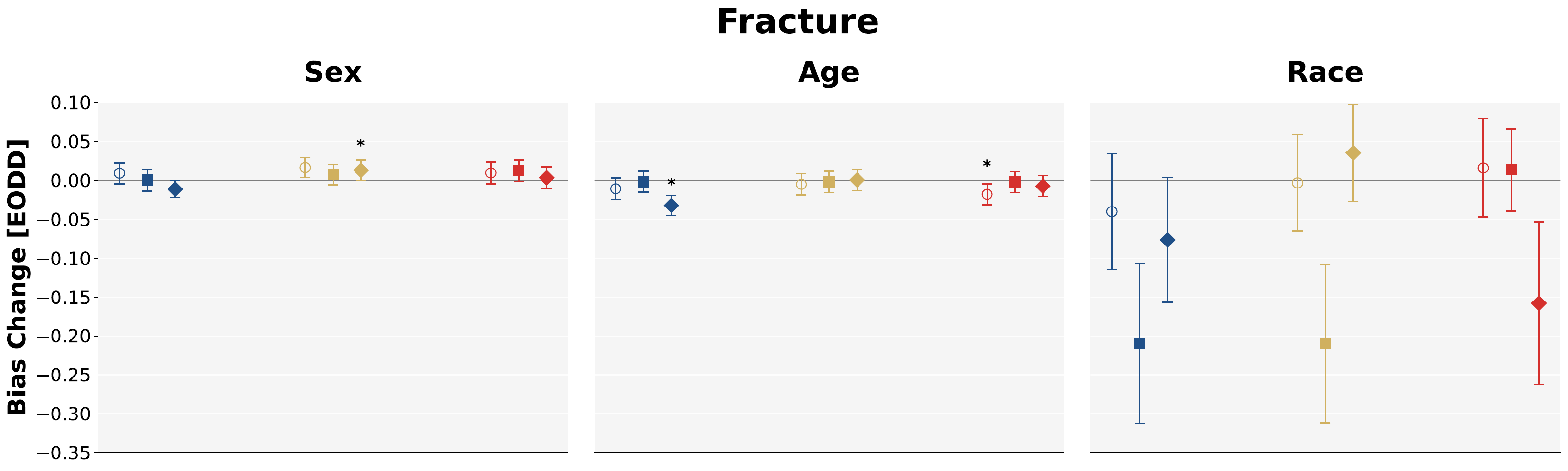}
\end{minipage}
}
\\[0.3cm]

\subfigure[]{%
\begin{minipage}[t]{0.49\textwidth}
\centering
\includegraphics[width=\linewidth]{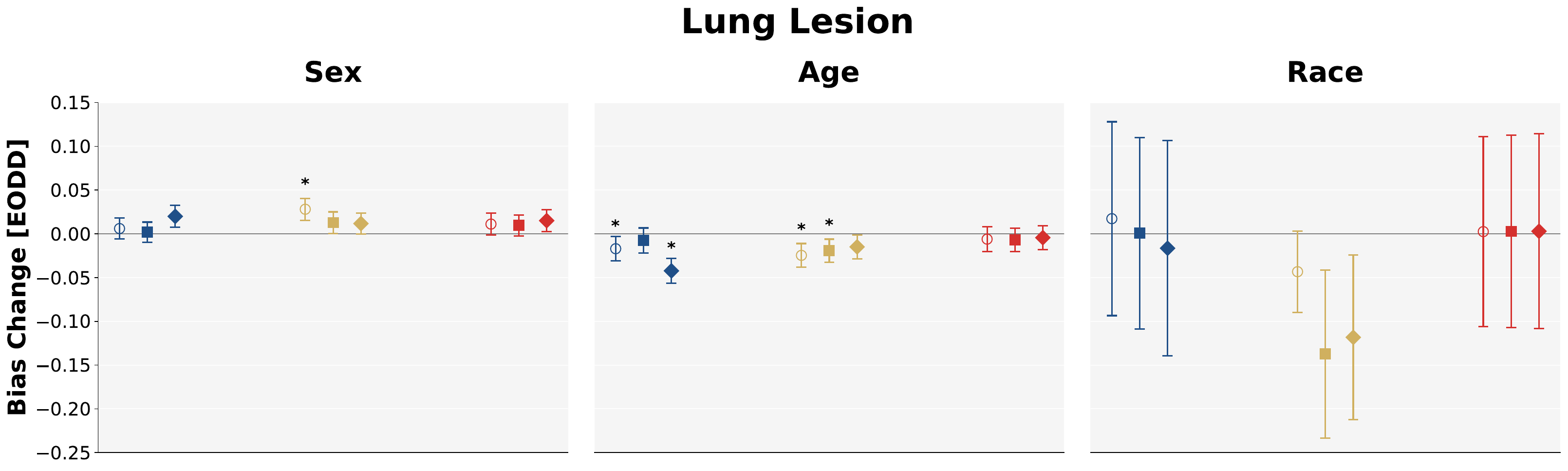}
\end{minipage}
}
&
\subfigure[]{%
\begin{minipage}[t]{0.49\textwidth}
\centering
\includegraphics[width=\linewidth]{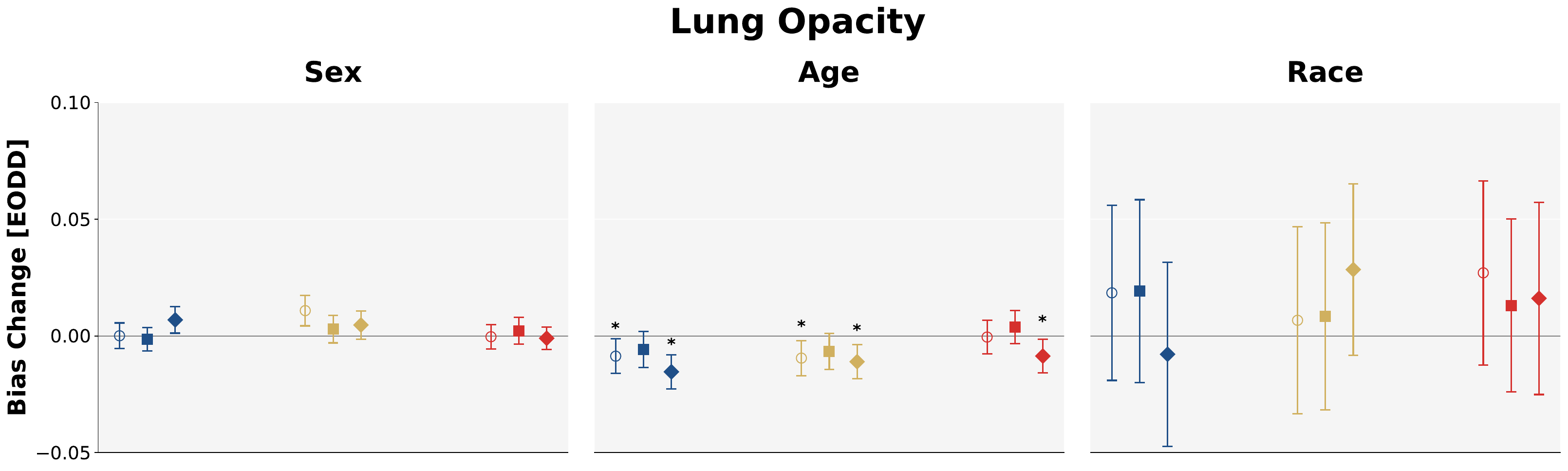}
\end{minipage}
}
\\[0.3cm]

\subfigure[]{%
\begin{minipage}[t]{0.49\textwidth}
\centering
\includegraphics[width=\linewidth]{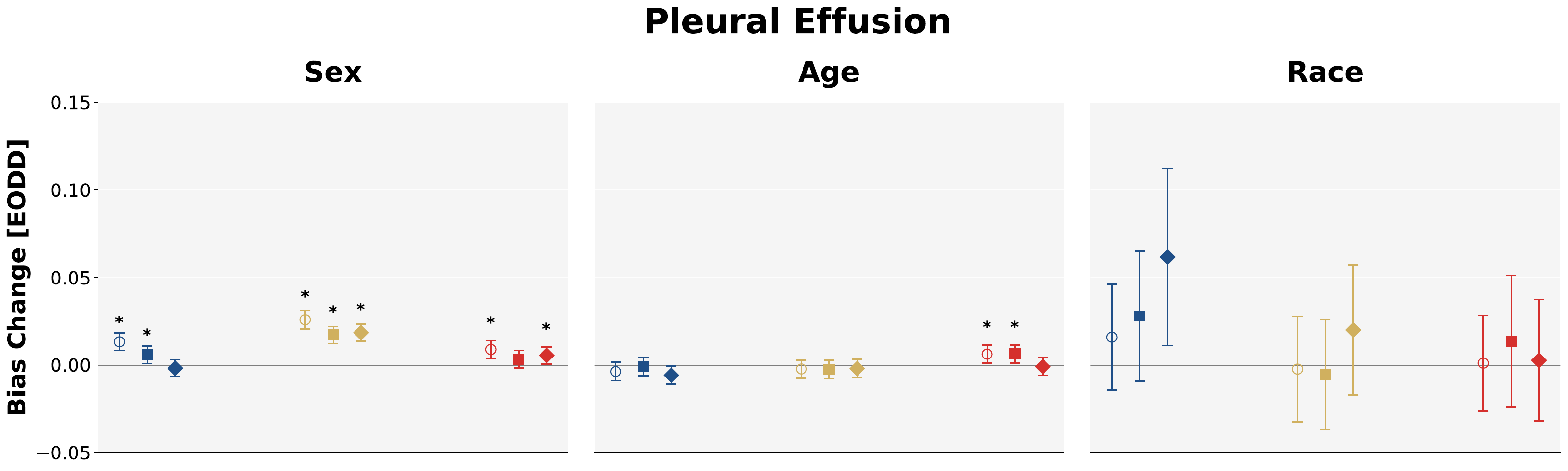}
\end{minipage}
}
&
\subfigure[]{%
\begin{minipage}[t]{0.49\textwidth}
\centering
\includegraphics[width=\linewidth]{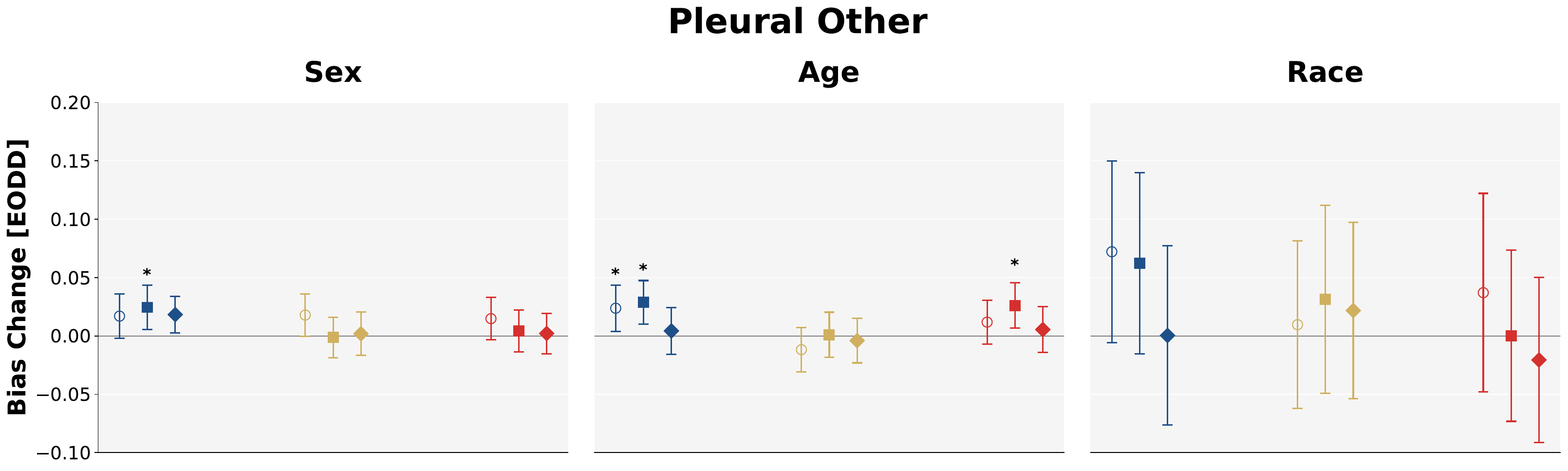}
\end{minipage}
}
\\[0.3cm]

\subfigure[]{%
\begin{minipage}[t]{0.49\textwidth}
\centering
\includegraphics[width=\linewidth]{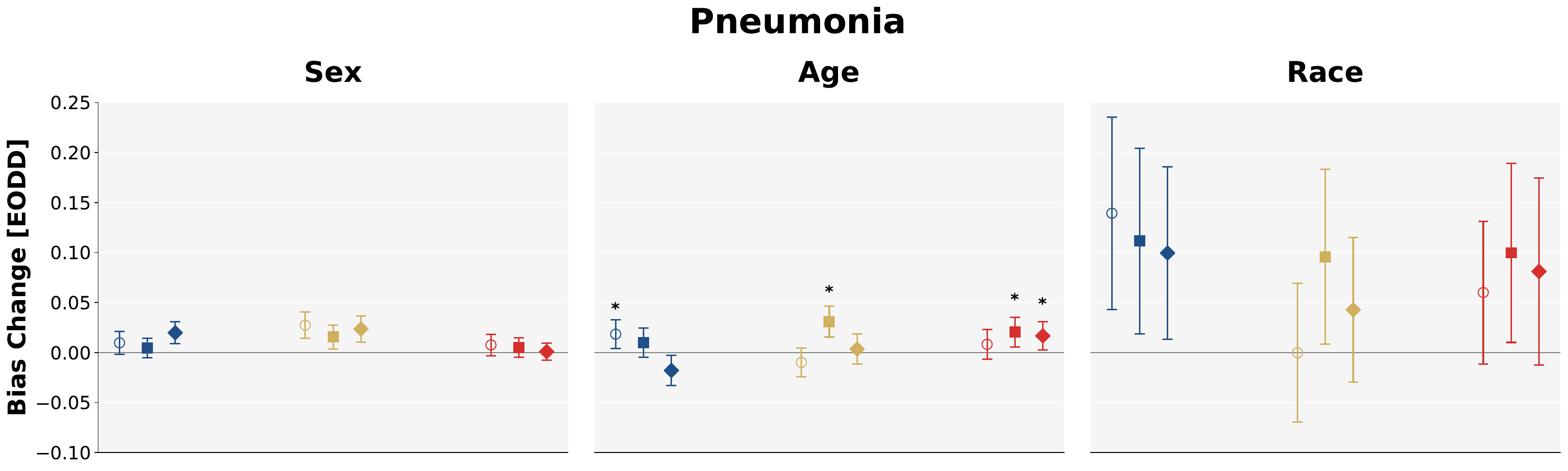}
\end{minipage}
}
&
\subfigure[]{%
\begin{minipage}[t]{0.49\textwidth}
\centering
\includegraphics[width=\linewidth]{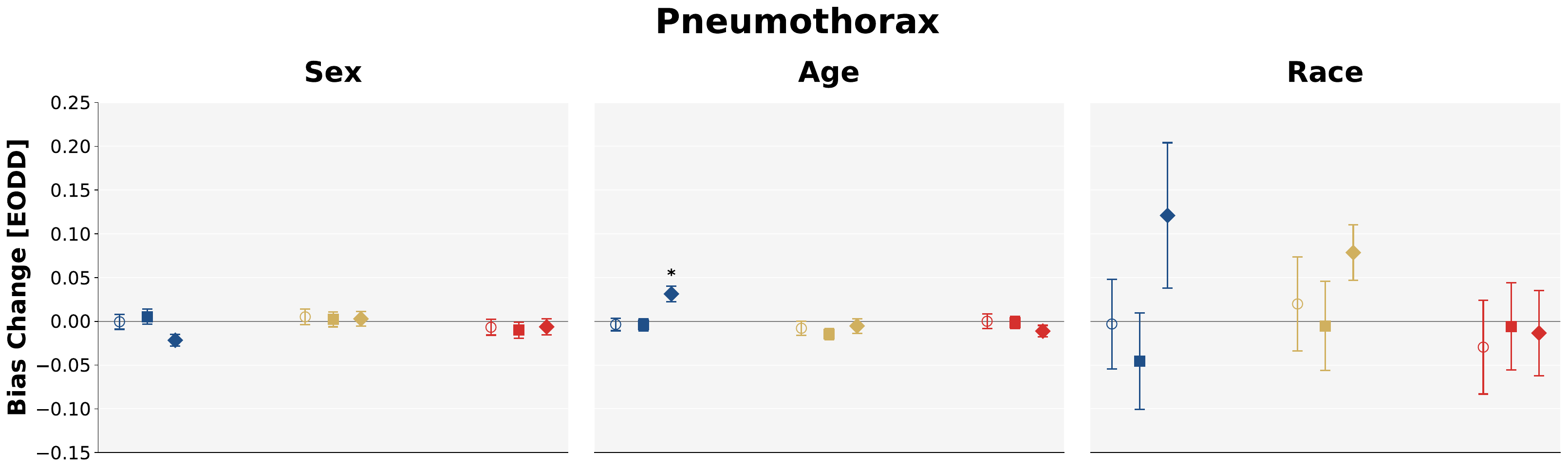}
\end{minipage}
}

\end{tabular}

\caption{Equalized odds bias change pre- and post-mitigation compared to predictions on original images for CheXpert. Pre-mitigation (``Reconstruction''), bias tends to increase slightly for sex; race exhibits high variance. Bias tends to decline slightly post-mitigation. Error bars represent standard deviation.}
\label{fig:bias_chexpert}

\end{figure*}

\begin{table}[ht]
\centering
  \begin{tabular}{l|ccc}
  \hline
                          & \multicolumn{1}{c|}{\textbf{Sex}} & \multicolumn{1}{c|}{\textbf{Age}} & {\textbf{Race}} \\ \hline
  \textbf{Classification} & $0.05 \pm 0.07$ & $0.17 \pm 0.08$ & $0.07 \pm 0.03$ \\ \hline
  \textbf{Segmentation}   & $1.10$ & $1.22$ & -- \\
  \hline
  \end{tabular}
  \caption{Baseline fairness of the classifiers (EODD) and the segmentation model (SER) for different sensitive attributes. For classification, mean and s.d. are reported across all classification tasks. Segmentation corresponds to UCSF-PDGM performance. Sex exhibits the lowest baseline bias.}
  \label{tab:baseline_bias}
\end{table}

Fig.~\ref{fig:histogram} displays the distribution of bias shifts when reconstructed images replace the original inputs. To provide a global overview, the histogram represents the bias shifts across all tasks, pathologies, and reconstruction models. As the diagnostic models exhibit bias on the original inputs (Table~\ref{tab:baseline_bias}), the bias shifts with reconstruction are plotted on a percentage scale compared to the original bias to highlight the relative effects. We find that the mode of these shifts is centered around zero, indicating little bias change in most instances. However, there is a noticeable tail towards positive bias changes, especially for sex, which exhibits a median increase of $24\%$. This is partly attributable to sex having a lower baseline bias than age and race (Table~\ref{tab:baseline_bias}).

The bias changes for each pathology and model are provided in Figures~\ref{fig:bias_chexpert} and~\ref{fig:bias_ucsf_class} (represented by the ``Reconstruction'' value in each plot).  
For UCSF-PDGM, no significant fairness deviations were observed when using the reconstructed images compared to the original images for either the segmentation or classification tasks (Figure~\ref{fig:bias_ucsf_class}). CheXpert shows more frequent significant shifts. Out of the 36 combinations (12 pathologies x 3 reconstruction models), there were 8 significant changes for sex (all in the positive direction) and 12 significant changes for age (4 in the positive direction). Due to large error bars, there were 0 significant changes for race, but alternative analysis which excluded subgroups with small sample sizes did reveal some significant changes (Appendix \ref{app:fairness_results}). Overall, the pathology-level findings support the histogram trend with a slight bias increase for sex and a slight decrease for age. The absolute magnitude of the effects were generally modest; however, some are of the order of a 0.05 change in EODD, corresponding to a 5\% difference in sensitivity/specificity, which is meaningful at the population level. Across reconstruction methods, the GAN and SDE-based models exhibited smaller bias shifts than the traditional U-Net (Table \ref{tab:model_bias}).

\begin{table}[ht]
\centering
  \begin{tabular}{l|ccc}
  \hline
                          & \multicolumn{1}{c|}{\textbf{U-Net}} & \multicolumn{1}{c|}{\textbf{GAN}} & {\textbf{SDE}} \\ \hline
  \textbf{Median}          & 2.28 & -0.21 & 1.59 \\ \hline
  \textbf{Absolute Median}            & 14.6 & 11.8 & 11.5 \\
  \hline
  \end{tabular}
    \caption{Median of bias change (\% change in EODD/SER) by reconstruction approach across all datasets, tasks, and attributes by model. SDE and GAN show a slightly smaller bias shift than U-Net.}  \label{tab:model_bias}
\end{table}

\subsection{Bias Mitigation}
While the impact of reconstruction on fairness was generally modest, applying mitigation strategies at the reconstruction stage could still reduce these effects or even improve the fairness of the underlying diagnostic models. We therefore tested two mitigation techniques inspired by classification literature but applied exclusively during reconstruction model training: sample reweighting and an equalized odds (EODD) constraint.

\begin{table}[ht]
\centering
  \begin{tabular}{l|ccc}
  \hline
                          & \multicolumn{1}{c|}{\textbf{Sex}} & \multicolumn{1}{c|}{\textbf{Age}} & {\textbf{Race}} \\ \hline
  \textbf{Standard}          & 24.1 & -1.88 & 3.30 \\ \hline
  \textbf{Reweighted}        & 10.6 & 0.03 & 1.05 \\ \hline
  \textbf{EODD}              & 7.56 & -2.01 & 0.52 \\ 
  \hline
  \end{tabular}
    \caption{Median bias change (\% change in EODD/SER) by mitigation strategy across all datasets, tasks, and models. Standard corresponds to the original results without mitigation applied. }  \label{tab:mitigation_bias}
\end{table}

\begin{figure*}[!t]
\centering

\includegraphics[width=\textwidth]{plots/fairness/eodd/evaluation_midl_camera_fairness_legend.pdf}

\vspace{-10pt}

\begin{minipage}[t]{0.32\textwidth}
\centering
\subfigure[]{%
\includegraphics[width=\linewidth]{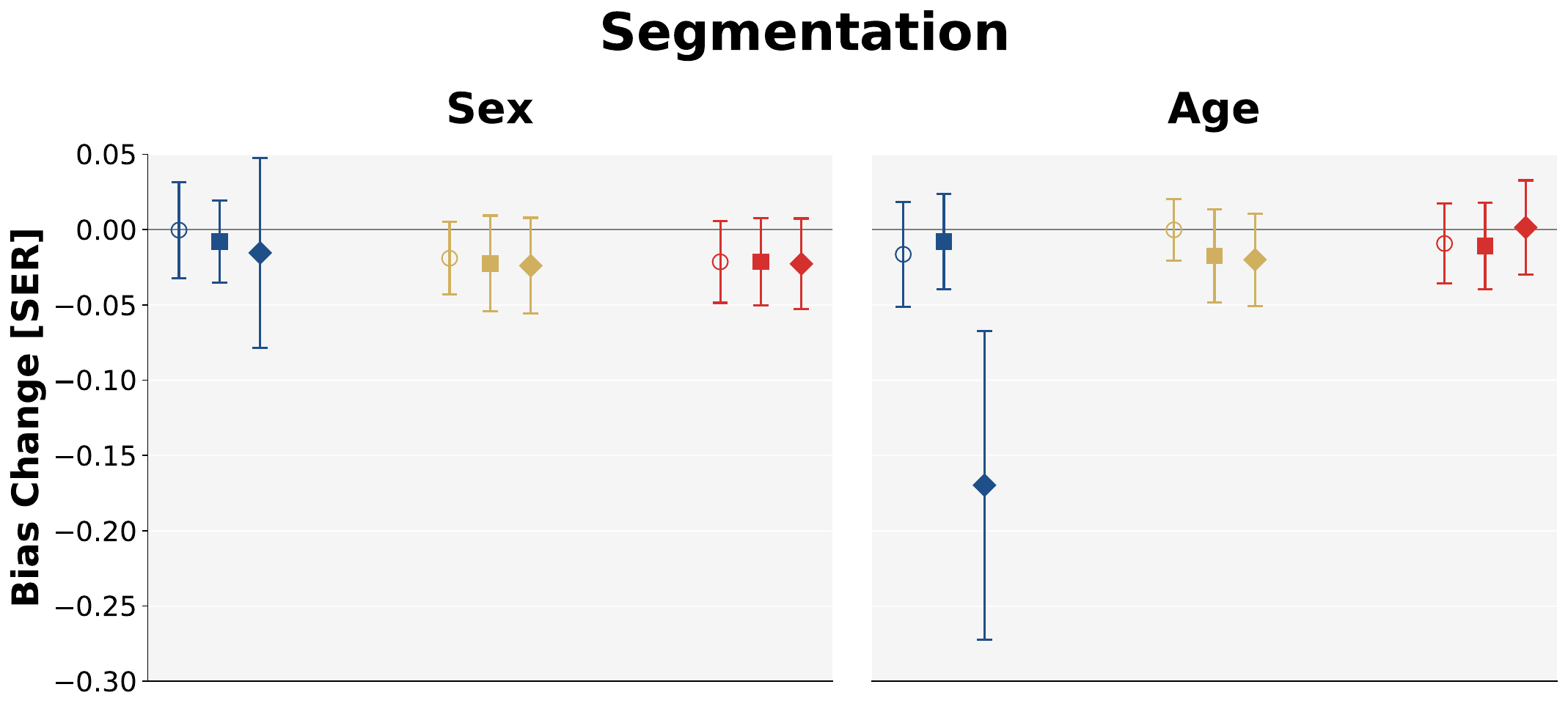}
}
\end{minipage}\hfill
\begin{minipage}[t]{0.32\textwidth}
\centering
\subfigure[]{%
\includegraphics[width=\linewidth]{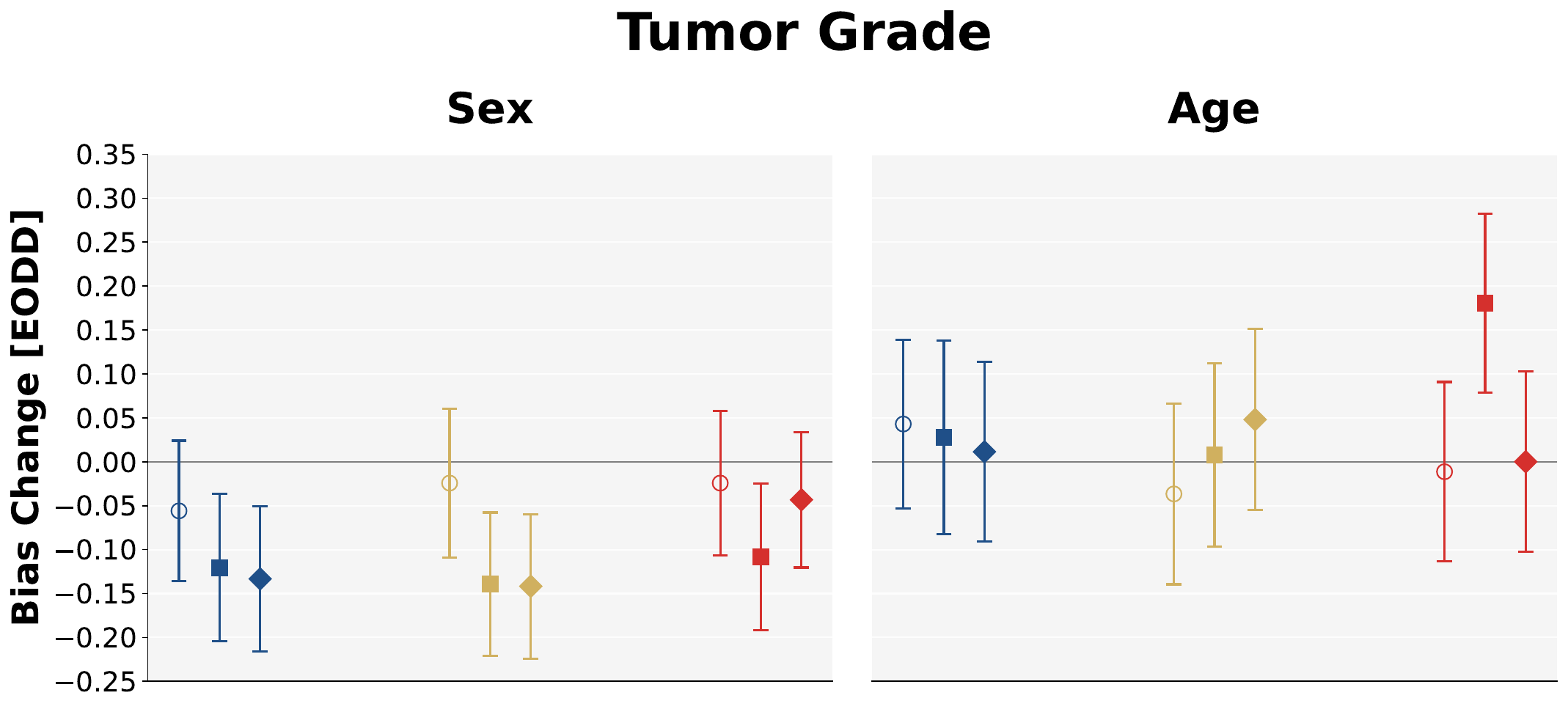}
}
\end{minipage}\hfill
\begin{minipage}[t]{0.32\textwidth}
\centering
\subfigure[]{%
\includegraphics[width=\linewidth]{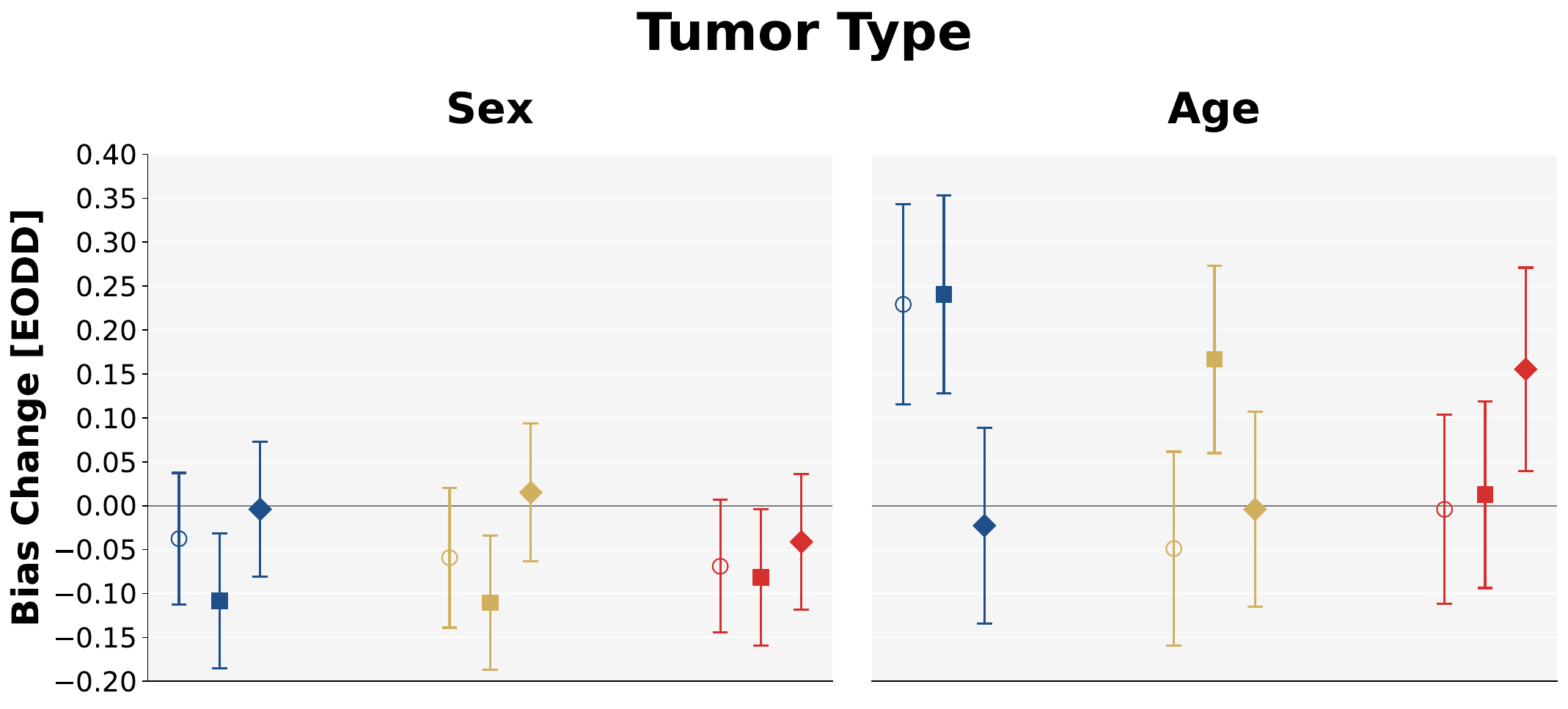}
}
\end{minipage}
\vspace{-10pt}
\caption{EODD and SER bias change pre- and post-mitigation compared to predictions on original images for UCSF-PDGM tasks. No consistent trends emerge for the classification tasks. Error bars represent standard deviation.}
\label{fig:bias_ucsf_class}

\end{figure*}

Table \ref{tab:mitigation_bias} summarizes the bias changes for the mitigated models compared to the standard models. The summary is presented as an aggregation over pathologies and reconstruction model types, with results for each combination presented in 
Figures \ref{fig:bias_chexpert} and \ref{fig:bias_ucsf_class}. 
Sex-related biases see the most substantial percentage improvements for both mitigation strategies.  The EODD mitigation approach exhibited slightly lower median bias for each sensitive attribute than the reweighted sampling strategy (Table \ref{tab:mitigation_bias}). For CheXpert the improved fairness for sex was  most notable for the U-Net and SDE models, and less so for the GAN-based Pix2Pix model (Figure \ref{fig:bias_chexpert}). For UCSF-PDGM segmentation, EODD reduced bias for most attributes and models, most strongly for U-Net (Figure~\ref{fig:bias_ucsf_class}). Classification fairness on UCSF-PDGM exhibits no consistent pattern, with fluctuations in both directions. Overall, while some fairness improvements are observed, the magnitudes are modest compared to the original bias (e.g., 16.5\% median improvement for sex with EODD mitigation) and can depend on the pathology and sensitive attribute. 

\begin{table}[ht]
\centering
  \setlength{\tabcolsep}{1mm}   
  \begin{tabular}{l|cc|cc} 
    \hline
            & \multicolumn{2}{c|}{\textbf{Reweighted}} & \multicolumn{2}{c}{\textbf{EODD}}  \\ \cline{2-5}
            & \textbf{Chex} & \textbf{UCSF} & \textbf{Chex} & \textbf{UCSF}  \\ \hline
    \textbf{PSNR}  & 0.54 & -0.75 & -0.64 & -7.28  \\ \hline
    \textbf{Down.}   & 0.07 & -1.97 &  0.02 & -2.94  \\ \hline
  \end{tabular}
  \caption{Mean change (\%) in PSNR and downstream performance (AUROC/Dice) per dataset after each mitigation averaged over reconstruction models and tasks. Performance drops are modest, except for PSNR in UCSF-PDGM for EODD.}
  \label{tab:fairness_perf}
\end{table}

Fairness gains can incur performance trade-offs, but the trade-offs observed here are modest. Table~\ref{tab:fairness_perf} reports the mean change in PSNR and downstream task performance across reconstruction models when the mitigation strategies are applied. CheXpert deviations are below \(1\,\%\) for PSNR and downstream AUROC. Downstream performance in UCSF-PDGM is also only moderately affected by the mitigation strategies, though PSNR shows larger drops with EODD (see Figure~\ref{app:mitigation_performance} in the Appendix). Reweighting incurs the smallest penalties overall.

Additional results using EOP and \(\Delta\)Dice fairness metrics before and after mitigation are provided in the Appendix (Figures~\ref{fig:bias_chexpert_eop} and~\ref{fig:bias_ucsf_class_eop}) and support the trends described above.
\section{Discussion}

We developed and applied an analysis framework that integrates reconstruction and prediction models to evaluate the effects of image reconstruction on downstream clinical performance and fairness, and investigate bias mitigation strategies at the reconstruction stage. Our analysis revealed several important insights, as summarized below.

\paragraph{Stability of Downstream Performance:} Despite notable reductions in image quality, indicated by decreased PSNR at higher noise levels, downstream segmentation and classification performances remained robust to image reconstruction. This stability suggests that current diagnostic models are largely resilient to reconstruction-induced image degradations, at least for the studied tasks, which implies that minor reconstruction noise might not adversely impact clinical diagnostic accuracy. This finding may be surprising given that deep learning classification models are often thought to lack robustness, such as showing changes if the data are heterogeneous or noisy \cite{biomedinformatics4020050}. This suggests a nuanced interpretation of robustness, where models may be robust to certain transformations (e.g., reconstruction noise) but not others. Critically, these findings also have implications for the studied reconstruction models. Even if the downstream models were robust, we would expect that the performance of these models would drop if the reconstruction models removed the true underlying information necessary to perform the tasks. Instead, we observe largely stable downstream performance even as PSNR decreases, suggesting retainment of diagnostic features for the studied tasks. Nonetheless, there was a mild dependence on task difficulty, where more subtle pathologies (e.g. lung lesions) showed larger performance effects, highlighting future opportunities of applying our framework to other subtle tasks where the implications of AI-based reconstruction are currently unknown. 

\FloatBarrier

\paragraph{Fairness Implications and Variability:} The aggregate effect of reconstruction on fairness was relatively modest, though certain pathologies and sensitive attributes showed significant shifts. These shifts varied in magnitude and direction, with a tendency toward increased bias, especially for patient sex. In most cases, the magnitude represented only a small fraction of the bias already present in the diagnostic models, though some would correspond to a \textasciitilde5\% difference in sensitivity/specificity between subgroups. Thus, reconstruction can contribute to bias in downstream tasks, but the overall bias appears to be largely driven by the downstream models themselves. 

\paragraph{Effectiveness of Mitigation Techniques:} Mitigation strategies reduced sex-related biases on CheXpert without measurable performance trade-offs in AUROC or PSNR (Figure~\ref{app:mitigation_performance} in the Appendix). However, similar mitigation strategies yielded inconsistent results on UCSF-PDGM, highlighting that their effectiveness is dataset-specific and dependent on the underlying task complexity and dataset characteristics.

\paragraph{Sensitivity of Model Choice:} 
The SDE and GAN-based reconstruction approaches introduced lower additional bias overall compared to the standard U-Net, which may be counterintuitive given the generative nature of the SDE and GAN models. The U-Net also exhibited larger degradations in downstream performance when fine-tuned with the fairness mitigation strategies (Figure \ref{app:mitigation_performance} in the Appendix). This sensitivity may arise from its inherently lower capacity than other methods, limiting simultaneous optimization of image fidelity and fairness constraints. 

\paragraph{Summary of Clinical Implications:}
The robustness of downstream performance to AI-based image reconstruction is encouraging, particularly as these technologies are increasingly integrated into clinical practice. However, some performance drops were observed, especially for more subtle pathologies, highlighting the importance of rigorous evaluation and real-world monitoring. The potential for fairness shifts also necessitates active monitoring and reporting. This is especially important because model behavior can change as data distributions shift.

\paragraph{Summary of Model Development Implications:}
Developers of reconstruction models should prioritize downstream task and fairness evaluations alongside traditional pixel-level metrics, recognizing that reconstruction-induced biases, though subtle, can propagate through diagnostic workflows. This is especially the case for patient sex, where anatomical differences can be more prominent and may explain the larger effects observed for this attribute in our results. Bias mitigation strategies applied at the reconstruction stage may help improve fairness, but our results suggest that direct intervention at the classifier stage should be prioritized. Future research should explore multi-stage bias mitigation, integrating reconstruction and classification levels to achieve balanced fairness and performance outcomes.

\paragraph{Limitations:} For comprehensiveness, we assessed multiple reconstruction models, downstream tasks, pathologies, and mitigation strategies, but this breadth necessarily creates challenges in data interpretation. As such, we have provided both summary level (e.g., Figure~\ref{fig:histogram}) and individual (e.g., Figure~\ref{fig:bias_chexpert}) results to enhance interpretability. Along with our studied datasets and tasks, it will be important in future work to apply our framework to additional datasets and clinical populations, including larger MRI datasets, to further probe generalization. 
Further generalization assessments for MRI should include additional sequences, directly modeling of k-space data rather than synthetic undersampling, and the inclusion of measurement noise rather than undersampling only.
We note that our framework relies on the existence of adequate downstream models, which may not exist for every intended clinical task and dataset. 
However, the framework is agnostic to the type of downstream model, and could be applied to volumetric or temporal models, or could rely on self-supervised models before fine-tuning when labels are scarce.
Additionally, while the algorithms used to create noisy images in this study simulate realistic acquisition degradations and are common approaches in the field \cite{FengRadial, Gibson2023APM}, they may not fully capture real-world variations.
\FloatBarrier
\section{Conclusion}
The increasing clinical prevalence of AI-based reconstruction models creates a critical need for quantitative assessments of their potential downstream impact.
We performed a scalable evaluation by using reconstruction and diagnostic AI models in tandem across multiple datasets, tasks, pathologies, and model types. We view our results as largely positive for the field -- downstream performance was much more robust to reconstruction noise than image-level metrics, and the biases introduced by reconstruction were generally modest. However, some trends of increased bias were observed, especially for patient sex. Altogether, supported by these findings, we argue for the importance of monitoring downstream performance and fairness when using AI-based reconstruction models, and for continued work to mitigate emerging biases.

\clearpage

\bibliography{midl26_37}

@ARTICLE{Diab2025-fm,
  title     = "Distinguishing between rigor and transparency in {FDA} marketing
               authorization of {AI-enabled} medical devices",
  author    = "Diab, Abdul Rahman and Lotter, William",
  journal   = "Radiol. Artif. Intell.",
  publisher = "Radiological Society of North America (RSNA)",
  volume    =  7,
  number    =  6,
  pages     = "e250369",
  month     =  nov,
  year      =  2025,
  language  = "en"
}

@ARTICLE{Sultan2025-co,
  title     = "An unsupervised method for {MRI} recovery: deep image prior with
               structured sparsity",
  author    = "Sultan, Muhammad Ahmad and Chen, Chong and Liu, Yingmin and Gil,
               Katarzyna and Zareba, Karolina and Ahmad, Rizwan",
  journal   = "MAGMA",
  publisher = "Springer Science and Business Media LLC",
  volume    =  38,
  number    =  5,
  pages     = "859--871",
  month     =  oct,
  year      =  2025,
  copyright = "https://creativecommons.org/licenses/by/4.0",
  language  = "en"
}

@ARTICLE{Chen2023-vn,
  title    = "Deep learning based unpaired image-to-image translation
              applications for medical physics: a systematic review",
  author   = "Chen, Junhua and Chen, Shenlun and Wee, Leonard and Dekker, Andre
              and Bermejo, Inigo",
  journal  = "Phys. Med. Biol.",
  volume   =  68,
  number   =  5,
  month    =  feb,
  year     =  2023,
  language = "en"
}

@article{CHEN2026102015,
title = {A multi-dynamic low-rank deep image prior for three-dimensional real-time cardiovascular magnetic resonance imaging},
journal = {Journal of Cardiovascular Magnetic Resonance},
volume = {28},
number = {1},
pages = {102015},
year = {2026},
issn = {1097-6647},
doi = {https://doi.org/10.1016/j.jocmr.2025.102015},
url = {https://www.sciencedirect.com/science/article/pii/S1097664725001772},
author = {Chong Chen and Marc Vornehm and Zhenyu Bu and Preethi Chandrasekaran and Muhammad A. Sultan and Syed M. Arshad and Yingmin Liu and Yuchi Han and Rizwan Ahmad}
}

@ARTICLE{Lee2024-ic,
  title     = "Image quality and diagnostic performance of low-dose liver {CT}
               with deep learning reconstruction versus standard-dose {CT}",
  author    = "Lee, Dong Ho and Lee, Jeong Min and Lee, Chang Hee and Afat,
               Saif and Othman, Ahmed",
  journal   = "Radiol. Artif. Intell.",
  publisher = "Radiological Society of North America (RSNA)",
  volume    =  6,
  number    =  2,
  pages     = "e230192",
  month     =  mar,
  year      =  2024,
  language  = "en"
}

@ARTICLE{Feuerriegel2023-rt,
  title    = "Evaluation of a deep learning-based reconstruction method for
              denoising and image enhancement of shoulder {MRI} in patients
              with shoulder pain",
  author   = "Feuerriegel, Georg C and Weiss, Kilian and Kronthaler, Sophia and
              Leonhardt, Yannik and Neumann, Jan and Wurm, Markus and Lenhart,
              Nicolas S and Makowski, Marcus R and Schwaiger, Benedikt J and
              Woertler, Klaus and Karampinos, Dimitrios C and Gersing,
              Alexandra S",
  journal  = "Eur. Radiol.",
  volume   =  33,
  number   =  7,
  pages    = "4875--4884",
  month    =  jul,
  year     =  2023,
  language = "en"
}

@article{Burlina2021Retinal,
author = {Burlina, Philippe and Joshi, Neil and Paul, William and Pacheco, Katia and Bressler, Neil},
year = {2021},
month = {02},
pages = {13},
title = {Addressing Artificial Intelligence Bias in Retinal Diagnostics},
volume = {10},
journal = {Translational Vision Science  Technology},
doi = {10.1167/tvst.10.2.13}
}

@article{Kalantari2021UnderdiagnosisCheX,
author = {Seyyed-Kalantari, Laleh and Zhang, Haoran and McDermott, Matthew and Chen, Irene and Ghassemi, Marzyeh},
year = {2021},
month = {12},
pages = {},
title = {Underdiagnosis bias of artificial intelligence algorithms applied to chest radiographs in under-served patient populations},
volume = {27},
journal = {Nature Medicine},
doi = {10.1038/s41591-021-01595-0}
}

@article{Stanley2022FairnessrelatedPA,
  title={Fairness-related performance and explainability effects in deep learning models for brain image analysis},
  author={Emma A. M. Stanley and Matthias Wilms and Pauline Mouches and Nils Daniel Forkert},
  journal={Journal of Medical Imaging},
  year={2022},
  volume={9},
  pages={061102 - 061102},
  url={https://api.semanticscholar.org/CorpusID:251876386}
}

@article{saumure2025humor,
  title={Humor as a window into generative AI bias},
  author={Saumure, R. and De Freitas, J. and Puntoni, S.},
  journal={Scientific Reports},
  volume={15},
  number={1},
  pages={1326},
  year={2025},
  doi={10.1038/s41598-024-83384-6},
  url={https://doi.org/10.1038/s41598-024-83384-6}
}

@inproceedings{ruggeri-nozza-2023-multi,
    title = "A Multi-dimensional study on Bias in Vision-Language models",
    author = "Ruggeri, Gabriele  and
      Nozza, Debora",
    booktitle = "Findings of the Association for Computational Linguistics: ACL 2023",
    month = jul,
    year = "2023",
    address = "Toronto, Canada",
    publisher = "Association for Computational Linguistics",
    url = "https://aclanthology.org/2023.findings-acl.403/",
    doi = "10.18653/v1/2023.findings-acl.403",
    pages = "6445--6455",
}

@inproceedings{luccioni2023stablebiasanalyzingsocietal,
author = {Luccioni, Alexandra Sasha and Akiki, Christopher and Mitchell, Margaret and Jernite, Yacine},
title = {Stable bias: evaluating societal representations in diffusion models},
year = {2023},
publisher = {Curran Associates Inc.},
address = {Red Hook, NY, USA},
booktitle = {Proceedings of the 37th International Conference on Neural Information Processing Systems},
articleno = {2458},
numpages = {14},
location = {New Orleans, LA, USA},
series = {NIPS '23}
}

@article{CHIU2024103188,
title = {Achieve fairness without demographics for dermatological disease diagnosis},
journal = {Medical Image Analysis},
volume = {95},
pages = {103188},
year = {2024},
issn = {1361-8415},
doi = {https://doi.org/10.1016/j.media.2024.103188},
url = {https://www.sciencedirect.com/science/article/pii/S1361841524001130},
author = {Ching-Hao Chiu and Yu-Jen Chen and Yawen Wu and Yiyu Shi and Tsung-Yi Ho},
}

@article{Glocker2023AlgorithmicEO,
  title={Algorithmic encoding of protected characteristics in chest X-ray disease detection models},
  author={Ben Glocker and Charles Jones and M{\'e}lanie Bernhardt and Stefan Winzeck},
  journal={eBioMedicine},
  year={2023},
  volume={89},
  url={https://api.semanticscholar.org/CorpusID:256858498}
}

@article{WANG2024105047,
title = {Drop the shortcuts: image augmentation improves fairness and decreases AI detection of race and other demographics from medical images},
journal = {eBioMedicine},
volume = {102},
pages = {105047},
year = {2024},
issn = {2352-3964},
doi = {https://doi.org/10.1016/j.ebiom.2024.105047},
url = {https://www.sciencedirect.com/science/article/pii/S2352396424000823},
author = {Ryan Wang and Po-Chih Kuo and Li-Ching Chen and Kenneth Patrick Seastedt and Judy Wawira Gichoya and Leo Anthony Celi},
}

@inproceedings{ioannou2022studydemographicbiascnnbased,
author = {Ioannou, Stefanos and Chockler, Hana and Hammers, Alexander and King, Andrew P.},
title = {A Study of Demographic Bias in CNN-Based Brain MR Segmentation},
year = {2022},
isbn = {978-3-031-17898-6},
publisher = {Springer-Verlag},
address = {Berlin, Heidelberg},
url = {https://doi.org/10.1007/978-3-031-17899-3_2},
doi = {10.1007/978-3-031-17899-3_2},
booktitle = {Machine Learning in Clinical Neuroimaging: 5th International Workshop, MLCN 2022, Held in Conjunction with MICCAI 2022, Singapore, September 18, 2022, Proceedings},
pages = {13–22},
numpages = {10},
location = {Singapore, Singapore}
}

@inproceedings{lee2022systematicstudyracesex,
author = {Lee, Tiarna and Puyol-Ant\'{o}n, Esther and Ruijsink, Bram and Shi, Miaojing and King, Andrew P.},
title = {A Systematic Study of Race and Sex Bias in CNN-Based Cardiac MR Segmentation},
year = {2022},
isbn = {978-3-031-23442-2},
publisher = {Springer-Verlag},
address = {Berlin, Heidelberg},
url = {https://doi.org/10.1007/978-3-031-23443-9_22},
doi = {10.1007/978-3-031-23443-9_22},
booktitle = {Statistical Atlases and Computational Models of the Heart. Regular and CMRxMotion Challenge Papers: 13th International Workshop, STACOM 2022, Held in Conjunction with MICCAI 2022},
pages = {233–244},
numpages = {12},
location = {Singapore, Singapore}
}

@article{puyol2022fairness,
  title={Fairness in Cardiac Magnetic Resonance Imaging: Assessing Sex and Racial Bias in Deep Learning-Based Segmentation},
  author={Puyol-Ant{\'o}n, E and Ruijsink, B and Mariscal Harana, J and Piechnik, SK and Neubauer, S and Petersen, SE and Razavi, R and Chowienczyk, P and King, AP},
  journal={Frontiers in Cardiovascular Medicine},
  volume={9},
  pages={859310},
  year={2022},
  month={Apr},
  doi={10.3389/fcvm.2022.859310},
  pmid={35463778},
  pmcid={PMC9021445}
}

@inproceedings{du2023unveilingfairnessbiasesdeep,
author = {Du, Yuning and Xue, Yuyang and Dharmakumar, Rohan and Tsaftaris, Sotirios A.},
title = {Unveiling Fairness Biases in Deep Learning-Based Brain MRI Reconstruction},
year = {2023},
isbn = {978-3-031-45248-2},
publisher = {Springer-Verlag},
address = {Berlin, Heidelberg},
url = {https://doi.org/10.1007/978-3-031-45249-9_10},
doi = {10.1007/978-3-031-45249-9_10},
booktitle = {Clinical Image-Based Procedures,  Fairness of AI in Medical Imaging, and Ethical and Philosophical Issues in Medical Imaging: 12th International Workshop, CLIP 2023 1st International Workshop, FAIMI 2023 and 2nd International Workshop, EPIMI 2023},
pages = {102–111},
numpages = {10},
location = {Vancouver, BC, Canada}
}

@inproceedings{Sheg24reconbias,
author = {Sheng, Yi and Yang, Junhuan and Lin, Youzuo and Jiang, Weiwen and Yang, Lei},
title = {Toward Fair Ultrasound Computing Tomography: Challenges, Solutions and Outlook},
year = {2024},
isbn = {9798400706059},
publisher = {Association for Computing Machinery},
address = {New York, NY, USA},
url = {https://doi.org/10.1145/3649476.3660387},
doi = {10.1145/3649476.3660387},
booktitle = {Proceedings of the Great Lakes Symposium on VLSI 2024},
pages = {748–753},
numpages = {6},
location = {Clearwater, FL, USA},
series = {GLSVLSI '24}
}

@article{mehrabi2022surveybiasfairnessmachine,
author = {Mehrabi, Ninareh and Morstatter, Fred and Saxena, Nripsuta and Lerman, Kristina and Galstyan, Aram},
title = {A Survey on Bias and Fairness in Machine Learning},
year = {2021},
issue_date = {July 2022},
publisher = {Association for Computing Machinery},
address = {New York, NY, USA},
volume = {54},
number = {6},
issn = {0360-0300},
url = {https://doi.org/10.1145/3457607},
doi = {10.1145/3457607},
journal = {ACM Comput. Surv.},
month = jul,
articleno = {115},
numpages = {35}
}

@inproceedings{DBLP:journals/corr/HardtPS16,
author = {Hardt, Moritz and Price, Eric and Srebro, Nathan},
title = {Equality of opportunity in supervised learning},
year = {2016},
isbn = {9781510838819},
publisher = {Curran Associates Inc.},
address = {Red Hook, NY, USA},
booktitle = {Proceedings of the 30th International Conference on Neural Information Processing Systems},
pages = {3323–3331},
numpages = {9},
location = {Barcelona, Spain},
series = {NIPS'16}
}

@InProceedings{Kinyanyui2020Dermatology,
author="Kinyanjui, Newton M.
and Odonga, Timothy
and Cintas, Celia
and Codella, Noel C. F.
and Panda, Rameswar
and Sattigeri, Prasanna
and Varshney, Kush R.",
title="Fairness of Classifiers Across Skin Tones in Dermatology",
booktitle="Medical Image Computing and Computer Assisted Intervention -- MICCAI 2020",
year="2020",
publisher="Springer International Publishing",
address="Cham",
pages="320--329",
isbn="978-3-030-59725-2"
}

@InProceedings{densenet,
author = {Huang, Gao and Liu, Zhuang and van der Maaten, Laurens and Weinberger, Kilian Q.},
title = {Densely Connected Convolutional Networks},
booktitle = {Proceedings of the IEEE Conference on Computer Vision and Pattern Recognition (CVPR)},
month = {July},
year = {2017}
}

@article{resnet,
  author       = {Kaiming He and
                  Xiangyu Zhang and
                  Shaoqing Ren and
                  Jian Sun},
  title        = {Deep Residual Learning for Image Recognition},
  journal      = {arXiv},
  volume       = {abs/1512.03385},
  year         = {2015},
  url          = {http://arxiv.org/abs/1512.03385},
  eprinttype    = {arXiv},
  eprint       = {1512.03385},
  bibsource    = {dblp computer science bibliography, https://dblp.org}
}

@article{jamanetworkopen.2023.42203,
    author = {Yuan, Chenxi and Linn, Kristin A. and Hubbard, Rebecca A.},
    title = {Algorithmic Fairness of Machine Learning Models for Alzheimer Disease Progression},
    journal = {JAMA Network Open},
    volume = {6},
    number = {11},
    pages = {e2342203-e2342203},
    year = {2023},
    month = {11},
    issn = {2574-3805},
    doi = {10.1001/jamanetworkopen.2023.42203},
    url = {https://doi.org/10.1001/jamanetworkopen.2023.42203},
    eprint = {https://jamanetwork.com/journals/jamanetworkopen/articlepdf/2811461/yuan\_2023\_oi\_231221\_1698415358.87588.pdf},
}

@inproceedings{Oguguo23,
  author={Oguguo, Tochi and Zamzmi, Ghada and Rajaraman, Sivaramakrishnan and Yang, Feng and Xue, Zhiyun and Antani, Sameer},
  booktitle={2023 IEEE 20th International Symposium on Biomedical Imaging (ISBI)}, 
  title={A Comparative Study of Fairness in Medical Machine Learning}, 
  year={2023},
  volume={},
  number={},
  pages={1-5},
  doi={10.1109/ISBI53787.2023.10230368}}

@INPROCEEDINGS {bissoto2019deconstructingbiasskinlesion,
author = { Bissoto, Alceu and Fornaciali, Michel and Valle, Eduardo and Avila, Sandra },
booktitle = { 2019 IEEE/CVF Conference on Computer Vision and Pattern Recognition Workshops (CVPRW) },
title = {{ (De) Constructing Bias on Skin Lesion Datasets }},
year = {2019},
volume = {},
ISSN = {},
pages = {2766-2774},
doi = {10.1109/CVPRW.2019.00335},
url = {https://doi.ieeecomputersociety.org/10.1109/CVPRW.2019.00335},
publisher = {IEEE Computer Society},
address = {Los Alamitos, CA, USA},
month =Jun}

@InProceedings{marcinkevičs2022debiasingdeepchestxray,
  title = 	 {Debiasing Deep Chest {X}-Ray Classifiers using Intra- and Post-processing Methods},
  author =       {Marcinkevics, Ricards and Ozkan, Ece and Vogt, Julia E.},
  booktitle = 	 {Proceedings of the 7th Machine Learning for Healthcare Conference},
  pages = 	 {504--536},
  year = 	 {2022},
  volume = 	 {182},
  series = 	 {Proceedings of Machine Learning Research},
  month = 	 {05--06 Aug},
  publisher =    {PMLR},
  url = 	 {https://proceedings.mlr.press/v182/marcinkevics22a.html},
}

@ARTICLE{Yang2024-dm,
  title     = "The limits of fair medical imaging {AI} in real-world
               generalization",
  author    = "Yang, Yuzhe and Zhang, Haoran and Gichoya, Judy W and Katabi,
               Dina and Ghassemi, Marzyeh",
  journal   = "Nat. Med.",
  publisher = "Springer Science and Business Media LLC",
  volume    =  30,
  number    =  10,
  pages     = "2838--2848",
  month     =  oct,
  year      =  2024,
  copyright = "https://creativecommons.org/licenses/by/4.0",
  language  = "en"
}

@ARTICLE{Lotter2024-sk,
  title     = "Acquisition parameters influence {AI} recognition of race in
               chest x-rays and mitigating these factors reduces underdiagnosis
               bias",
  author    = "Lotter, William",
  journal   = "Nat. Commun.",
  publisher = "Springer Science and Business Media LLC",
  volume    =  15,
  number    =  1,
  pages     = "7465",
  month     =  aug,
  year      =  2024,
  copyright = "https://creativecommons.org/licenses/by-nc-nd/4.0",
  language  = "en"
}

@InProceedings{wu2022fairpruneachievingfairnesspruning,
author="Wu, Yawen
and Zeng, Dewen
and Xu, Xiaowei
and Shi, Yiyu
and Hu, Jingtong",
title="FairPrune: Achieving Fairness Through Pruning for Dermatological Disease Diagnosis",
booktitle="Medical Image Computing and Computer Assisted Intervention -- MICCAI 2022",
year="2022",
publisher="Springer Nature Switzerland",
address="Cham",
pages="743--753",
isbn="978-3-031-16431-6"
}

@article{FengRadial,
author = {Feng, Li},
year = {2022},
month = {04},
pages = {},
title = {Golden‐Angle Radial MRI: Basics, Advances, and Applications},
volume = {56},
journal = {Journal of Magnetic Resonance Imaging},
doi = {10.1002/jmri.28187}
}

@article{SiddiquiFairSeg,
author = {Siddiqui, Ismaeel and Littlefield, Nickolas and Carlson, Luke and Gong, Matthew and Chhabra, Avani and Menezes, Zoe and Mastorakos, George and Thakar, Sakshi and Abedian, Mehrnaz and Lohse, Ines and Weiss, Kurt and Plate, Johannes and Moradi, Hamidreza and Amirian, Soheyla and P. Tafti, Ahmad},
year = {2024},
month = {07},
pages = {},
title = {Fair AI-powered orthopedic image segmentation: addressing bias and promoting equitable healthcare},
volume = {14},
journal = {Scientific Reports},
doi = {10.1038/s41598-024-66873-6}
}

@article{Gibson2023APM,
  title={A practical method to simulate realistic reduced-exposure CT images by the addition of computationally generated noise.},
  author={Nicholas Mark Gibson and Amy Lee and Martin Bencsik},
  journal={Radiological physics and technology},
  year={2023},
  url={https://api.semanticscholar.org/CorpusID:265148810}
}

@inproceedings{Unet,
author={Ronneberger, Olaf
and Fischer, Philipp
and Brox, Thomas},
title={U-Net: Convolutional Networks for Biomedical Image Segmentation},
booktitle={Medical Image Computing and Computer-Assisted Intervention -- MICCAI 2015},
year={2015},
publisher={Springer International Publishing},
address={Cham},
pages={234-241},
isbn={978-3-319-24574-4}
}

@article{pix2pix2017,
  title={Image-to-Image Translation with Conditional Adversarial Networks},
  author={Isola, Phillip and Zhu, Jun-Yan and Zhou, Tinghui and Efros, Alexei A},
  journal={CVPR},
  year={2017}
}

@InProceedings{sde,
  title = 	 {Image Restoration with Mean-Reverting Stochastic Differential Equations},
  author =       {Luo, Ziwei and Gustafsson, Fredrik K. and Zhao, Zheng and Sj\"{o}lund, Jens and Sch\"{o}n, Thomas B.},
  booktitle = 	 {Proceedings of the 40th International Conference on Machine Learning},
  pages = 	 {23045--23066},
  year = 	 {2023},
  volume = 	 {202},
  series = 	 {Proceedings of Machine Learning Research},
  month = 	 {23--29 Jul},
  publisher =    {PMLR},
  url = 	 {https://proceedings.mlr.press/v202/luo23b.html}
}

@inproceedings{CheXpert,
author = {Irvin, Jeremy and Rajpurkar, Pranav and Ko, Michael and Yu, Yifan and Ciurea-Ilcus, Silviana and Chute, Chris and Marklund, Henrik and Haghgoo, Behzad and Ball, Robyn and Shpanskaya, Katie and Seekins, Jayne and Mong, David A. and Halabi, Safwan S. and Sandberg, Jesse K. and Jones, Ricky and Larson, David B. and Langlotz, Curtis P. and Patel, Bhavik N. and Lungren, Matthew P. and Ng, Andrew Y.},
title = {CheXpert: a large chest radiograph dataset with uncertainty labels and expert comparison},
year = {2019},
isbn = {978-1-57735-809-1},
publisher = {AAAI Press},
url = {https://doi.org/10.1609/aaai.v33i01.3301590},
doi = {10.1609/aaai.v33i01.3301590},
booktitle = {Proceedings of the Thirty-Third AAAI Conference on Artificial Intelligence},
articleno = {73},
numpages = {8},
location = {Honolulu, Hawaii, USA},
series = {AAAI'19}
}

@article{Calabrese_2022,
   title={The University of California San Francisco Preoperative Diffuse                     Glioma MRI Dataset},
   volume={4},
   ISSN={2638-6100},
   url={http://dx.doi.org/10.1148/ryai.220058},
   DOI={10.1148/ryai.220058},
   number={6},
   journal={Radiology: Artificial Intelligence},
   publisher={Radiological Society of North America (RSNA)},
   author={Calabrese, Evan and Villanueva-Meyer, Javier E. and Rudie, Jeffrey D. and Rauschecker, Andreas M. and Baid, Ujjwal and Bakas, Spyridon and Cha, Soonmee and Mongan, John T. and Hess, Christopher P.},
   year={2022},
   month=nov }

@inproceedings{lpips,
  title={The Unreasonable Effectiveness of Deep Features as a Perceptual Metric},
  author={Zhang, Richard and Isola, Phillip and Efros, Alexei A and Shechtman, Eli and Wang, Oliver},
  booktitle={CVPR},
  year={2018}
}

@inproceedings{Cohen2021TorchXRayVisionAL,
  title={TorchXRayVision: A library of chest X-ray datasets and models},
  author={Joseph Paul Cohen and Joseph D. Viviano and Paul Bertin and Paul Morrison and Parsa Torabian and Matteo Guarrera and Matthew P. Lungren and Akshay Chaudhari and Rupert Brooks and Mohammad Hashir and Hadrien Bertrand},
  booktitle={International Conference on Medical Imaging with Deep Learning},
  year={2021},
  url={https://api.semanticscholar.org/CorpusID:240353861}
}

@inproceedings{DRO,
  author = {Sagawa, Shiori and Koh, Pang Wei and Hashimoto, Tatsunori B and Liang, Percy},
  booktitle = {International Conference on Learning Representations (ICLR)},
  title = {Distributionally robust neural networks for group shifts: On the importance of regularization for worst-case generalization},
  year = 2020
}

@InProceedings{JTT,
  title = 	 {Just Train Twice: Improving Group Robustness without Training Group Information},
  author =       {Liu, Evan Z and Haghgoo, Behzad and Chen, Annie S and Raghunathan, Aditi and Koh, Pang Wei and Sagawa, Shiori and Liang, Percy and Finn, Chelsea},
  booktitle = 	 {Proceedings of the 38th International Conference on Machine Learning},
  pages = 	 {6781--6792},
  year = 	 {2021},
  volume = 	 {139},
  series = 	 {Proceedings of Machine Learning Research},
  month = 	 {18--24 Jul},
  publisher =    {PMLR},
  url = 	 {https://proceedings.mlr.press/v139/liu21f.html}
}

@inproceedings{Creager2019FlexiblyFR,
  title={Flexibly Fair Representation Learning by Disentanglement},
  author={Elliot Creager and David Madras and J{\"o}rn-Henrik Jacobsen and Marissa A. Weis and Kevin Swersky and Toniann Pitassi and Richard S. Zemel},
  booktitle={International Conference on Machine Learning},
  year={2019},
  url={https://api.semanticscholar.org/CorpusID:174800294}
}

@inproceedings{Sarhan,
author = {Sarhan, Mhd Hasan and Navab, Nassir and Eslami, Abouzar and Albarqouni, Shadi},
title = {Fairness by Learning Orthogonal Disentangled Representations},
year = {2020},
isbn = {978-3-030-58525-9},
publisher = {Springer-Verlag},
address = {Berlin, Heidelberg},
url = {https://doi.org/10.1007/978-3-030-58526-6_44},
doi = {10.1007/978-3-030-58526-6_44},
booktitle = {Computer Vision – ECCV 2020: 16th European Conference, Glasgow, UK, August 23–28, 2020, Proceedings, Part XXIX},
pages = {746–761},
numpages = {16},
location = {Glasgow, United Kingdom}
}

@article{Gong,
  title={Mitigating Face Recognition Bias via Group Adaptive Classifier},
  author={Sixue Gong and Xiaoming Liu and Anil K. Jain},
  journal={2021 IEEE/CVF Conference on Computer Vision and Pattern Recognition (CVPR)},
  year={2020},
  pages={3413-3423},
  url={https://api.semanticscholar.org/CorpusID:219687431}
}

@inproceedings{Zhang,
author = {Zhang, Brian Hu and Lemoine, Blake and Mitchell, Margaret},
title = {Mitigating Unwanted Biases with Adversarial Learning},
year = {2018},
isbn = {9781450360128},
publisher = {Association for Computing Machinery},
address = {New York, NY, USA},
url = {https://doi.org/10.1145/3278721.3278779},
doi = {10.1145/3278721.3278779},
booktitle = {Proceedings of the 2018 AAAI/ACM Conference on AI, Ethics, and Society},
pages = {335–340},
numpages = {6},
location = {New Orleans, LA, USA},
series = {AIES '18}
}

@InProceedings{KimKimKimKimKim,
author = {Kim, Byungju and Kim, Hyunwoo and Kim, Kyungsu and Kim, Sungjin and Kim, Junmo},
title = {Learning Not to Learn: Training Deep Neural Networks With Biased Data},
booktitle = {The IEEE Conference on Computer Vision and Pattern Recognition (CVPR)},
month = {June},
year = {2019}
}

@INPROCEEDINGS {Wang,
author = { Wang, Zeyu and Qinami, Klint and Karakozis, Ioannis Christos and Genova, Kyle and Nair, Prem and Hata, Kenji and Russakovsky, Olga },
booktitle = { 2020 IEEE/CVF Conference on Computer Vision and Pattern Recognition (CVPR) },
title = {{ Towards Fairness in Visual Recognition: Effective Strategies for Bias Mitigation }},
year = {2020},
volume = {},
ISSN = {},
pages = {8916-8925},
doi = {10.1109/CVPR42600.2020.00894},
url = {https://doi.ieeecomputersociety.org/10.1109/CVPR42600.2020.00894},
publisher = {IEEE Computer Society},
address = {Los Alamitos, CA, USA},
month =Jun}

@INPROCEEDINGS{repair,
  author={Li, Yi and Vasconcelos, Nuno},
  booktitle={2019 IEEE/CVF Conference on Computer Vision and Pattern Recognition (CVPR)}, 
  title={REPAIR: Removing Representation Bias by Dataset Resampling}, 
  year={2019},
  volume={},
  number={},
  pages={9564-9573},
  doi={10.1109/CVPR.2019.00980}}

@inproceedings{WenlongOrtho,
author = {Deng, Wenlong and Zhong, Yuan and Dou, Qi and Li, Xiaoxiao},
title = {On Fairness of Medical Image Classification with Multiple Sensitive Attributes via Learning Orthogonal Representations},
year = {2023},
isbn = {978-3-031-34047-5},
publisher = {Springer-Verlag},
address = {Berlin, Heidelberg},
url = {https://doi.org/10.1007/978-3-031-34048-2_13},
doi = {10.1007/978-3-031-34048-2_13},
booktitle = {Information Processing in Medical Imaging: 28th International Conference, IPMI 2023, San Carlos de Bariloche, Argentina, June 18–23, 2023, Proceedings},
pages = {158–169},
numpages = {12},
location = {San Carlos de Bariloche, Argentina}
}

@InProceedings{FairDisCo,
author={Du, Siyi
and Hers, Ben
and Bayasi, Nourhan
and Hamarneh, Ghassan
and Garbi, Rafeef},
title={FairDisCo: Fairer AI in Dermatology via Disentanglement Contrastive Learning},
booktitle={Computer Vision -- ECCV 2022 Workshops},
year={2023},
publisher={Springer Nature Switzerland},
address={Cham},
pages={185-202},
isbn={978-3-031-25069-9}
}

@article{Adeli2019RepresentationLW,
  title={Representation Learning with Statistical Independence to Mitigate Bias},
  author={Ehsan Adeli and Qingyu Zhao and Adolf Pfefferbaum and Edith V. Sullivan and Li Fei-Fei and Juan Carlos Niebles and Kilian M. Pohl},
  journal={2021 IEEE Winter Conference on Applications of Computer Vision (WACV)},
  year={2019},
  pages={2512-2522},
  url={https://api.semanticscholar.org/CorpusID:211069024}
}

@Article{Singh2025.03.13.25323924,
author={Singh, Rohan
and Bapna, Monika
and Diab, Abdul Rahman
and Ruiz, Emily S.
and Lotter, William},
title={How AI is used in FDA-authorized medical devices: a taxonomy across 1,016 authorizations},
journal={npj Digital Medicine},
year={2025},
month={Jul},
day={01},
volume={8},
number={1},
pages={388},
issn={2398-6352},
doi={10.1038/s41746-025-01800-1},
url={https://doi.org/10.1038/s41746-025-01800-1}
}

@inproceedings{groh2021evaluating,
  title={Evaluating deep neural networks trained on clinical images in dermatology with the fitzpatrick 17k dataset},
  author={Groh, Matthew and Harris, Caleb and Soenksen, Luis and Lau, Felix and Han, Rachel and Kim, Aerin and Koochek, Arash and Badri, Omar},
  booktitle={Proceedings of the IEEE/CVF conference on computer vision and pattern recognition},
  pages={1820--1828},
  year={2021}
}

@Article{biomedinformatics4020050,
AUTHOR = {Chuah, Joshua and Yan, Pingkun and Wang, Ge and Hahn, Juergen},
TITLE = {Towards the Generation of Medical Imaging Classifiers Robust to Common Perturbations},
JOURNAL = {BioMedInformatics},
VOLUME = {4},
YEAR = {2024},
NUMBER = {2},
PAGES = {889--910},
URL = {https://www.mdpi.com/2673-7426/4/2/50},
ISSN = {2673-7426},
DOI = {10.3390/biomedinformatics4020050}
}

@article{bousse2024review,
 author = {Bousse, Alexandre and Kandarpa, Venkata Sai Sundar and Shi, Kuangyu and Gong, Kuang and Lee, Jae Sung and Liu, Chi and Visvikis, Dimitris},
 doi = {10.1109/TRPMS.2023.3349194},
 journal = {IEEE Transactions on Radiation and Plasma Medical Sciences},
 publisher = {IEEE},
 title = {A Review on Low-Dose Emission Tomography Post-Reconstruction Denoising With Neural Network Approaches},
 url = {https://arxiv.org/abs/2401.00232},
 year = {2024}
}

@Article{Heckel2024,
author={Heckel, Reinhard
and Jacob, Mathews
and Chaudhari, Akshay
and Perlman, Or
and Shimron, Efrat},
title={Deep learning for accelerated and robust MRI reconstruction},
journal={Magnetic Resonance Materials in Physics, Biology and Medicine},
year={2024},
month={Jul},
day={01},
volume={37},
number={3},
pages={335-368},
issn={1352-8661},
doi={10.1007/s10334-024-01173-8},
url={https://doi.org/10.1007/s10334-024-01173-8}
}

@article{AHISHAKIYE2021118,
title = {A survey on deep learning in medical image reconstruction},
journal = {Intelligent Medicine},
volume = {1},
number = {3},
pages = {118-127},
year = {2021},
issn = {2667-1026},
doi = {https://doi.org/10.1016/j.imed.2021.03.003},
url = {https://www.sciencedirect.com/science/article/pii/S2667102621000061},
author = {Emmanuel Ahishakiye and Martin Bastiaan {Van Gijzen} and Julius Tumwiine and Ruth Wario and Johnes Obungoloch}
}

@article{Kingma2014AdamAM,
  title={Adam: A Method for Stochastic Optimization},
  author={Diederik P. Kingma and Jimmy Ba},
  journal={CoRR},
  year={2014},
  volume={abs/1412.6980},
  url={https://api.semanticscholar.org/CorpusID:6628106}
}

@INPROCEEDINGS{Grad-CAM,
  author={Selvaraju, Ramprasaath R. and Cogswell, Michael and Das, Abhishek and Vedantam, Ramakrishna and Parikh, Devi and Batra, Dhruv},
  booktitle={2017 IEEE International Conference on Computer Vision (ICCV)}, 
  title={Grad-CAM: Visual Explanations from Deep Networks via Gradient-Based Localization}, 
  year={2017},
  volume={},
  number={},
  pages={618-626},
  doi={10.1109/ICCV.2017.74}}

\clearpage
\setcounter{secnumdepth}{1}
\appendix
\section{  }


\begin{figure}[ht]
    \centering
    \includegraphics[width=\linewidth]{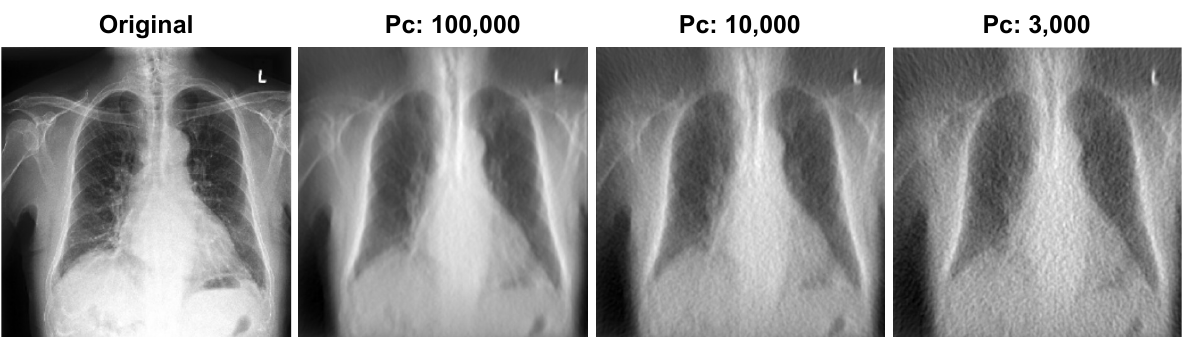}
    \caption{X-Ray images with photon count 100,000, 10,000, 3,000.}
    \label{fig:noise-chex}
\end{figure}
\begin{figure}[ht]
    \centering
    \includegraphics[width=\linewidth]{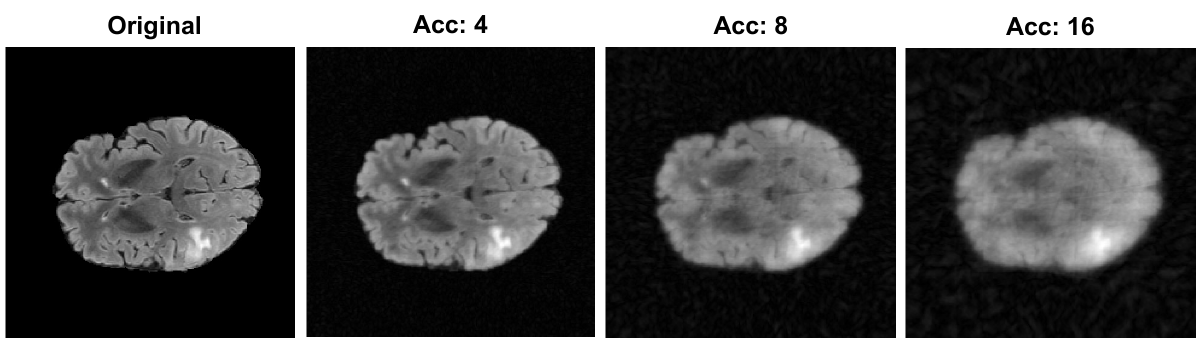}
    \caption{MRI images with acceleration 4, 8, 16.}
    \label{fig:noise-ucsf}
\end{figure}

\begin{figure}[ht]
    \centering
    \includegraphics[width=\linewidth]{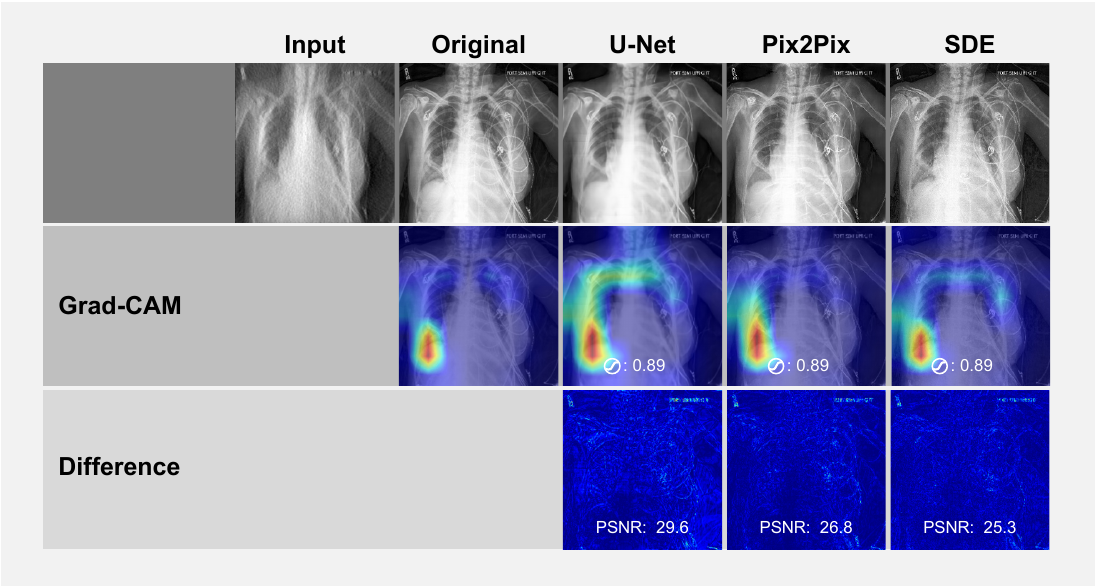}
    \caption{Reconstruction example from photon count 10,000 for the different models. Grad-CAM \cite{Grad-CAM} and logit score correspond to the lung lesion prediction of the pre-trained classifier, indicating similar predictions on the reconstructed images.}
    \label{fig:chex-example}
\end{figure}
\begin{figure}[ht]
    \centering
    \includegraphics[width=\linewidth]{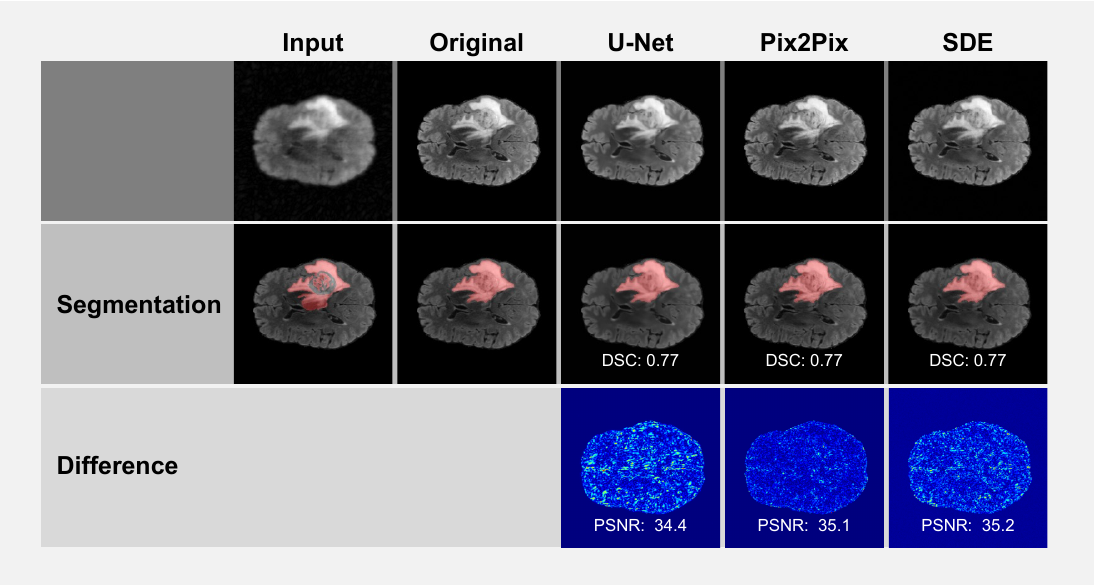}
    \caption{Reconstruction with corresponding segmentation and Dice score of an MRI image with acceleration 8 for the different models.}
    \label{fig:ucsf-example}
\end{figure}

\paragraph{Diagnostic Hyperparameters.}\label{app:hyperparams}
The segmentation network was optimized with Adam \cite{Kingma2014AdamAM} using a learning rate of 0.001 and a batch size of 8 without data augmentation for 20 training epochs. The training loss consisted of Dice and L1, equally weighted at 0.5 each.
The network used a sigmoid activation and a threshold of 0.5 was used at inference to compute the Dice performance.
The model was trained on a per-slice level using all available MRI slices.
At inference, Dice performance was computed using slices 60-130, as this range is representative of the regions where the ground truth masks appear and thus is more representative of performance. The Dice scores were computed separately for each slice, then averaged across slices per patient, followed by averaging across patients to compute final performance. 
For the UCSF‐PDGM ResNet classifiers, we trained for 20 epochs with a learning rate of 0.0001 and a batch size of 16 without augmentation.
Each task was treated as binary classification (subtype: glioblastoma vs not glioblastoma, grade: (II, III)  vs IV) using binary cross entropy loss.
All MRI slices were again used for training, followed by using slices 60-130 at inference. Prediction scores were generated separately for each slice, followed by computing a patient-level score as the median across slices to serve as input to patient-level AUROC calculations. The median across slices was used to improve robustness to outliers. All UCSF-PDGM diagnostic models were trained using images pre-processed using min-max normalization to the 0-1 range and resized to 256x256.
The CheXpert DenseNet classifier was trained using TorchXRayVision \cite{Cohen2021TorchXRayVisionAL}. 
The default image preprocessing was used, with an input size of 224x224 pixels and normalization to a range of -1024 to 1024. The model was trained without data augmentation for 50 epochs using the Adam optimizer with a learning rate of 1e-3 and a weight decay of 1e-5.

\paragraph{Reconstruction Hyperparameters.}
No data augmentation was applied to any of the reconstruction pipelines.
A U-Net was trained for 20 epochs on both UCSF-PDGM and CheXpert, using Adam with MSE loss, a learning rate of 0.001, and a batch size of 16.
The GAN (Pix2Pix) was trained for 200 epochs on each dataset with Adam, a learning rate 0.0002, and a batch size of 32 to compensate for the smaller data volume.
For the SDE model, we employed Adam with a learning rate of 0.0001, a cosine learning-rate schedule, and a batch size of 8; training ran for 40 epochs on CheXpert and 300 epochs on UCSF-PDGM.
We note that the number of epochs varied between models because the different approaches take longer to converge (e.g., GANs are inherently less stable than a standard MSE loss), but in each case, the final weights were selected via validation loss monitoring, consistent with standard practice.
During mitigation with the EODD-constraint, we employed \(\tau=0.5\) for the threshold, \(T=0.3\) for the temperature, and a momentum value of 0.1 for the EMA.
The remaining hyperparameters and architectural details were adopted unchanged from the original U-Net \cite{Unet}, Pix2Pix \cite{pix2pix2017}, and SDE \cite{sde} publications. Image pre-processing consisted of min-max normalization to the 0-1 range and resizing to 256x256 for all reconstruction models.

The models were trained on a single NVIDIA A40 or A100 GPU. The SDE model was computationally most expensive and needed a maximum of 48 hours to train from scratch. For all models, the final weights were chosen based on performance on the validation split during training.

\paragraph{MRI preprocessing and reconstruction details.}\label{app:mri}
UCSF-PDGM provides reconstructed single-channel images (no multi-coil raw k-space data or complex-valued images). As a result, no coil combination, coil compression, or sensitivity map estimation was performed. To simulate undersampled MRI acquisitions, reconstructed images were retrospectively transformed to synthetic k-space using a discrete Fourier transform, implicitly assuming zero phase. Radial undersampling masks \cite{FengRadial} were applied in k-space, and zero-filled reconstructions were obtained via inverse Fourier transform. The models were then trained in an image-to-image fashion to map the zero-filled images to the original reconstructed images. This pipeline was chosen to align with standard practice in MRI reconstruction studies when raw k-space is unavailable. The original UCSF-PDGM dataset was acquired using a 3.0 tesla scanner and a dedicated 8-channel head coil \cite{Calabrese_2022}. Two gadolinium-based contrast agents were used across the cohort: gadobutrol at a dose of 0.1 mL/kg and gadoterate at a dose of 0.2 mL/kg.

\paragraph{Proof of Proportionality.}\label{app:proof}
When the protected attribute A takes more than two categories (e.g., multiple races, genders, or age groups), we compare all pairs $a_i, a_j$ of subgroups. Then, we take the maximum of the pairwise disparities in true positive and false positive rates:

\begin{align*}
EODD &= \max_{1 \le i < j \le k}
\Bigl[
\;\bigl|P(\hat{Y}=1 \mid Y=1, A=a_i) \\
   \;&-\; P(\hat{Y}=1 \mid Y=1, A=a_j)\bigr| 
\\
\quad &+\;\bigl|P(\hat{Y}=1 \mid Y=0, A=a_i)\\ 
   \;&-\; P(\hat{Y}=1 \mid Y=0, A=a_j)\bigr| 
\Bigr]
\end{align*}

Each pairwise comparison is handled exactly as in the binary case by treating $a_i, a_j$ as $0,1$. Therefore, all the steps below—derived under a binary setup—apply pairwise to any two subgroups. Taking the maximum over these pairwise disparities then yields the multi-group measure.

\noindent This proof is based on the derivation by \cite{marcinkevičs2022debiasingdeepchestxray}, and adjusted for EODD.

\noindent EODD measures the disparity between subgroups in true positive rate (TPR) and false positive rate (FPR). In the binary case: 
\begin{align*}
    EODD &= P_{X,Y, A} (\hat{Y} = 1 | Y = 1, A=1)\\ &- P_{X,Y|A} (\hat{Y} = 1 | Y = 1, A=0) \\
    &+ P_{X,Y, A} (\hat{Y} = 1 | Y = 0, A=1)\\ &- P_{X,Y, A} (\hat{Y} = 1 | Y = 0, A=0)
\end{align*}
\noindent This can be expressed by the following proxy function. 
\begin{align}
    EODD &= \frac{\sum_{i=1}^{n} f_{\theta}(x_i) a_i y_i}{\sum_{i=1}^{n} a_i y_i} \\
    &- \frac{\sum_{i=1}^{n} f_{\theta}(x_i)(1-a_i)y_i}{\sum_{i=1}^{n} (1-a_i)y_i} \tag{1}\\
    &\quad + \frac{\sum_{i=1}^{n} f_{\theta}(x_i) a_i (1-y_i)}{\sum_{i=1}^{n} a_i (1-y_i)} \\
    &- \frac{\sum_{i=1}^{n} f_{\theta}(x_i) (1-a_i)(1-y_i)}{\sum_{i=1}^{n} (1-a_i)(1-y_i)} \tag{2}
\end{align}

\noindent To start, let's define the conditional covariance:
\begin{align}
    \text{cov}&(A, X | Y = y) =\\ &\mathbb{E}[(A - \mathbb{E}[A | Y = y]) (X - \mathbb{E}[X | Y = y]) | Y = y] \notag \\ 
    &= \mathbb{E}[AX | Y = y] - \mathbb{E}[A | Y = y] \mathbb{E}[X | Y = y] \tag{3}
\end{align}

\noindent We can use the law of total covariance to prove the validity:

\begin{align}
    \text{cov}(A, X) &= \mathbb{E} \Big[ \text{cov}(A, X | Y) \Big] \\ &+ \text{cov} \Big( \mathbb{E}[A | Y], \mathbb{E}[X | Y] \Big) \tag{4}
\end{align}

\noindent Expanding the first expectation term with (3):
\begin{align}
    \mathbb{E} [\text{cov}(A, X | Y)] &= \mathbb{E} \Big[ \mathbb{E}[AX | Y] - \mathbb{E}[A | Y] \mathbb{E}[X | Y] \Big] \notag \\
    &= \mathbb{E}[AX] - \mathbb{E}[\mathbb{E}[A | Y] \mathbb{E}[X | Y]] \tag{5}
\end{align}

\noindent Expanding the second covariance term:

\begin{align}
    \text{cov}(\mathbb{E}[X | Z], \mathbb{E}[Y | Z]) &= \mathbb{E}[\mathbb{E}[X | Z] \mathbb{E}[Y | Z]] \tag{6} \\& - \mathbb{E}[X] \mathbb{E}[Y] \tag{6}
\end{align}

\noindent Substituting (5) and (6) into (4):

\begin{align*}
    \text{cov}(X, Y) &= \mathbb{E}[XY] - \mathbb{E}[\mathbb{E}[X | Z] \mathbb{E}[Y | Z]] \\ &+ \mathbb{E}[\mathbb{E}[X | Z] \mathbb{E}[Y | Z]] - \mathbb{E}[X] \mathbb{E}[Y] \\
    &= \mathbb{E}[XY] - \mathbb{E}[X] \mathbb{E}[Y] \\
    &= \text{cov}(X, Y)
\end{align*}
\noindent We want to show that $\Delta_{OOD} \propto \widehat{\text{Cov}}(A, f_{\theta}(X) | Y=1) +  
\widehat{\text{Cov}}(A, f_{\theta}(X) | Y=0)$

\noindent Let $\sum_{i} a_i y_i = S_{AY}, \quad \sum_{i} a_i = S_A, \quad \sum_{i} y_i = S_Y.$ \\

\noindent \textbf{Expanding EODD}:

\noindent Expanding (1):

\begin{align*}
    &\frac{\sum_{i=1}^{N} f_{\theta}(x_i) a_i y_i}{\sum_{i=1}^{N} a_i y_i}
    - \frac{\sum_{i=1}^{N} f_{\theta}(x_i) (1 - a_i) y_i}{\sum_{i=1}^{N} y_i (1 - a_i) y_i} \\
    &= \frac{1}{S_{AY}} \sum_{i=1}^{N} f_{\theta}(x_i) a_i y_i
    - \frac{1}{S_Y - S_A} \sum_{i=1}^{N} f_{\theta}(x_i)  
    \\ &+ \frac{1}{S_Y - S_{AY}} \sum_{i=1}^{N} f_{\theta}(x_i) a_i y_i  \\
    &= \frac{S_Y}{S_{AY} (S_Y - S_{AY})} 
    \sum_{i=1}^{N} f_{\theta}(x_i) y_i a_i
    \\ &- \frac{1}{S_Y - S_{AY}} \sum_{i=1}^{N} f_{\theta}(x_i) y_i 
\end{align*}

\noindent Note that: 
\begin{align*}
    \widehat{\text{Cov}}&(A, f_{\theta}(X) | Y=1)
    \\ &= \frac{\sum_{i=1}^{n} f_{\theta}(x_i) a_i y_i}{\sum_{i=1}^{n} y_i} 
    \\ &- \frac{\sum_{i=1}^{n} a_i y_i}{\sum_{i=1}^{n} y_i}
    \frac{\sum_{i=1}^{n} f_{\theta}(x_i) y_i}{\sum_{i=1}^{n} y_i}  \\
    &= \frac{1}{S_Y} \sum_{i=1}^{n} f_{\theta}(x_i) a_i y_i 
    \\ &- \frac{S_{AY}}{S_Y^2} \sum_{i=1}^{n} f_{\theta}(x_i) y_i.
\end{align*}
Showing $(5) \propto \widehat{\text{Cov}}(A, f_{\theta}(X) | Y=1)\\ \quad
\text{with factor} \quad \frac{S_Y^2}{S_{AY} (S_Y - S_{AY})}$, independent of $f_{\theta}$.\\

\noindent Expanding (2):

\begin{align*}
    &\frac{\sum_{i=1}^{n} f_{\theta}(x_i) a_i (1-y_i)}{\sum_{i=1}^{n} a_i (1-y_i)} 
    \\ &- \frac{\sum_{i=1}^{n} f_{\theta}(x_i) (1-a_i)(1-y_i)}{\sum_{i=1}^{n} (1-a_i)(1-y_i)} \\
    &= \frac{N - S_Y}{(N - S_Y - S_A + S_{AY})(S_A - S_{AY})} 
    \sum_{i=1}^{N} f_{\theta}(x_i) a_i  \\
    &\quad - \frac{N - S_Y}{(N - S_Y - S_A + S_{AY})(S_A - S_{AY})} 
    \sum_{i=1}^{N} f_{\theta}(x_i) a_i y_i  \\
    &\quad - \frac{1}{N - S_Y - S_A + S_{AY}} 
    \sum_{i=1}^{N} f_{\theta}(x_i) y_i  \\
    &\quad - \frac{N}{N - S_Y - S_A + S_{AY}} 
    \sum_{i=1}^{N} f_{\theta}(x_i)
\end{align*}

\noindent Similarly: 
\begin{align*}
    \widehat{\text{Cov}}&(A, f_0(X) | Y = 0) \\ &=  
    \frac{\sum_{i=1}^{N} f_0(x_i) a_i (1 - y_i)}{\sum_{i=1}^{N} (1 - y_i)}
    \\ &- \frac{\sum_{i=1}^{N} a_i (1 - y_i)}{\sum_{i=1}^{N} (1 - y_i)}
    \cdot \frac{\sum_{i=1}^{N} f_0(x_i) (1 - y_i)}{\sum_{i=1}^{N} (1 - y_i)}
     \\
    &= \frac{1}{N - S_Y} \sum_{i=1}^{N} f_0(x_i) a_i
    \\&- \frac{N}{N - S_Y} \sum_{i=1}^{N} f_0(x_i) a_i y_i
    \notag \\
    &\quad - \frac{S_A - S_{AY}}{(N - S_Y)^2} \sum_{i=1}^{N} f_0(x_i)
    \\ &- \frac{S_A \cdot S_{AY}}{(N - S_Y)^2} \sum_{i=1}^{N} f_0(x_i) y_i
\end{align*}
Showing $(6) \propto \widehat{\text{Cov}}(A, f_{\theta}(X) | Y=0) \quad 
\text{with factor} \quad 
\frac{(S_A - S_{AY}) (N - S_Y - S_A + S_{AY})}{(N - S_Y)^2}$, independent of $f_{\theta}$.\\

\noindent Therefore, $EODD \propto \widehat{\text{Cov}}(A, f_{\theta}(X) | Y=1) +  
\widehat{\text{Cov}}(A, f_{\theta}(X) | Y=0)$.

\clearpage

\begin{table*}[h]
\centering
\begin{tabular}{l|cccccc|c}
\hline
 & \textbf{AI/AN} & \textbf{Asian} & \textbf{Black} & \textbf{NH/PI} & \textbf{Other} & \textbf{White} &  \\
\hline
\textbf{Female, $> 62$} & 54  & 1539  & 923  & 314  & 2518  & 6456  & 11804 \\
\textbf{Female, $\leq 62$} & 39  & 1739  & 608  & 136  & 1710  & 9500  & 13732 \\
\textbf{Male, $> 62$} & 56  & 1734  & 1023  & 240  & 3553  & 8984  & 15590 \\
\textbf{Male, $\leq 62$} & 27  & 1924  & 539  & 171  & 1853  & 11170  & 15684 \\
\hline
\textbf{} & 176  & 6936  & 3093  & 861  & 9634  & 36110  & 56810 \\
\hline
\end{tabular}
\caption{Patient-wise groups used for analysis based on sex, age, and race for the CheXpert dataset. Unequally distributed with very few samples for American Indian or Alaska Native (AI/AN) and Native Hawaiian or Other Pacific Islander (NH/PI).}\label{tab:chex_dataset}
\end{table*}

\begin{table*}[h]
\centering
\begin{tabular}{l|cc|c}
\hline
 & \textbf{Male} & \textbf{Female} & \\
\hline
\textbf{$\leq 58$} & 155 & 92 & 147\\
\textbf{$> 58$} & 144 & 110 & 254\\
\hline
\textbf{} & 299 & 202 & 501\\
\hline
\end{tabular}
\caption{Patient distribution by sex and age for the UCSF-PDGM dataset. Patients under 58 and females represent minority groups.}\label{tab:ucsf_dataset}
\end{table*}


\begin{figure}[h]
    \centering
    \includegraphics[width=0.6\linewidth]{plots/evaluation_performance/ucsf/ucsf-evaluation_performance_legend.pdf}
    \vspace{1em}

    \includegraphics[width=0.6\linewidth]{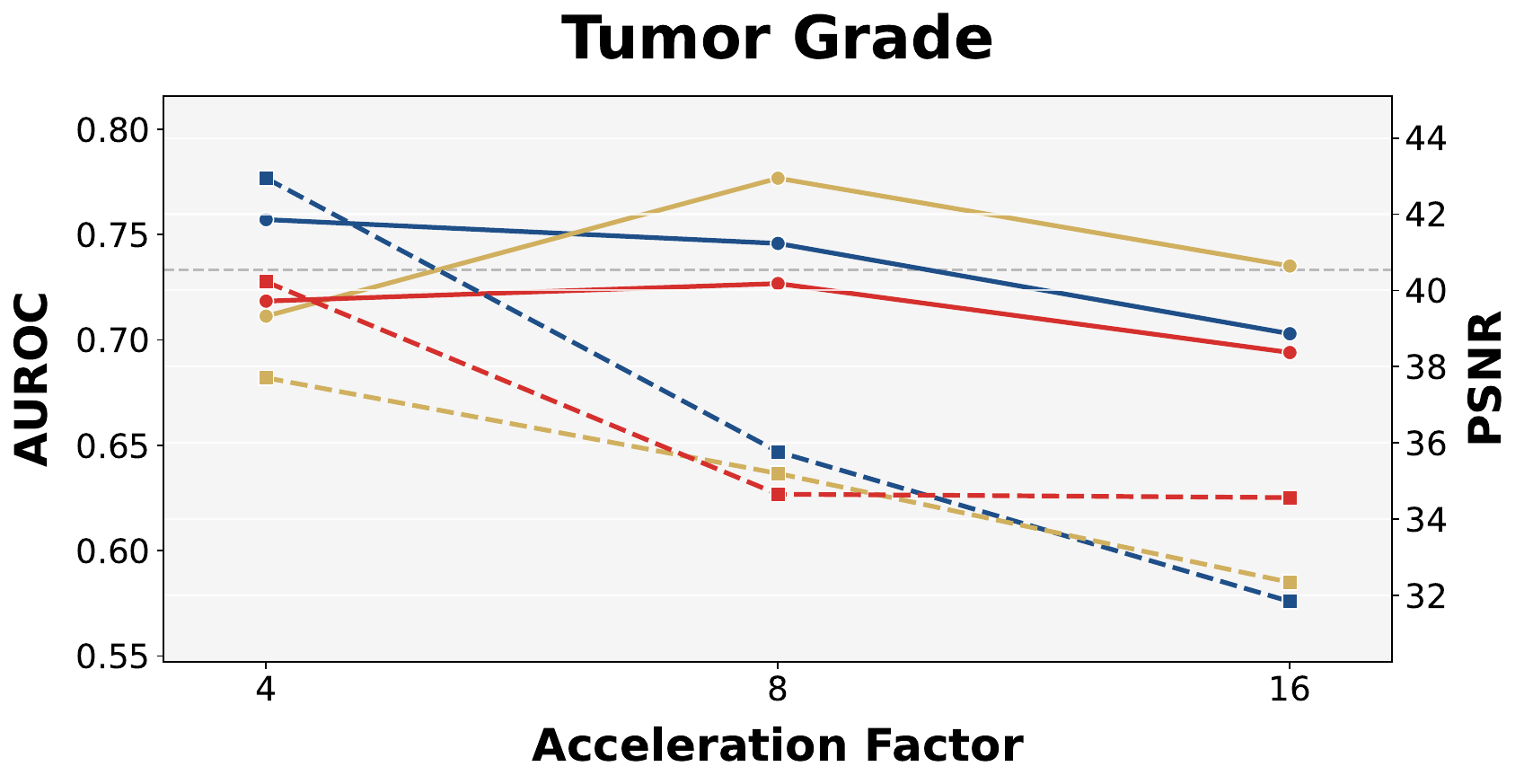}
    \vspace{1em} 

    \includegraphics[width=0.6\linewidth]{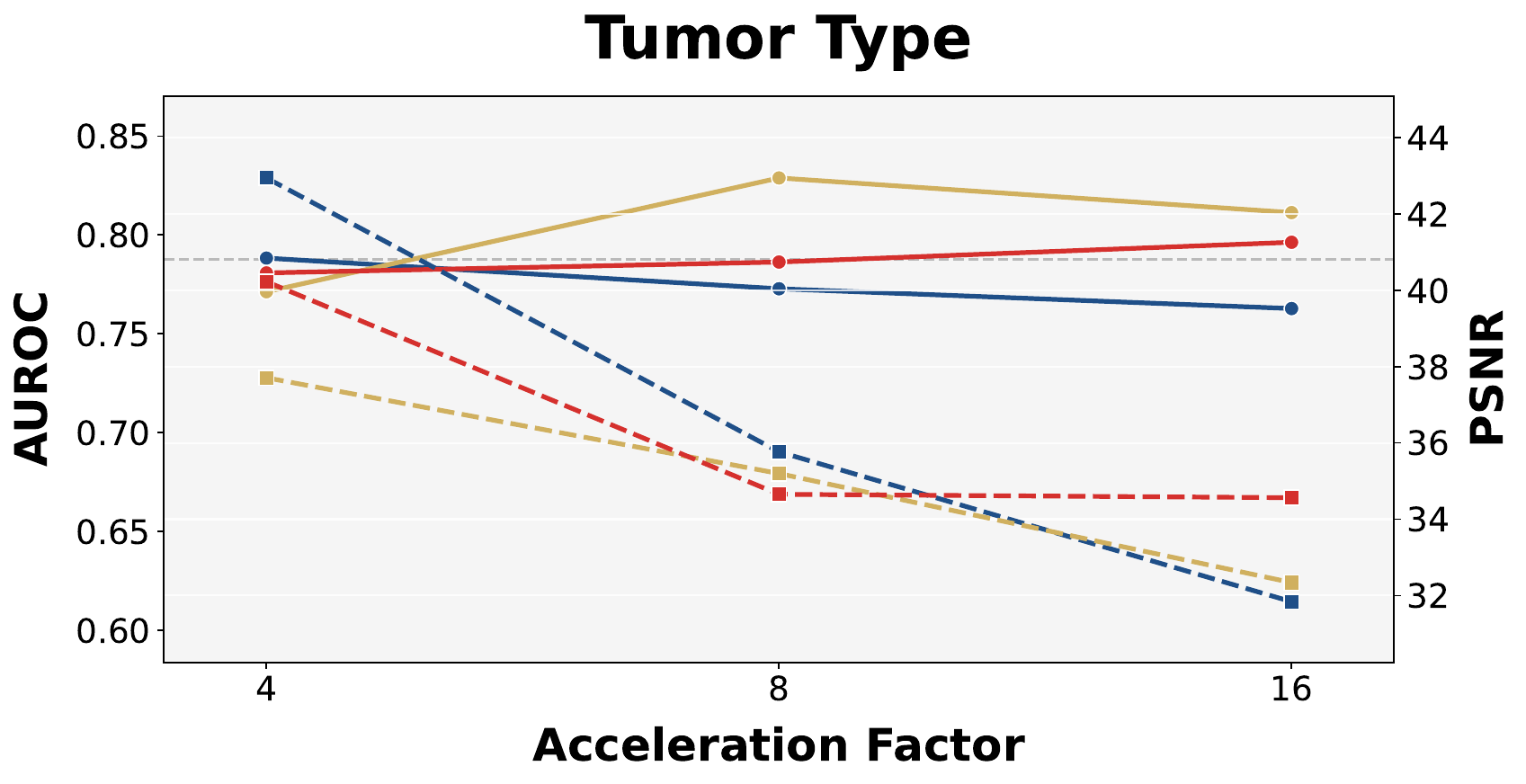}

    \caption{Tumor Type and Tumor Grade and PSNR values for different noise levels on UCSF-PDGM. The image quality and diagnostic performance axes are on a similar percentage scale. Task performance metrics show high stability across models and noise conditions, while PSNR drops with increasing noise.}
    \label{fig:performance_ucsf}
\end{figure}

\begin{table*}[t]
\centering
\footnotesize
\begin{tabular}{cll|llll}
\hline
\multicolumn{1}{l}{\textbf{Photon Count}} & \multicolumn{2}{c|}{\textbf{Metrics}}         & \textbf{Baseline} & \textbf{U-Net} & \textbf{GAN} & \textbf{SDE} \\ \hline
\multirow{15}{*}{100,000}               & \multirow{13}{*}{\textbf{AUROC}} & \textbf{Atalectasis}  & 0.87 & 0.87 & 0.86 & 0.87 \\
                                 &                        & \textbf{Cardiomegaly} & 0.91 & 0.91 & 0.91 & 0.91 \\
                                 &                        & \textbf{Consolidation} & 0.91 & 0.91 & 0.91 & 0.91 \\ 
                                 &                        & \textbf{Edema} & 0.90 & 0.90 & 0.90 & 0.90 \\ 
                                 &                        & \textbf{EC} & 0.79 & 0.78 & 0.78 & 0.79 \\ 
                                 &                        & \textbf{Fracture} & 0.76 & 0.75 & 0.75 & 0.76 \\ 
                                 &                        & \textbf{Lung Lesion} & 0.80 & 0.79 & 0.79 & 0.79 \\ 
                                 &                        & \textbf{Lung Opacity} & 0.88 & 0.88 & 0.88 & 0.88 \\ 
                                 &                        & \textbf{Pleural Effusion} & 0.93 & 0.92 & 0.92 & 0.92 \\ 
                                 &                        & \textbf{Pleural Other} & 0.83 & 0.82 & 0.81 & 0.82 \\ 
                                 &                        & \textbf{Pneumonia} & 0.83 & 0.83 & 0.83 & 0.83 \\ 
                                 &                        & \textbf{Pneumothorax} & 0.77 & 0.75 & 0.76 & 0.77 \\
                                 &                        & \textbf{Average} & 0.85 & 0.84 & 0.84 & 0.85 \\ \cline{2-3}
                                 & \multicolumn{2}{l|}{\textbf{PSNR}}            &  & 31.60 & 30.16 & 29.98 \\
                                 & \multicolumn{2}{l|}{\textbf{LPIPS}}           &  & 0.13 & 0.08 & 0.08 \\ \hline
\multirow{15}{*}{10,000}               & \multirow{13}{*}{\textbf{AUROC}} & \textbf{Atalectasis}  & 0.87 & 0.87 & 0.86 & 0.87 \\
                                 &                        & \textbf{Cardiomegaly} & 0.91 & 0.90 & 0.90 & 0.91 \\
                                 &                        & \textbf{Consolidation} & 0.91 & 0.91 & 0.90 & 0.91 \\ 
                                 &                        & \textbf{Edema} & 0.90 & 0.89 & 0.89 & 0.90 \\ 
                                 &                        & \textbf{EC} & 0.79 & 0.78 & 0.78 & 0.78 \\ 
                                 &                        & \textbf{Fracture} & 0.76 & 0.75 & 0.74 & 0.75 \\ 
                                 &                        & \textbf{Lung Lesion} & 0.80 & 0.78 & 0.78 & 0.79 \\ 
                                 &                        & \textbf{Lung Opacity} & 0.88 & 0.88 & 0.87 & 0.88 \\ 
                                 &                        & \textbf{Pleural Effusion} & 0.93 & 0.92 & 0.91 & 0.92 \\ 
                                 &                        & \textbf{Pleural Other} & 0.83 & 0.81 & 0.80 & 0.82 \\ 
                                 &                        & \textbf{Pneumonia} & 0.83 & 0.82 & 0.82 & 0.82 \\ 
                                 &                        & \textbf{Pneumothorax} & 0.77 & 0.75 & 0.75 & 0.77 \\
                                 &                        & \textbf{Average} & 0.85 & 0.84 & 0.83 & 0.84 \\ \cline{2-3}
                                 & \multicolumn{2}{l|}{\textbf{PSNR}}            &  & 30.52 & 28.62 & 27.12 \\
                                 & \multicolumn{2}{l|}{\textbf{LPIPS}}           &  & 0.19 & 0.11 & 0.15 \\ \hline
\multirow{15}{*}{3000}               & \multirow{13}{*}{\textbf{AUROC}} & \textbf{Atalectasis}  & 0.87 & 0.86 & 0.85 & 0.86 \\
                                 &                        & \textbf{Cardiomegaly} & 0.91 & 0.90 & 0.90 & 0.91 \\
                                 &                        & \textbf{Consolidation} & 0.91 & 0.91 & 0.90 & 0.90 \\ 
                                 &                        & \textbf{Edema} & 0.90 & 0.89 & 0.89 & 0.89 \\ 
                                 &                        & \textbf{EC} & 0.79 & 0.78 & 0.78 & 0.78 \\ 
                                 &                        & \textbf{Fracture} & 0.76 & 0.74 & 0.73 & 0.75 \\ 
                                 &                        & \textbf{Lung Lesion} & 0.80 & 0.77 & 0.77 & 0.78 \\ 
                                 &                        & \textbf{Lung Opacity} & 0.88 & 0.87 & 0.87 & 0.87 \\ 
                                 &                        & \textbf{Pleural Effusion} & 0.93 & 0.91 & 0.91 & 0.92 \\ 
                                 &                        & \textbf{Pleural Other} & 0.83 & 0.80 & 0.78 & 0.81 \\ 
                                 &                        & \textbf{Pneumonia} & 0.83 & 0.82 & 0.80 & 0.82 \\ 
                                 &                        & \textbf{Pneumothorax} & 0.77 & 0.74 & 0.74 & 0.77 \\
                                 &                        & \textbf{Average} & 0.85 & 0.83 & 0.83 & 0.84 \\ \cline{2-3}
                                 & \multicolumn{2}{l|}{\textbf{PSNR}}            &  & 28.89 & 27.36 & 26.83 \\
                                 & \multicolumn{2}{l|}{\textbf{LPIPS}}           &  & 0.22 & 0.14 & 0.15 \\ \hline
\end{tabular}
\vspace{-10pt}
\caption{CheXpert performance across reconstruction models and photon counts.
Pathologies with lower baseline AUROC (e.g., fracture, pneumothorax, lung lesion) experience greater performance drops under noise compared to more easily detectable conditions (e.g., effusion, cardiomegaly). Baseline corresponds to original images.}\label{tab:chex_perf}
\end{table*}

\begin{table*}[t]
    \centering
    \begin{tabular}{cll|cccc}
    \hline
    \multicolumn{1}{l}{\textbf{Acceleration}} & \multicolumn{2}{c|}{\textbf{Metrics}}         & \textbf{Baseline} & \textbf{U-Net} & \textbf{GAN} & \textbf{SDE} \\ \hline
    \multirow{5}{*}{4}               & \multirow{2}{*}{\textbf{AUROC}} & \textbf{Tumor Type}  &    0.79      &   0.79   &  0.77   &     0.78     \\
                                     &                        & \textbf{Tumor Grade} &   0.73       &   0.76   &  0.71   &     0.72     \\ \cline{2-3}
                                     & \multicolumn{2}{l|}{\textbf{Dice}}            &  0.72        &   0.72   &  0.71   &     0.72     \\ \cline{2-3}
                                     & \multicolumn{2}{l|}{\textbf{PSNR}}            &         &   42.94   &  37.71   &     40.23     \\
                                     & \multicolumn{2}{l|}{\textbf{LPIPS}}           &          &   0.01   &  0.02   &     0.00     \\ \hline
    \multirow{5}{*}{8}               & \multirow{2}{*}{\textbf{AUROC}} & \textbf{Tumor Type}  &   0.79       &   0.77   &  0.83   &     0.79     \\
                                     &                        & \textbf{Tumor Grade} &   0.73       &   0.75   &  0.78   &     0.73     \\ \cline{2-3}
                                     & \multicolumn{2}{l|}{\textbf{Dice}}            &  0.72        &   0.70   &  0.71   &     0.71     \\ \cline{2-3}
                                     & \multicolumn{2}{l|}{\textbf{PSNR}}            &         &   35.77   &  35.20   &     34.65     \\
                                     & \multicolumn{2}{l|}{\textbf{LPIPS}}           &          &   0.03   &  0.02   &     0.02     \\ \hline
    \multirow{5}{*}{16}               & \multirow{2}{*}{\textbf{AUROC}} & \textbf{Tumor Type}  &   0.79       &   0.76   &  0.81   &     0.80     \\
                                     &                        & \textbf{Tumor Grade} &   0.73       &   0.70   &  0.74   &     0.69     \\ \cline{2-3}
                                     & \multicolumn{2}{l|}{\textbf{Dice}}            &  0.72        &   0.67   &  0.70   &     0.71     \\ \cline{2-3}
                                     & \multicolumn{2}{l|}{\textbf{PSNR}}            &         &   31.84   &  32.34   &     34.56     \\
                                     & \multicolumn{2}{l|}{\textbf{LPIPS}}           &          &   0.06   &  0.04   &     0.02     \\ \hline
    \end{tabular}
    \caption{Performance metrics for UCSF-PDGM across reconstruction models and noise levels. While PSNR varies with noise and model, downstream segmentation and classification metrics remain relatively stable, indicating robust task performance across conditions.}\label{tab:ucsf_perf}
    \end{table*}



\begin{figure*}[t]
\centering

\subfigure[]{%
\begin{minipage}[t]{\textwidth}
\centering
\includegraphics[width=0.3\textwidth]{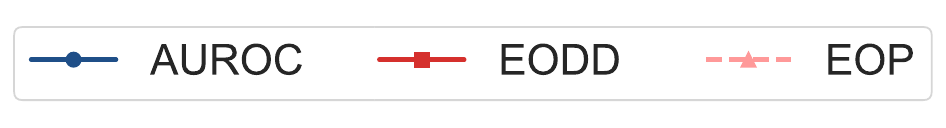}
\end{minipage}}

\begin{tabular}{c@{\hspace{0.2cm}}c@{\hspace{0.2cm}}c}

\subfigure[Age]{%
\begin{minipage}[t]{0.3\textwidth}
\centering
\includegraphics[width=\linewidth]{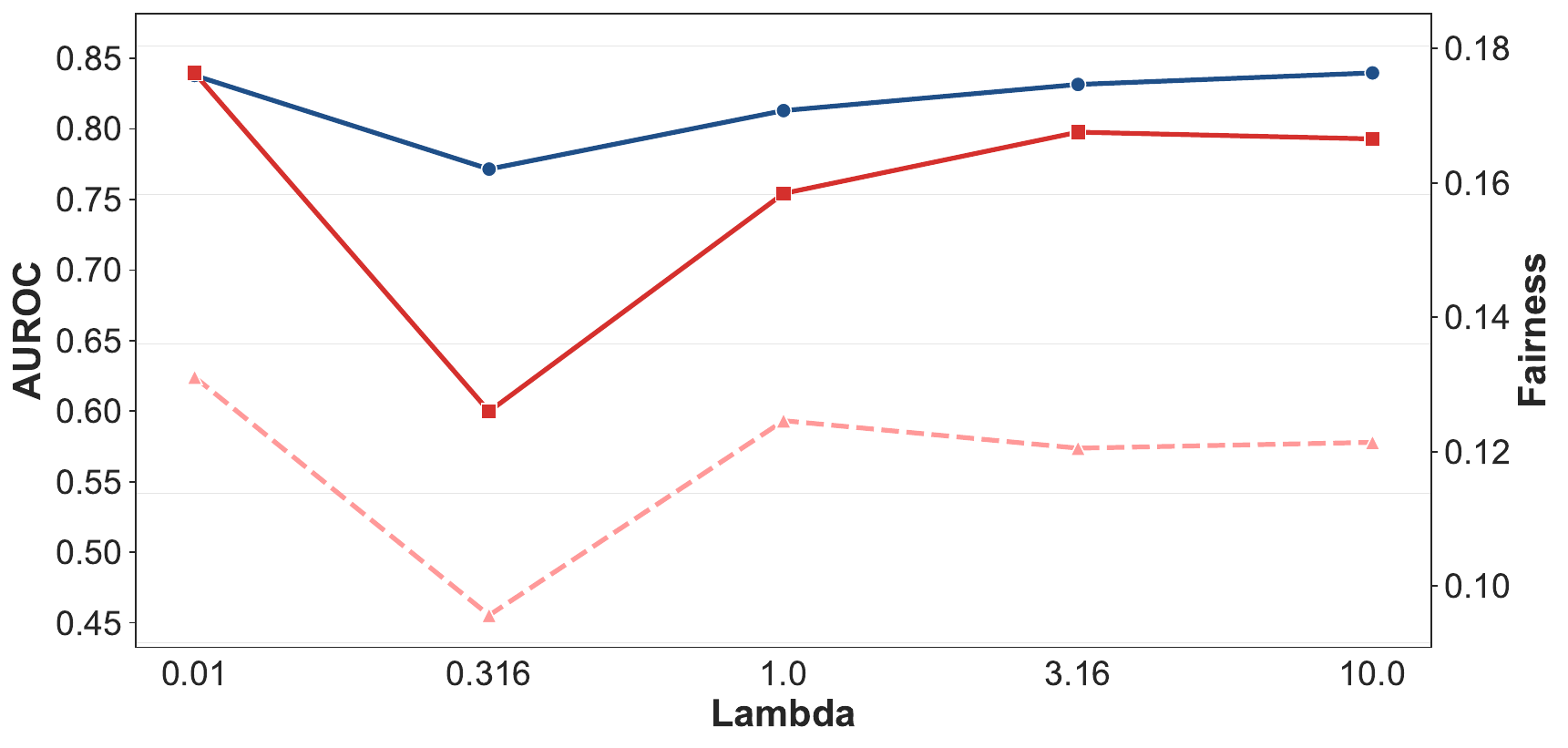}
\end{minipage}}
&
\subfigure[Sex]{%
\begin{minipage}[t]{0.3\textwidth}
\centering
\includegraphics[width=\linewidth]{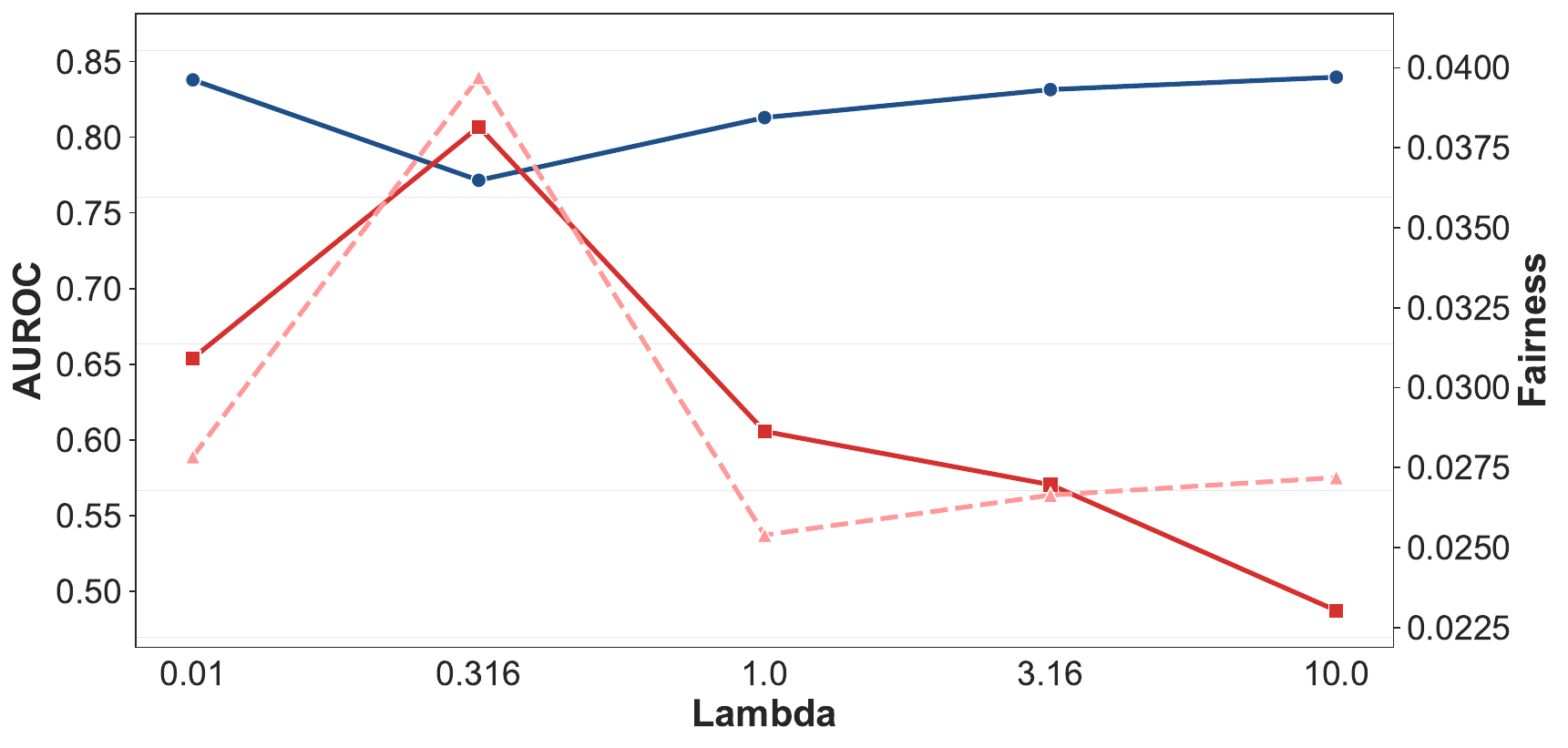}
\end{minipage}}
&
\subfigure[Race]{%
\begin{minipage}[t]{0.3\textwidth}
\centering
\includegraphics[width=\linewidth]{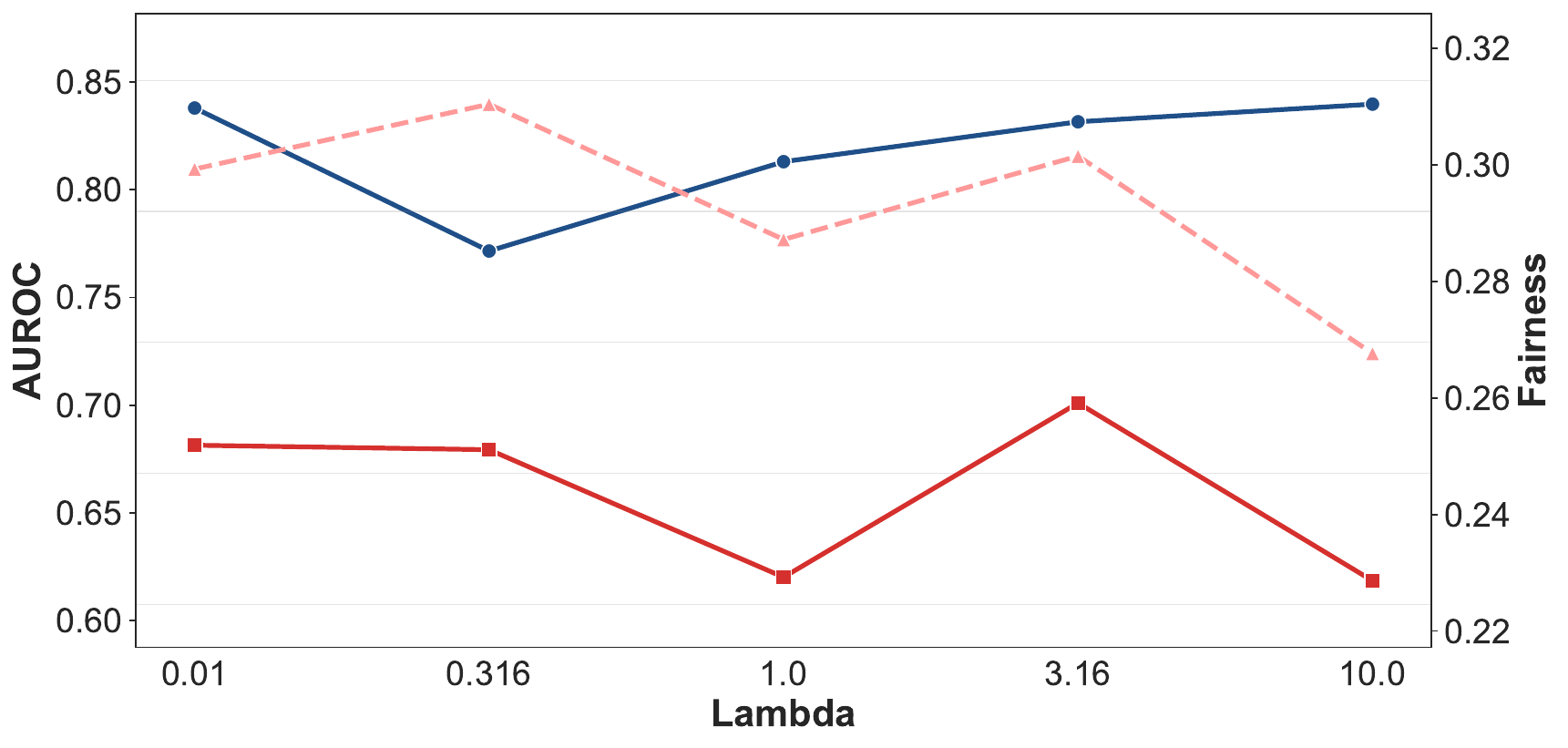}
\end{minipage}}

\end{tabular}

    \caption{Influence of fairness weighting parameter ($\lambda_{\mathrm{fair}}$) on classifier AUROC performance and fairness metrics for the Equalized Odds (EODD) mitigation constraint, evaluated with U-Net on the CheXpert dataset. There is minor sensitivity of AUROC to lambda; fairness metrics show greater variance but minimal substantial improvement with increased $\lambda$.}
    \label{fig:eodd-lambda-auroc}
    \end{figure*}

    \begin{figure*}[t]
  \centering

\subfigure[]{%
\begin{minipage}[t]{\textwidth}
\centering
\includegraphics[width=0.3\textwidth]{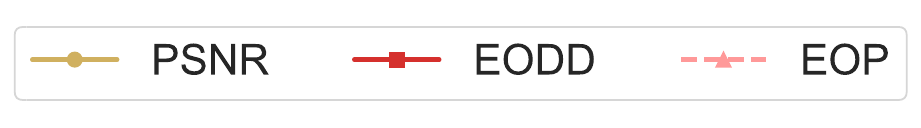}
\end{minipage}}

\begin{tabular}{c@{\hspace{0.2cm}}c@{\hspace{0.2cm}}c}

\subfigure[Age]{%
\begin{minipage}[t]{0.3\textwidth}
\centering
\includegraphics[width=\linewidth]{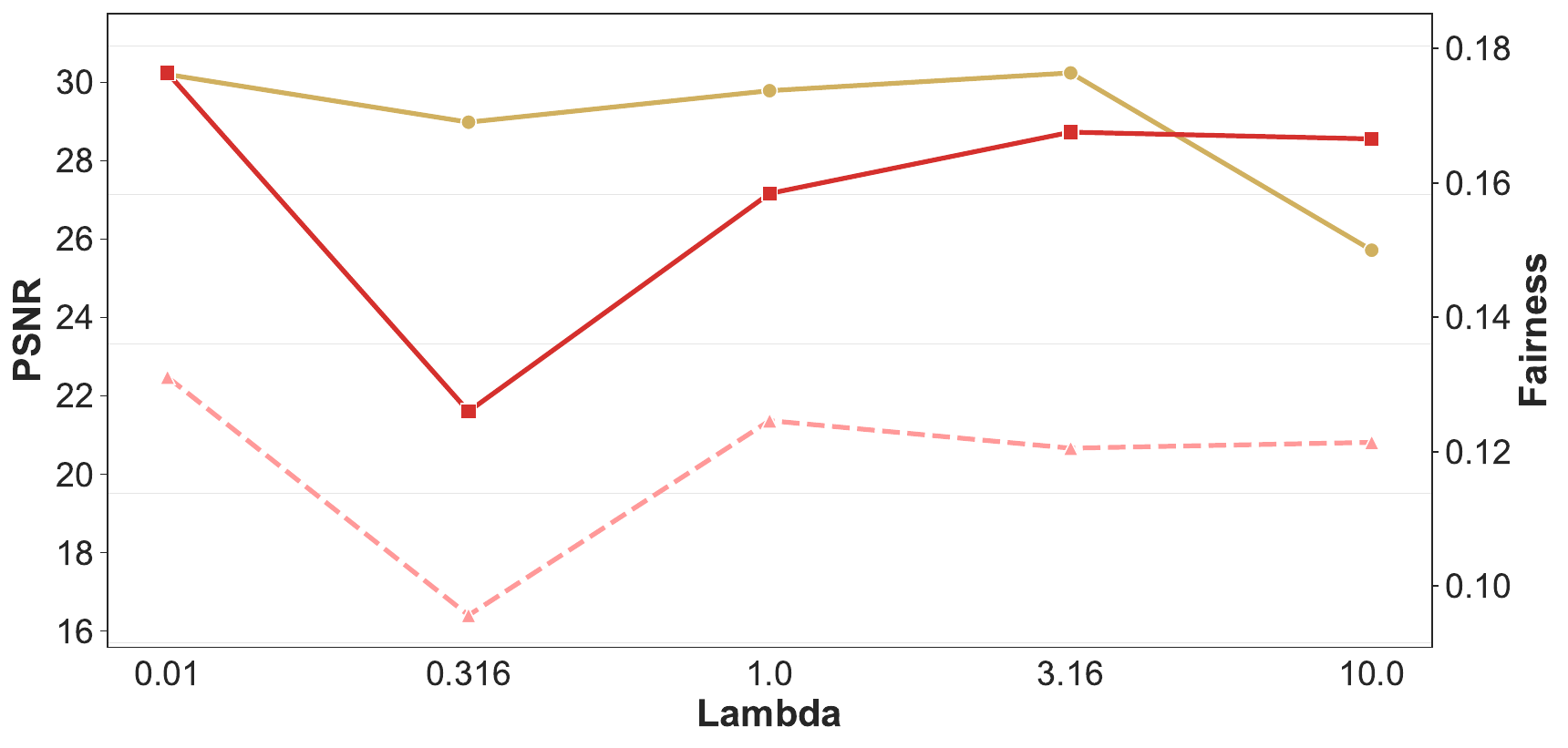}
\end{minipage}}
&
\subfigure[Sex]{%
\begin{minipage}[t]{0.3\textwidth}
\centering
\includegraphics[width=\linewidth]{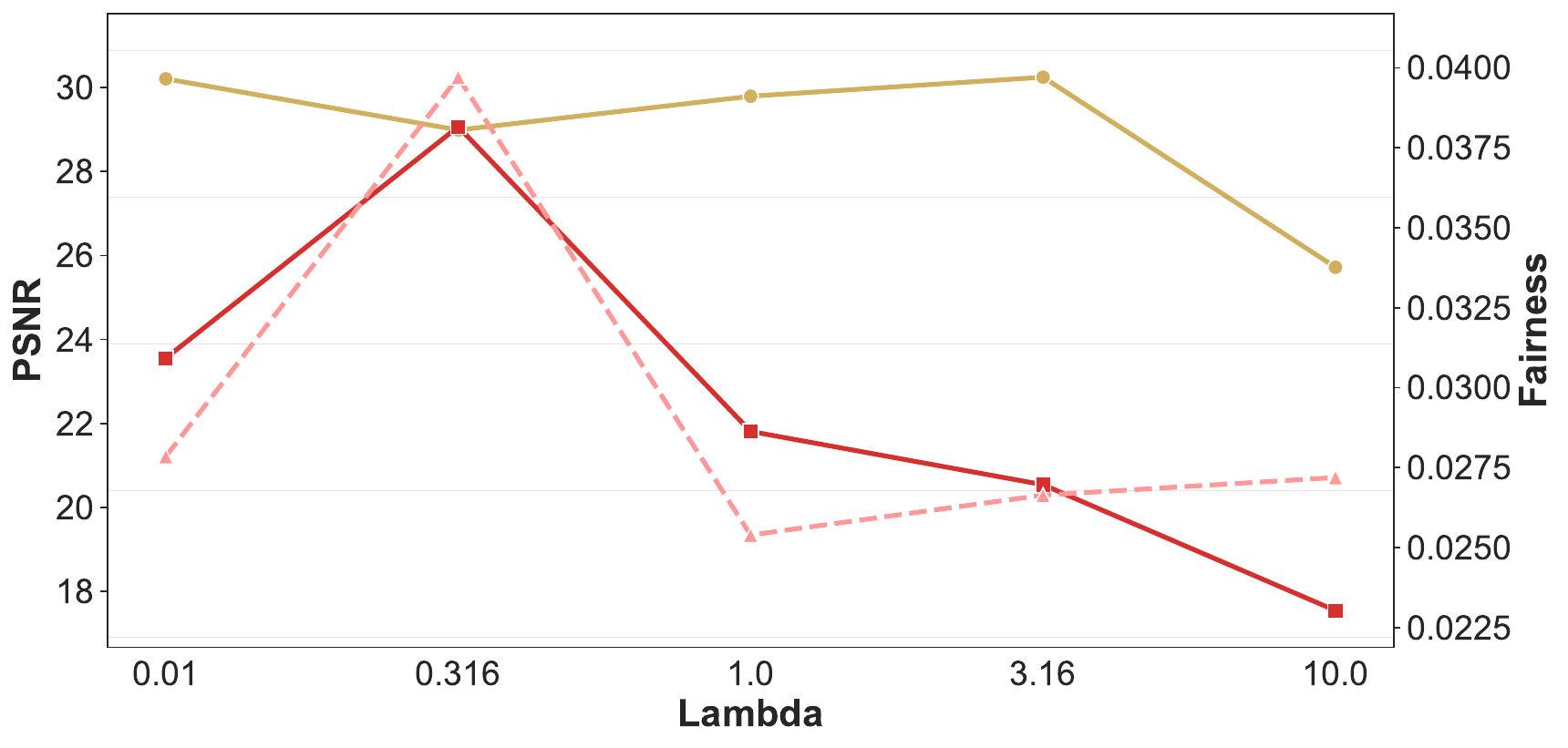}
\end{minipage}}
&
\subfigure[Race]{%
\begin{minipage}[t]{0.3\textwidth}
\centering
\includegraphics[width=\linewidth]{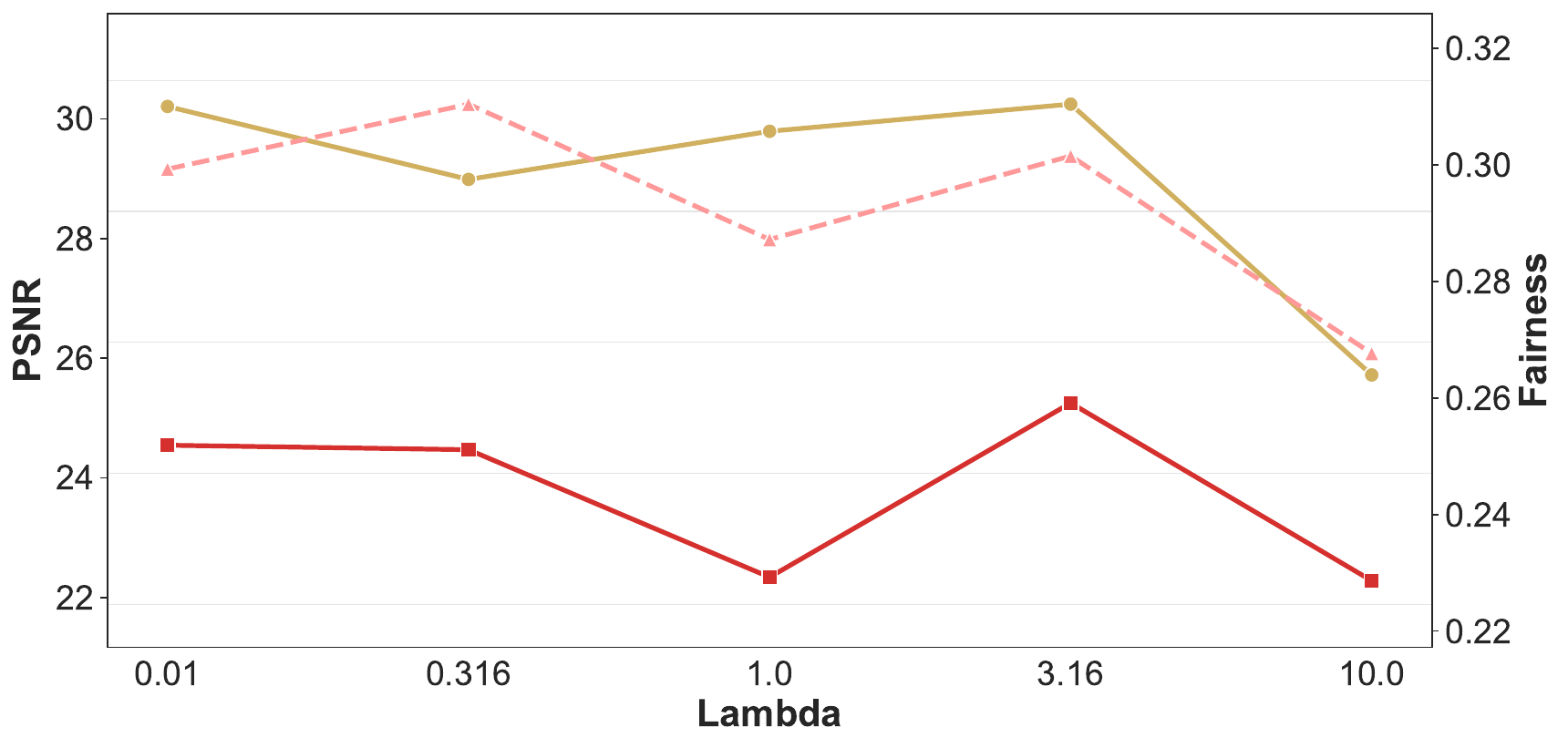}
\end{minipage}}

\end{tabular}
    \caption{Impact of $\lambda_{\mathrm{fair}}$ on reconstruction quality (PSNR) compared to fairness for the EODD constraint mitigation. PSNR remains stable across lambda variations, while fairness shows slight variation without substantial improvement.}
    \label{fig:eodd-lambda-psnr}
    \end{figure*}


\begin{figure*}[t]
\centering

\includegraphics[width=0.5\linewidth]{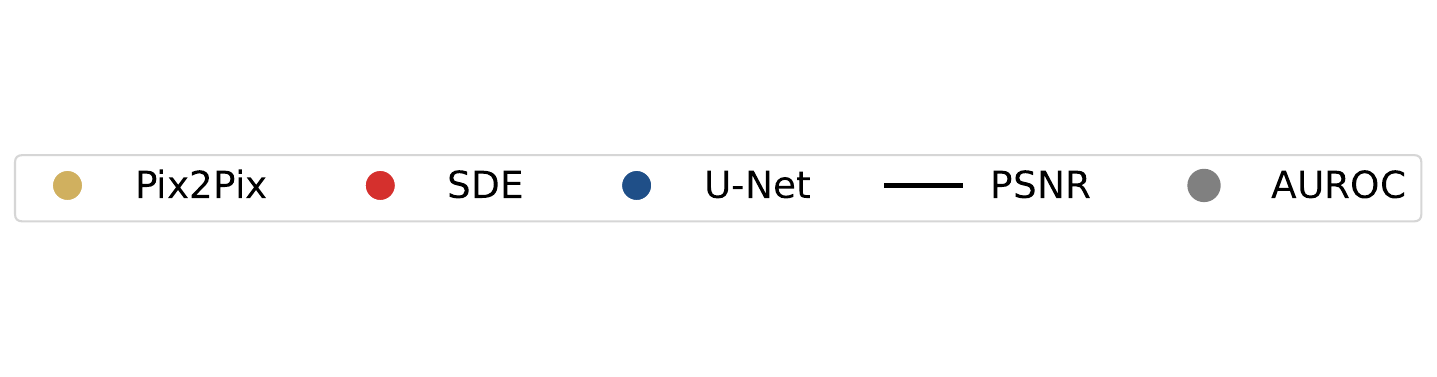}
\vspace{0.5em}

\subfigure[Reweighting]{%
\begin{minipage}[t]{\textwidth}
\centering
\includegraphics[width=0.45\linewidth]{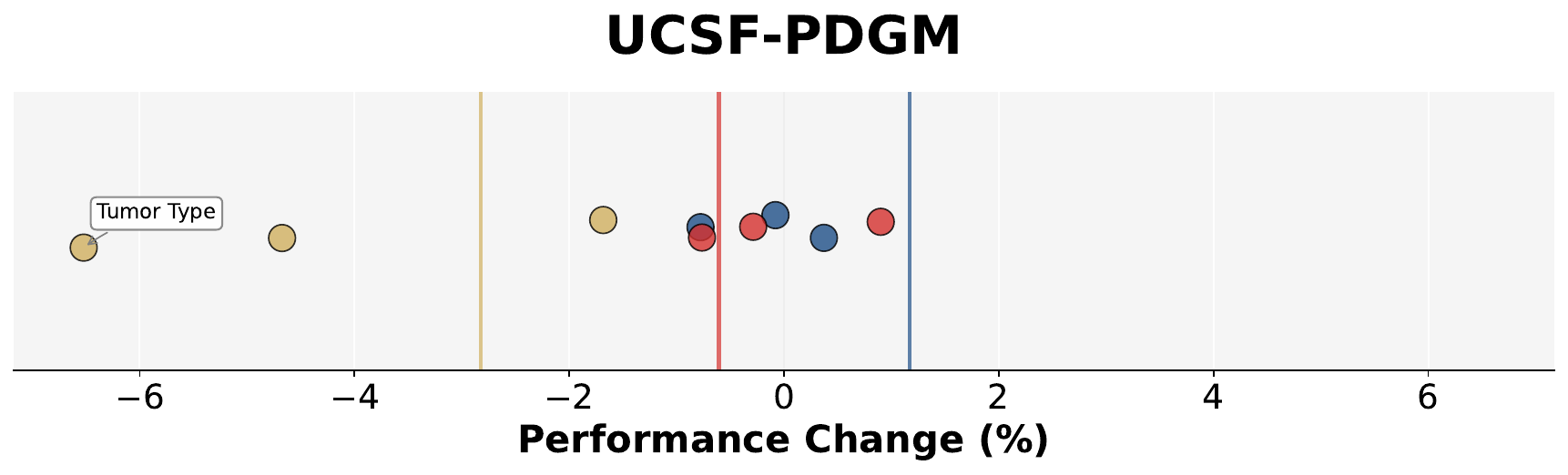}
\hspace{1em}
\includegraphics[width=0.45\linewidth]{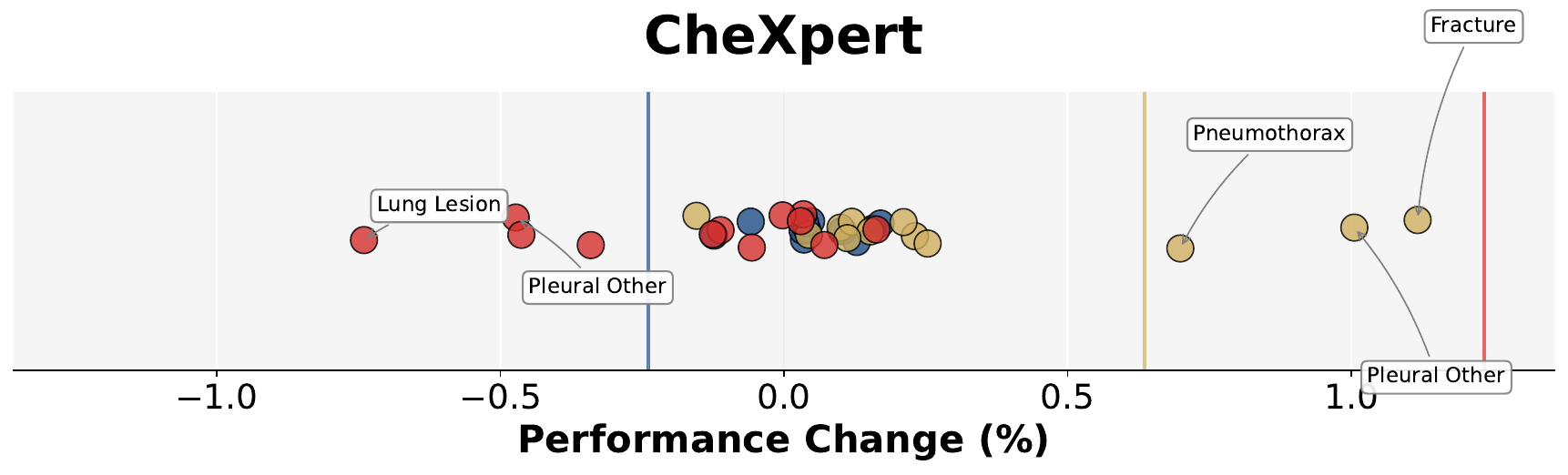}
\end{minipage}}
\vspace{1em}

\subfigure[Equalized odds constraint]{%
\begin{minipage}[t]{\textwidth}
\centering
\includegraphics[width=0.45\linewidth]{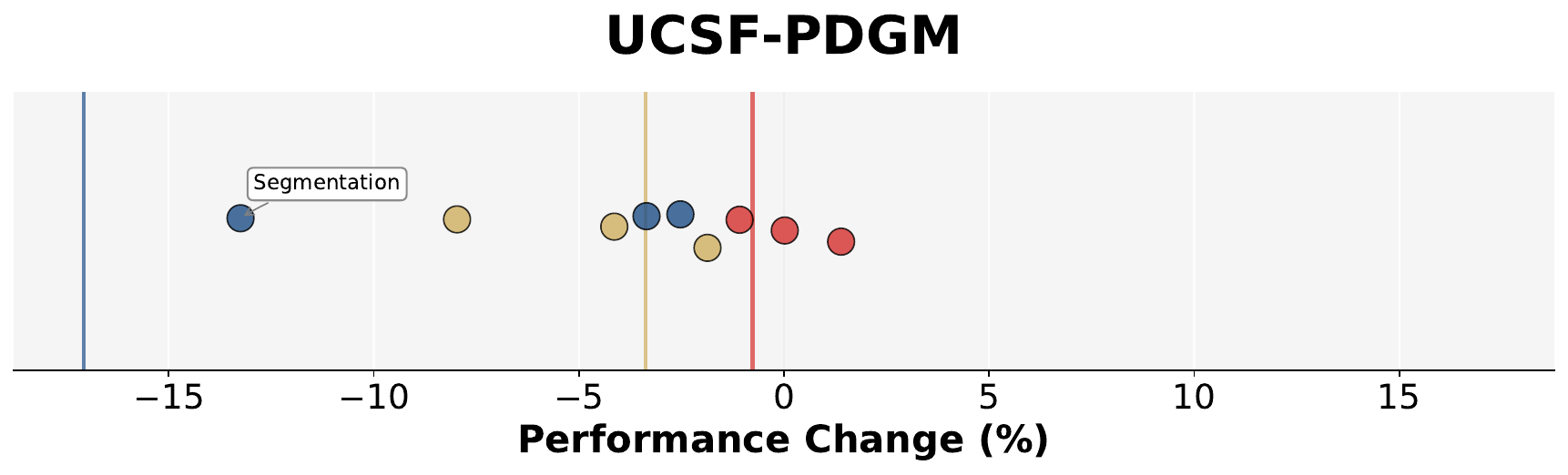}
\hspace{1em}
\includegraphics[width=0.45\linewidth]{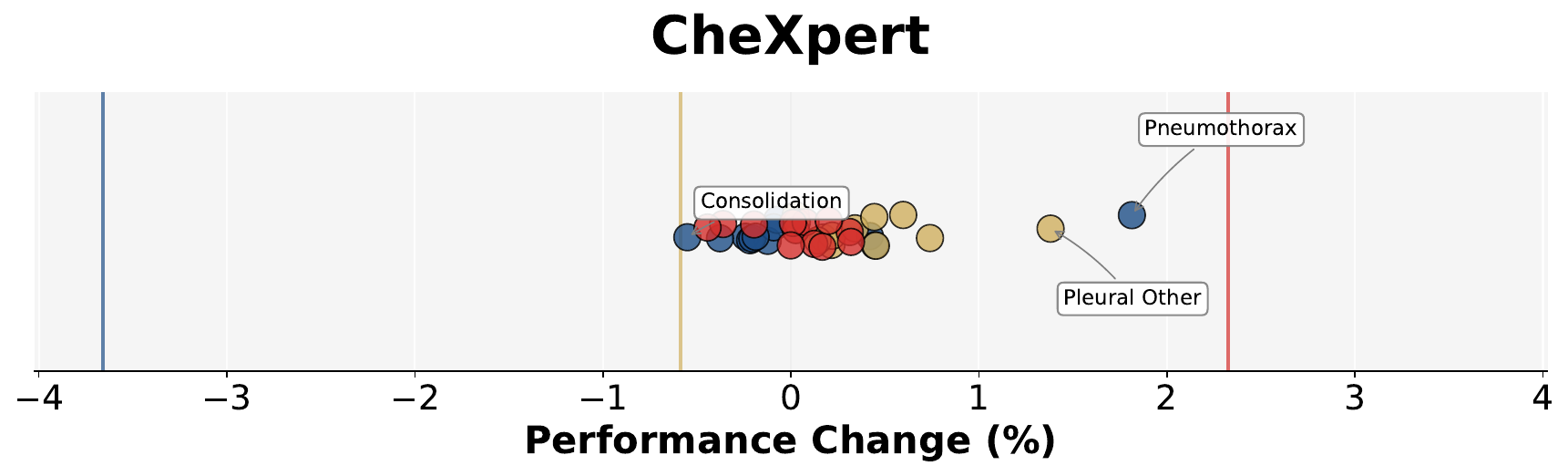}
\end{minipage}}
\vspace{1em}

    \caption{Change in prediction performance after applying bias mitigation techniques. Each row compares two datasets for a given method: (a) Reweighted sampling, (b) Equalized odds constraint. UCSF-PDGM experiences more performance degradation. However, all techniques show good stability in task performance, with few outliers in the UCSF-PDGM dataset.}
    \label{app:mitigation_performance}
\end{figure*}

\subsection{Additional Fairness Results}\label{app:fairness_results}
In addition to Equalized Odds and Skewed Error Ratio in the main text, we investigate two additional bias metrics: 

\paragraph{Equality of Opportunity (EOP):} 
\begin{align*}
P&(\hat{Y} = 1 \mid Y = 1, A = 0) \\ &= P(\hat{Y} = 1 \mid Y = 1, A = 1).
\end{align*}

We report the worst case Equality of Opportunity \cite{DBLP:journals/corr/HardtPS16} difference between groups
\begin{align*}
max_{i,j}|P&(\hat{Y} = 1 \mid Y = 1, A = i) \\&- P(\hat{Y} = 1 \mid Y = 1, A = j)|, \\& \quad 
\forall \quad \text{A} \in \mathcal{A}.
\end{align*}

EOP is a relaxation of EODD, requiring fairness only concerning the positive class (\(Y=1\)).

\paragraph{\(\Delta \text{Dice}\):}
Given the limited availability of dedicated segmentation fairness metrics, we also compute:
\[
\Delta \text{Dice} = \max_{i, j} \left|\text{Dice}_{A_i} - \text{Dice}_{A_j}\right|, A \in \mathcal{A}
\]
which represents the maximum difference in Dice across all protected subgroups \(\mathcal{A}\).

\begin{figure*}[t]
\centering

\includegraphics[width=\textwidth]{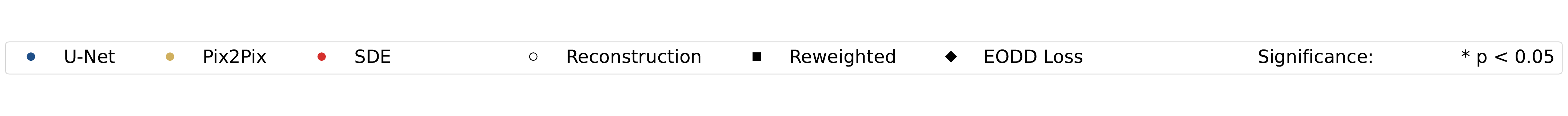}

\begin{tabular}{c@{\hspace{0.2cm}}c}

\subfigure[]{%
\begin{minipage}[t]{0.49\textwidth}
\centering
\includegraphics[width=\linewidth]{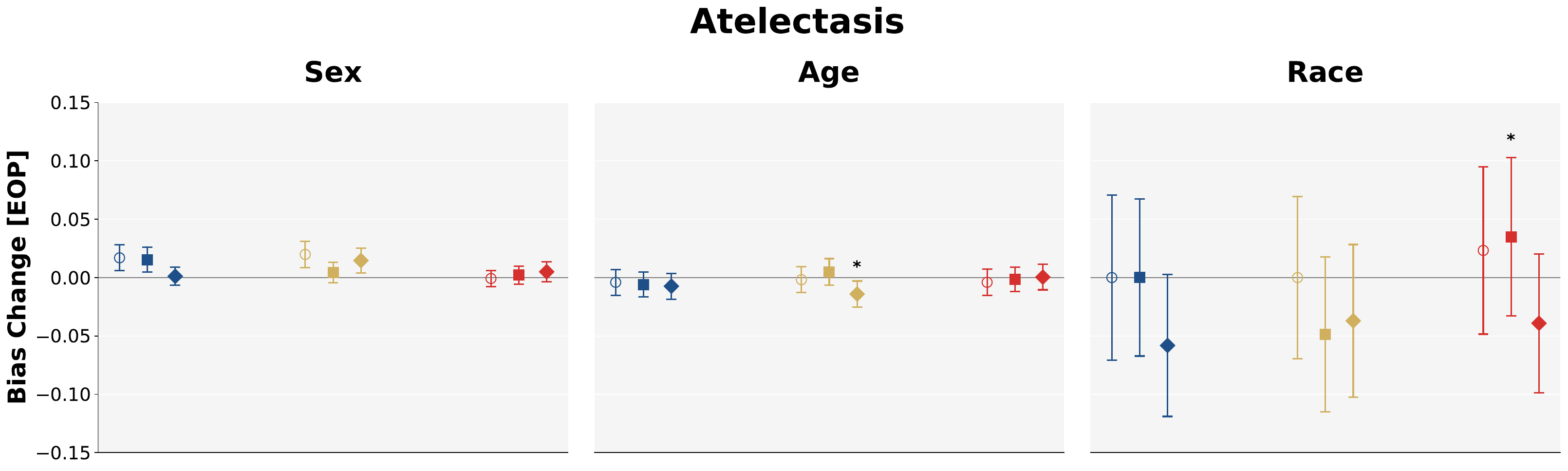}
\end{minipage}} 
&
\subfigure[]{%
\begin{minipage}[t]{0.49\textwidth}
\centering
\includegraphics[width=\linewidth]{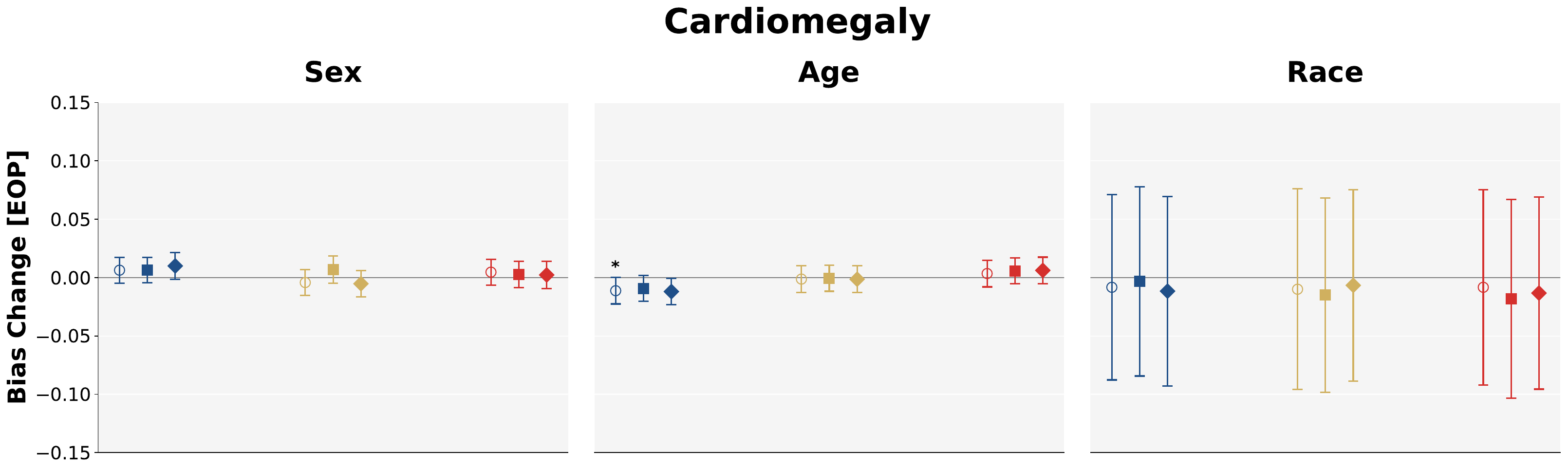}
\end{minipage}}
\\[0.3cm]

\subfigure[]{%
\begin{minipage}[t]{0.49\textwidth}
\centering
\includegraphics[width=\linewidth]{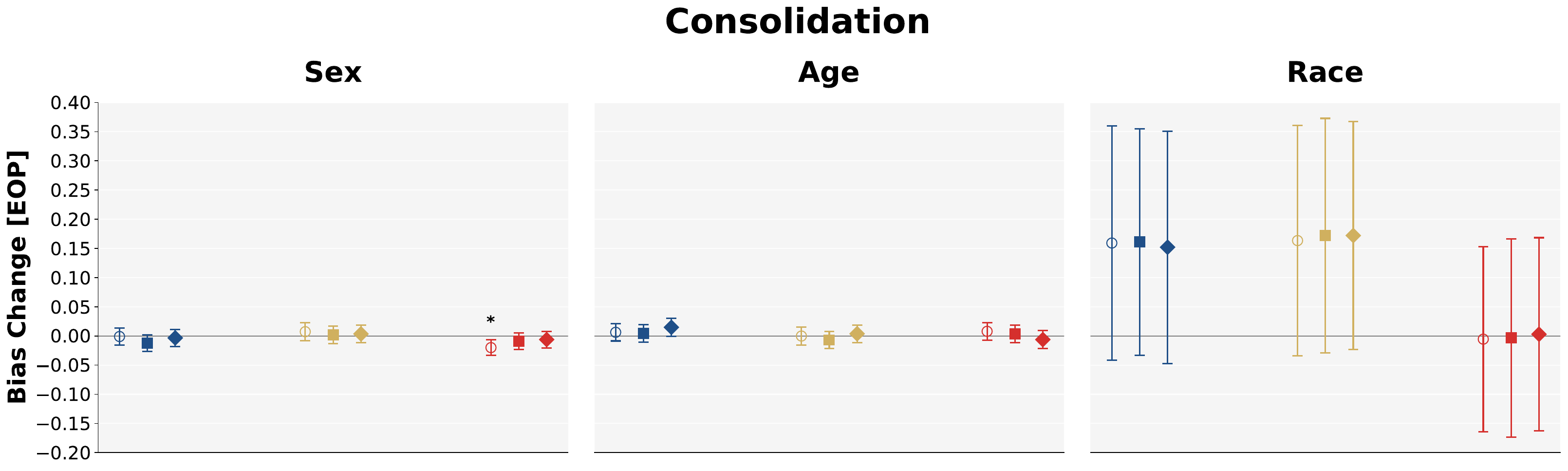}
\end{minipage}}
&
\subfigure[]{%
\begin{minipage}[t]{0.49\textwidth}
\centering
\includegraphics[width=\linewidth]{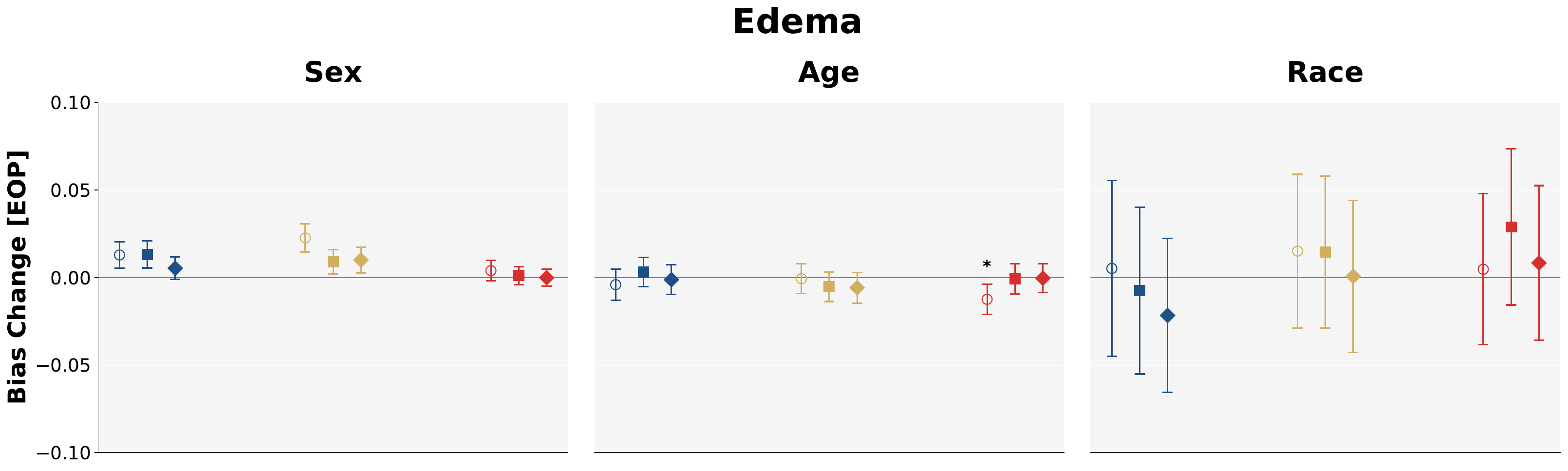}
\end{minipage}}
\\[0.3cm]

\subfigure[]{%
\begin{minipage}[t]{0.49\textwidth}
\centering
\includegraphics[width=\linewidth]{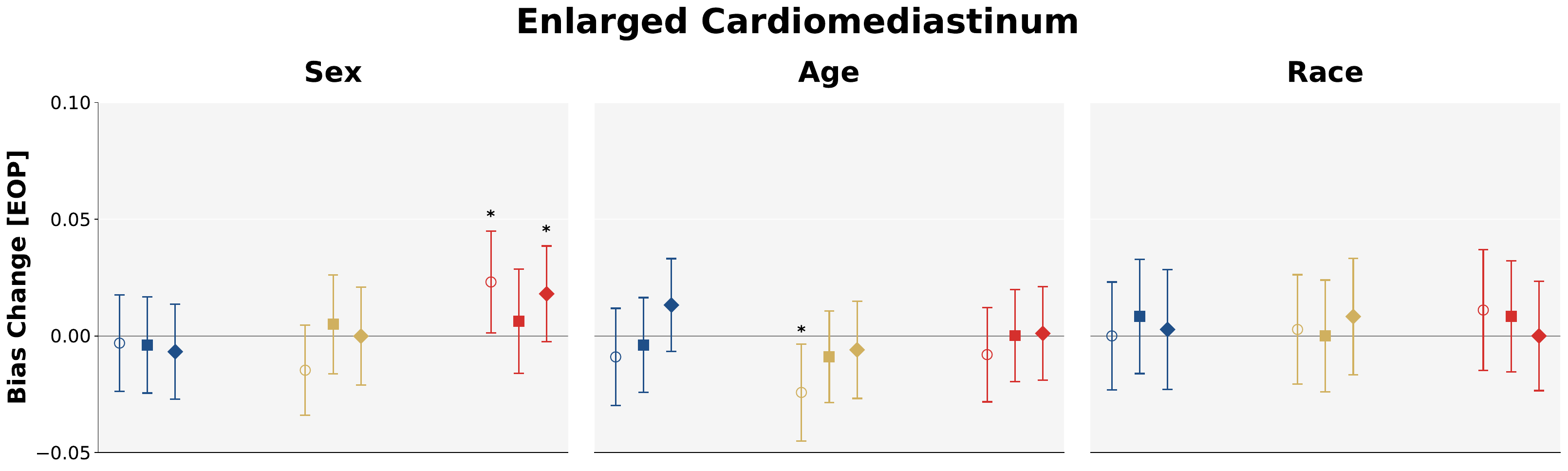}
\end{minipage}}
&
\subfigure[]{%
\begin{minipage}[t]{0.49\textwidth}
\centering
\includegraphics[width=\linewidth]{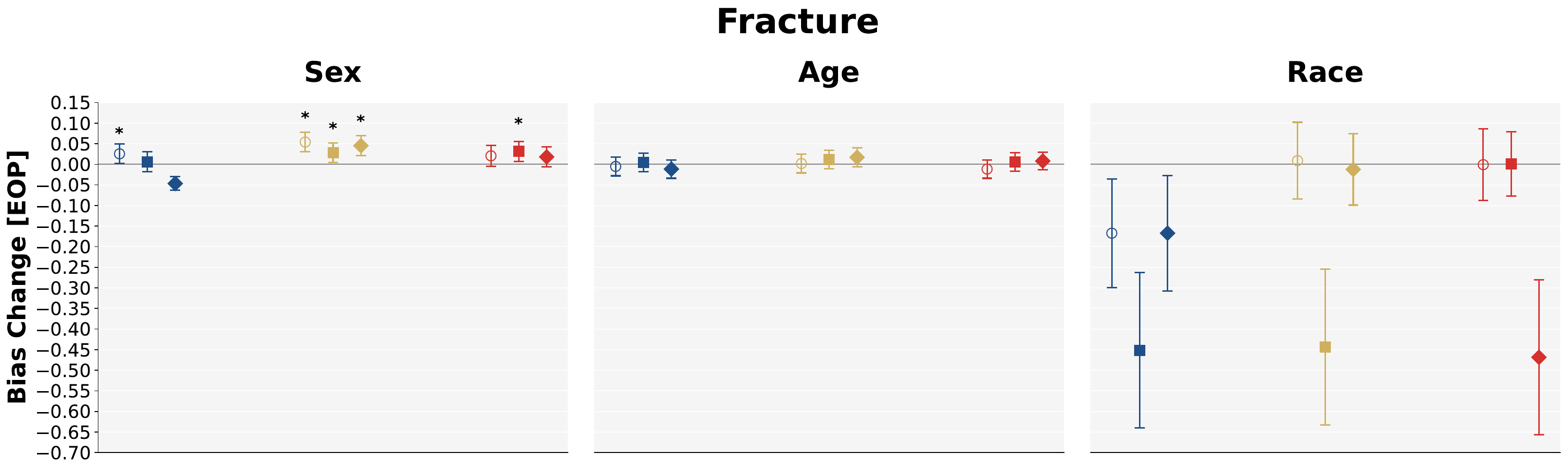}
\end{minipage}}
\\[0.3cm]

\subfigure[]{%
\begin{minipage}[t]{0.49\textwidth}
\centering
\includegraphics[width=\linewidth]{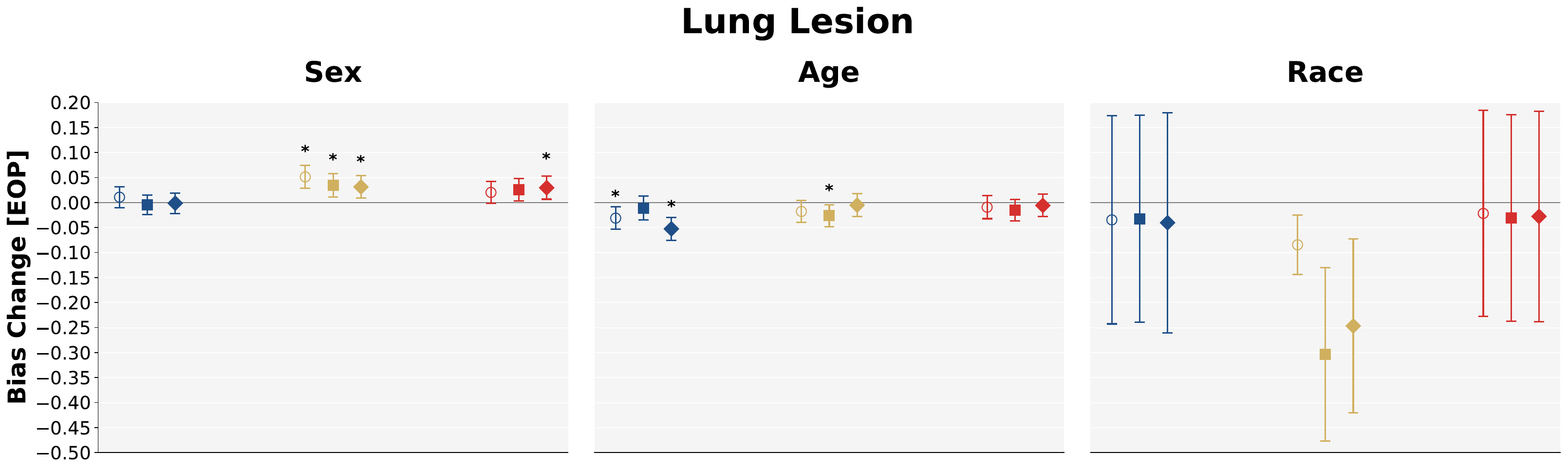}
\end{minipage}}
&
\subfigure[]{%
\begin{minipage}[t]{0.49\textwidth}
\centering
\includegraphics[width=\linewidth]{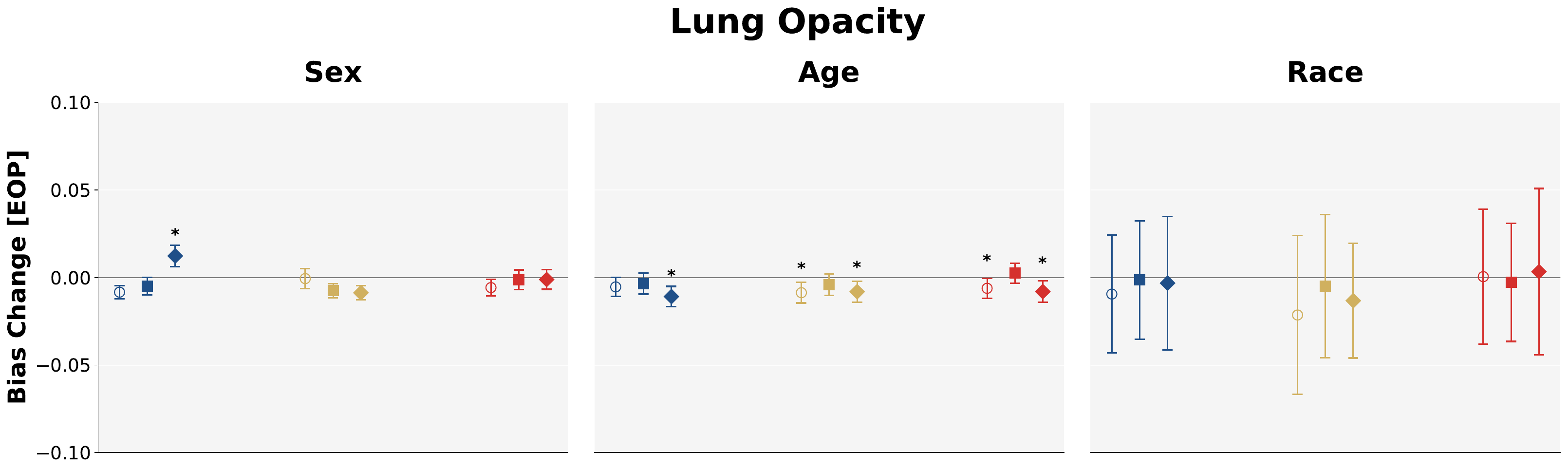}
\end{minipage}}
\\[0.3cm]

\subfigure[]{%
\begin{minipage}[t]{0.49\textwidth}
\centering
\includegraphics[width=\linewidth]{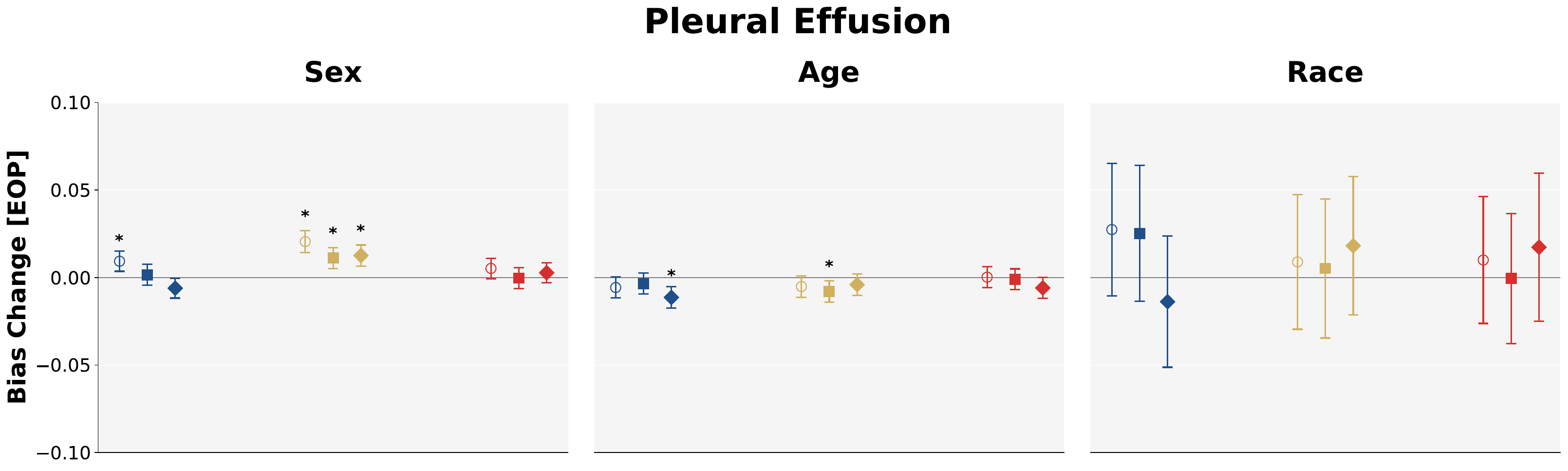}
\end{minipage}}
&
\subfigure[]{%
\begin{minipage}[t]{0.49\textwidth}
\centering
\includegraphics[width=\linewidth]{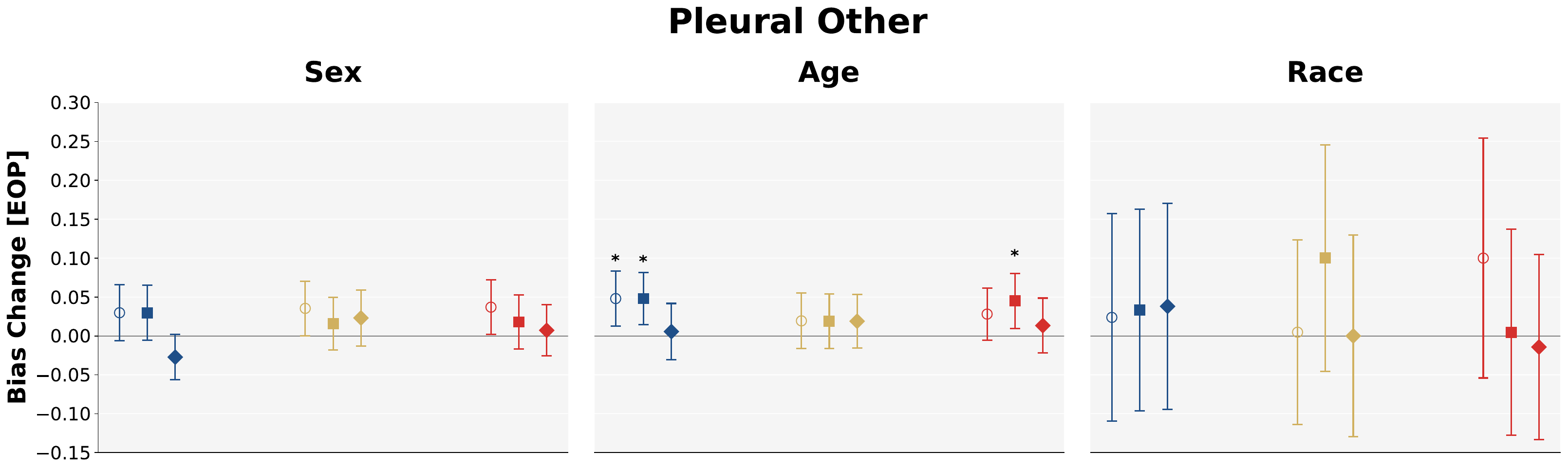}
\end{minipage}}
\\[0.3cm]

\subfigure[]{%
\begin{minipage}[t]{0.49\textwidth}
\centering
\includegraphics[width=\linewidth]{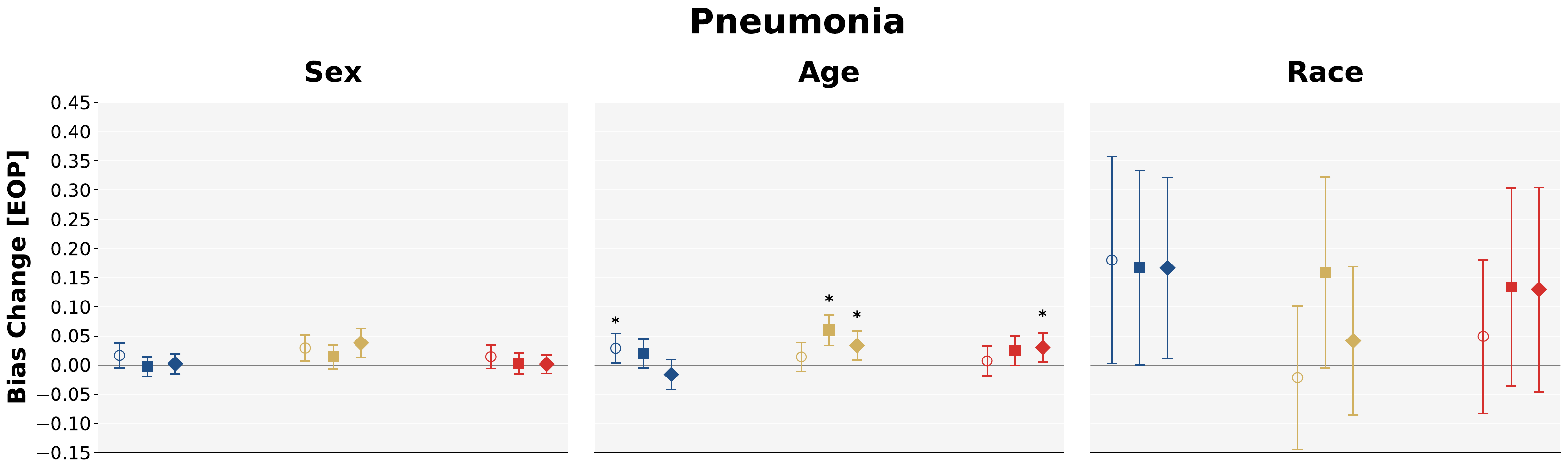}
\end{minipage}}
&
\subfigure[]{%
\begin{minipage}[t]{0.49\textwidth}
\centering
\includegraphics[width=\linewidth]{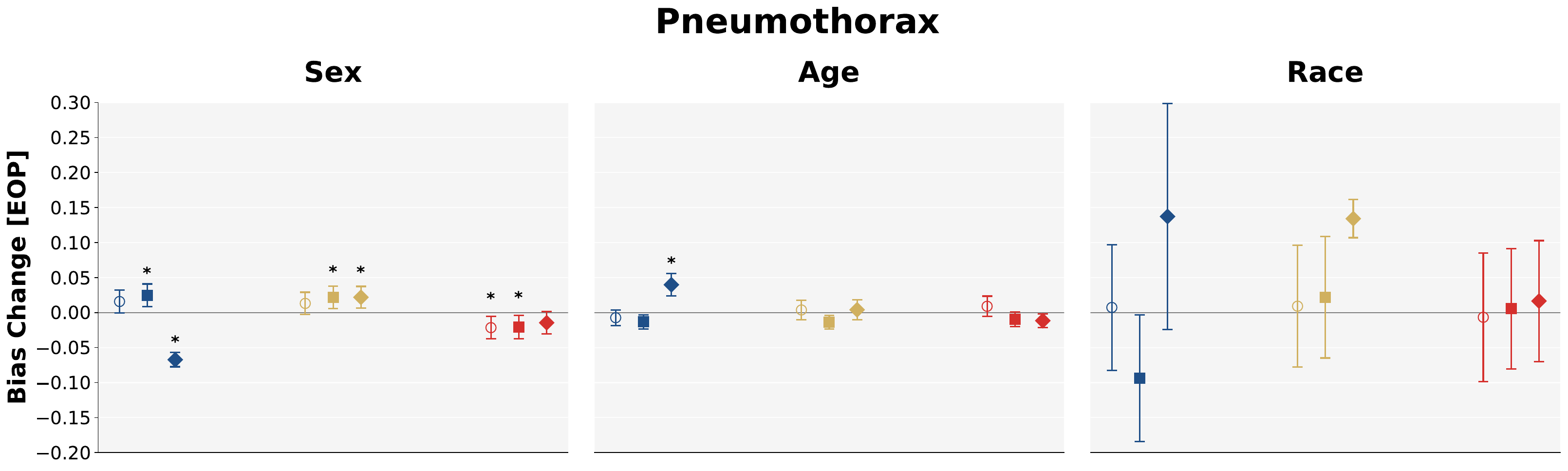}
\end{minipage}}

\end{tabular}

\caption{Equality of opportunity (EOP) bias change pre- and post-mitigation compared to predictions on original images for CheXpert classification. Pre-mitigation, bias tends to increase slightly for sex; race exhibits high variance. Bias tends to decline slightly post-mitigation.}
\label{fig:bias_chexpert_eop}

\end{figure*}

\begin{figure*}[t]
\centering

\includegraphics[width=\textwidth]{plots/fairness/eop/evaluation_midl_camera_fairness_legend.pdf}

\begin{tabular}{ccc}

\subfigure[]{%
\begin{minipage}[t]{0.3\textwidth}
\centering
\includegraphics[width=\linewidth]{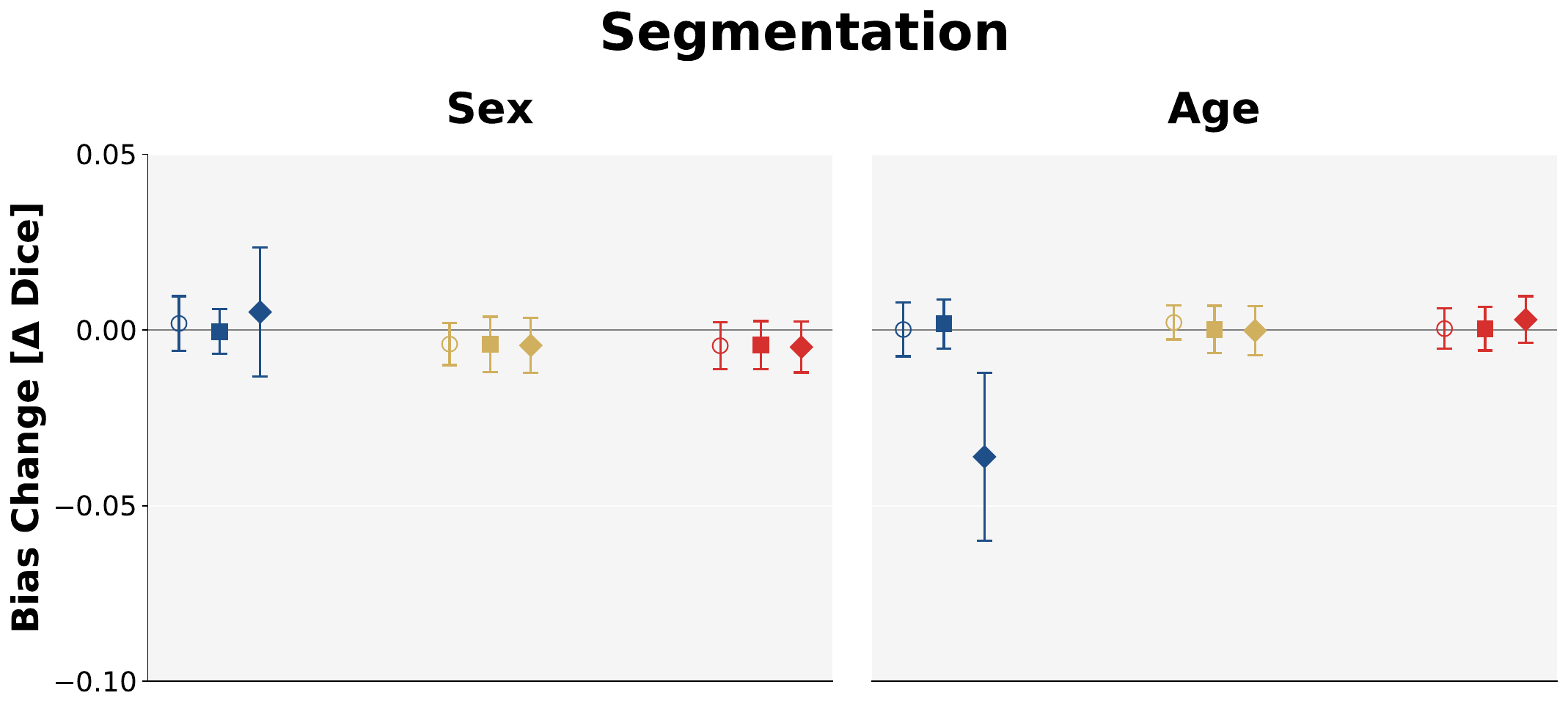}
\end{minipage}}
&
\subfigure[]{%
\begin{minipage}[t]{0.3\textwidth}
\centering
\includegraphics[width=\linewidth]{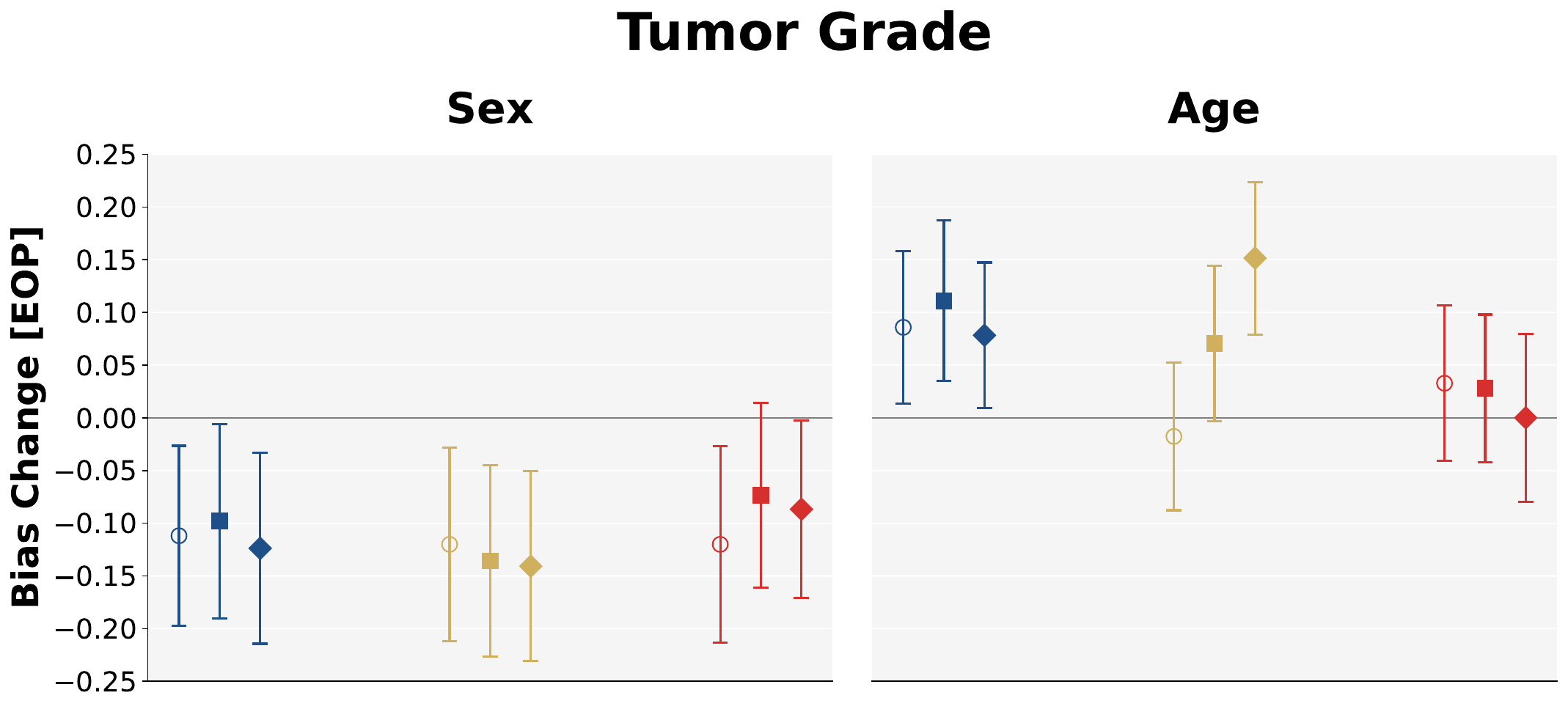}
\end{minipage}}
&
\subfigure[]{%
\begin{minipage}[t]{0.3\textwidth}
\centering
\includegraphics[width=\linewidth]{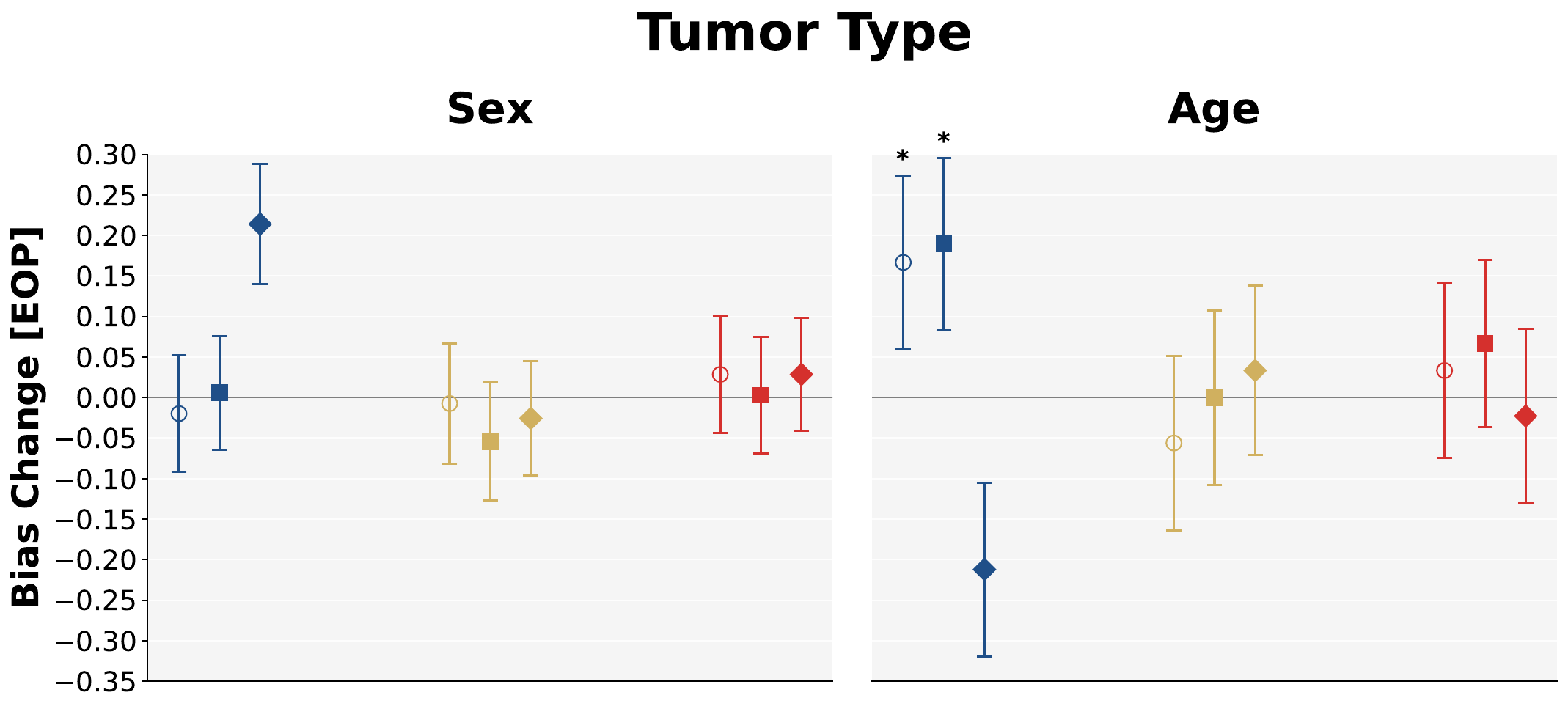}
\end{minipage}}

\end{tabular}

\caption{Equality of opportunity (EOP) and $\Delta$ Dice bias change compared to predictions on original images pre- and post-mitigation for UCSF-PDGM classification and segmentation.}
\label{fig:bias_ucsf_class_eop}

\end{figure*}

Plots containing the results of these additional evaluations can be found in Figure \ref{fig:bias_chexpert_eop} and \ref{fig:bias_ucsf_class_eop}.

\begin{figure}[t!]
 \centering
 \includegraphics[width=0.7\linewidth]{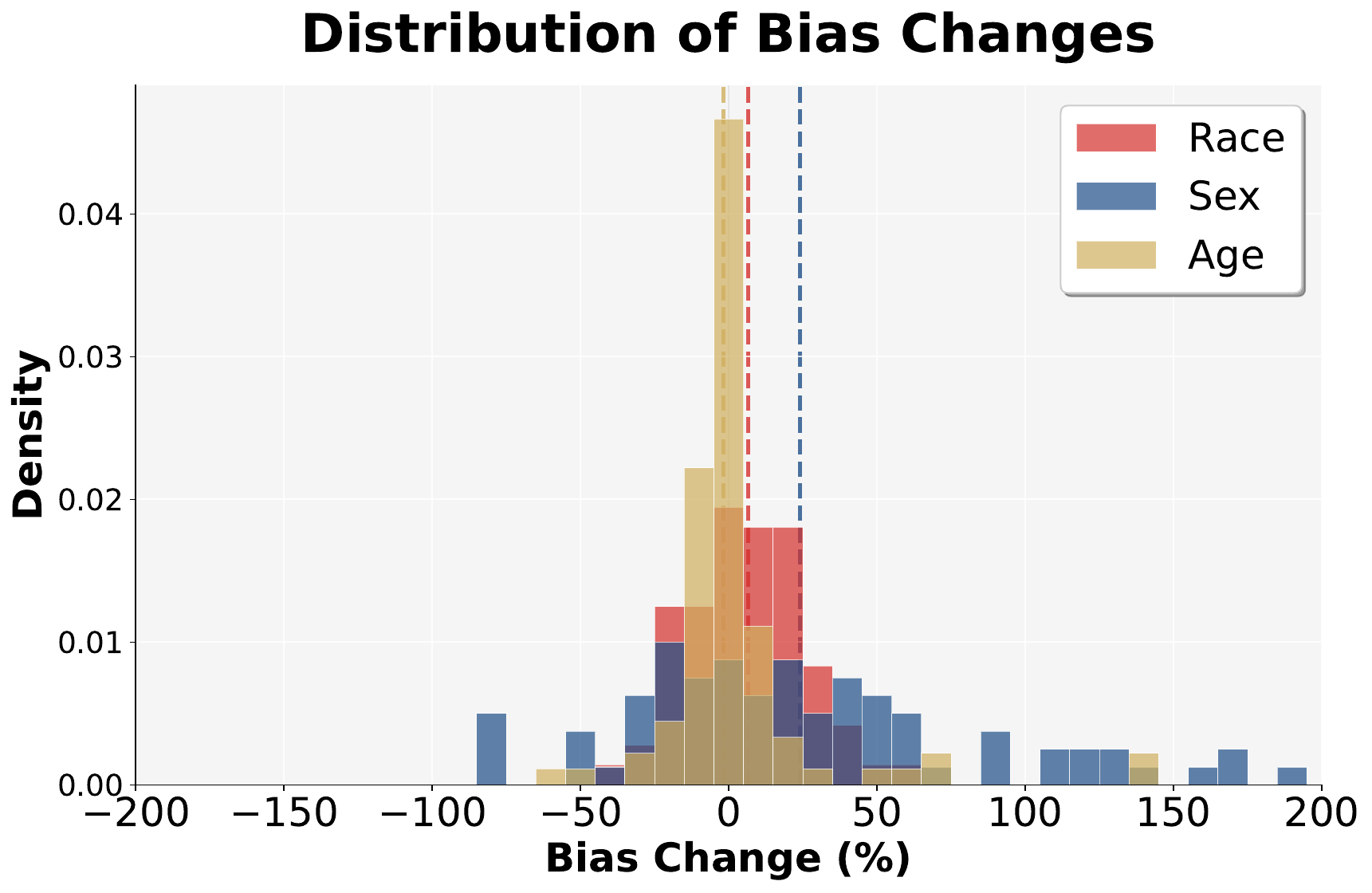}
 \vspace{-10pt}
 \caption{Distribution of bias changes when using alternative race subgroups for CheXpert calculations.}
    \label{fig:histogram_race}
\end{figure}

\begin{figure*}[!t]
\centering

\includegraphics[width=\textwidth]{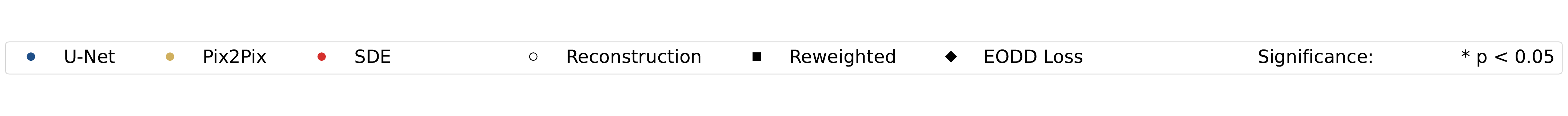}

\begin{tabular}{c@{\hspace{0.2cm}}c}

\subfigure[]{%
\begin{minipage}[t]{0.49\textwidth}
\centering
\includegraphics[width=\linewidth]{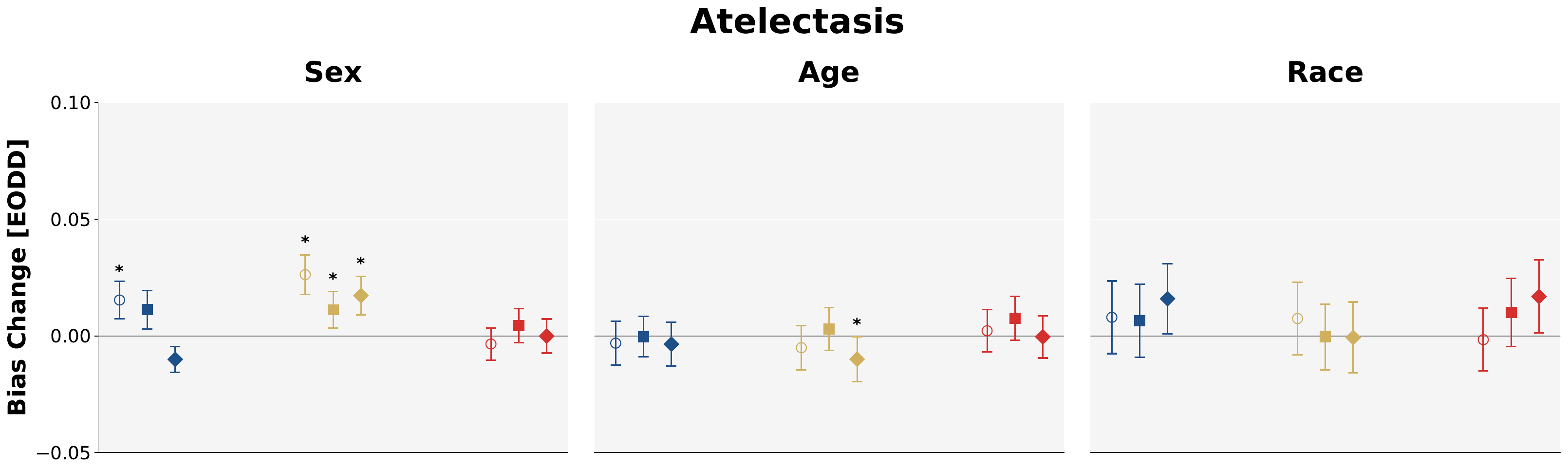}
\end{minipage}
}
&
\subfigure[]{%
\begin{minipage}[t]{0.49\textwidth}
\centering
\includegraphics[width=\linewidth]{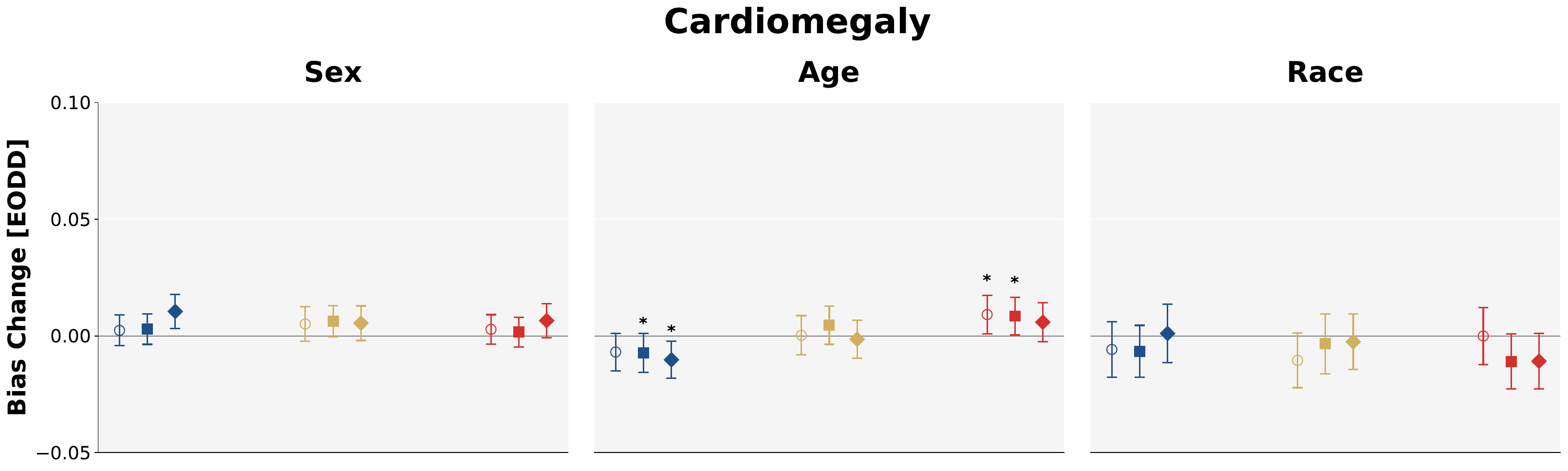}
\end{minipage}
}
\\[0.3cm]

\subfigure[]{%
\begin{minipage}[t]{0.49\textwidth}
\centering
\includegraphics[width=\linewidth]{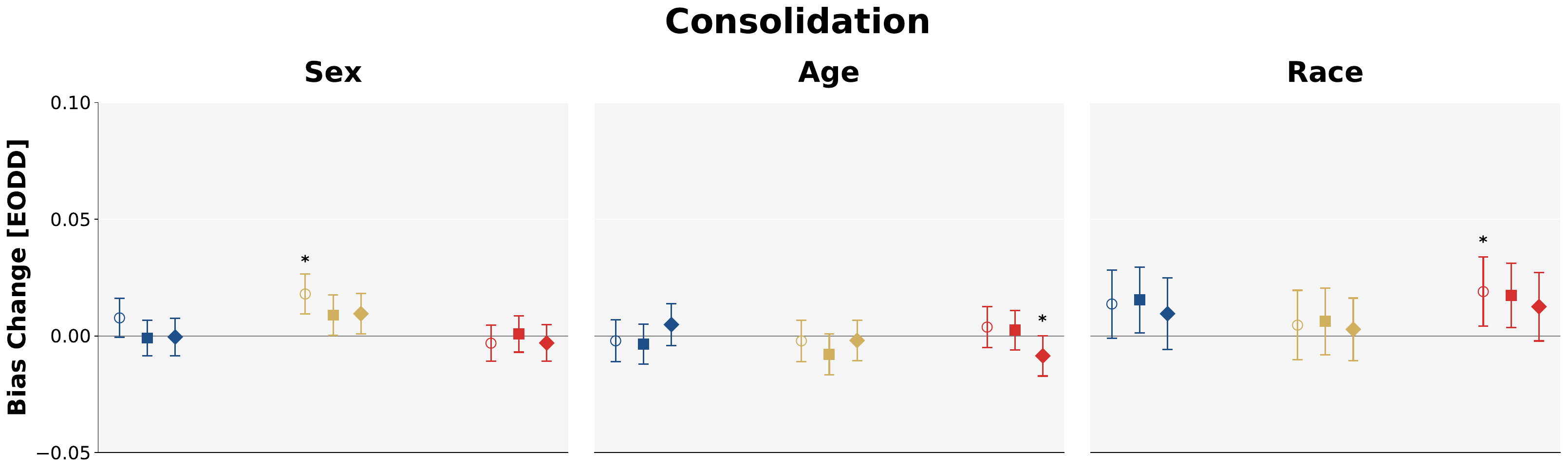}
\end{minipage}
}
&
\subfigure[]{%
\begin{minipage}[t]{0.49\textwidth}
\centering
\includegraphics[width=\linewidth]{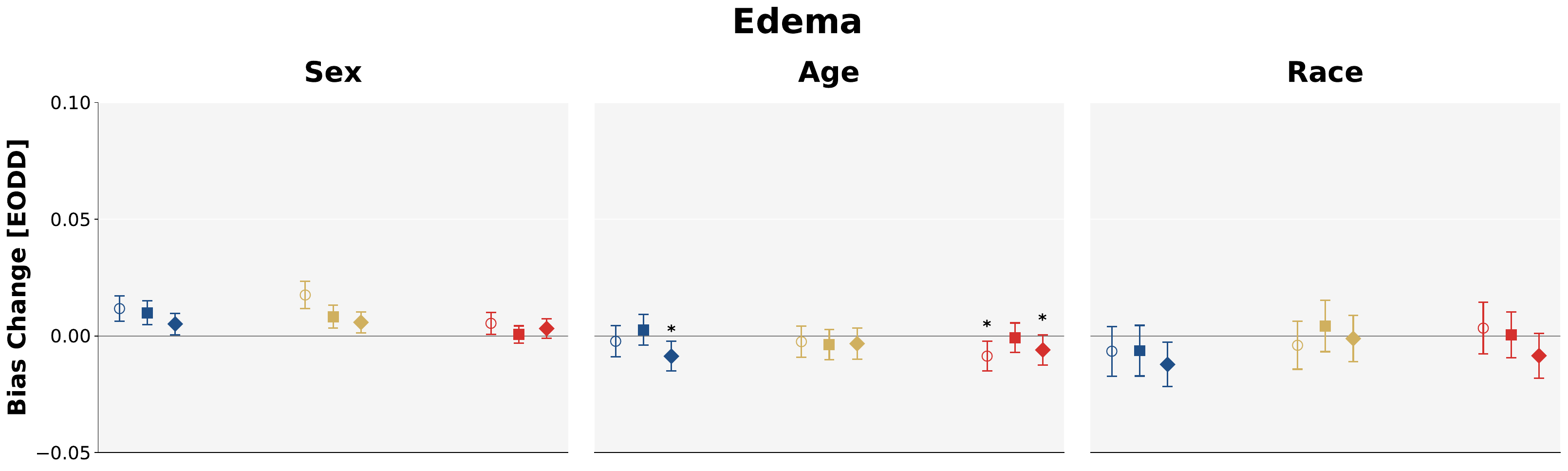}
\end{minipage}
}
\\[0.3cm]

\subfigure[]{%
\begin{minipage}[t]{0.49\textwidth}
\centering
\includegraphics[width=\linewidth]{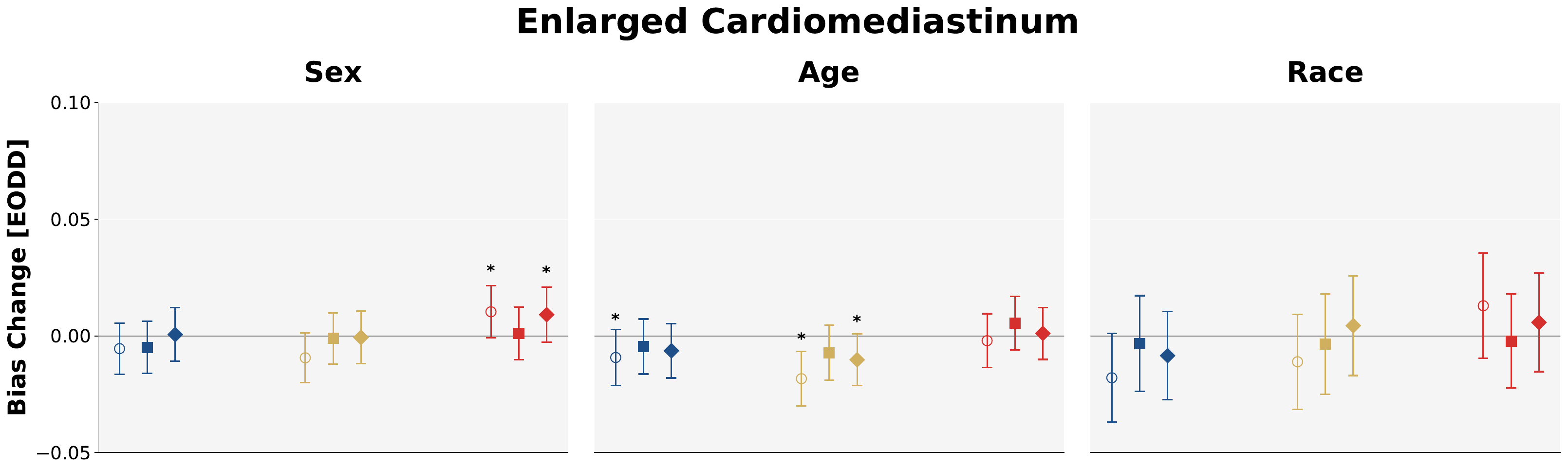}
\end{minipage}
}
&
\subfigure[]{%
\begin{minipage}[t]{0.49\textwidth}
\centering
\includegraphics[width=\linewidth]{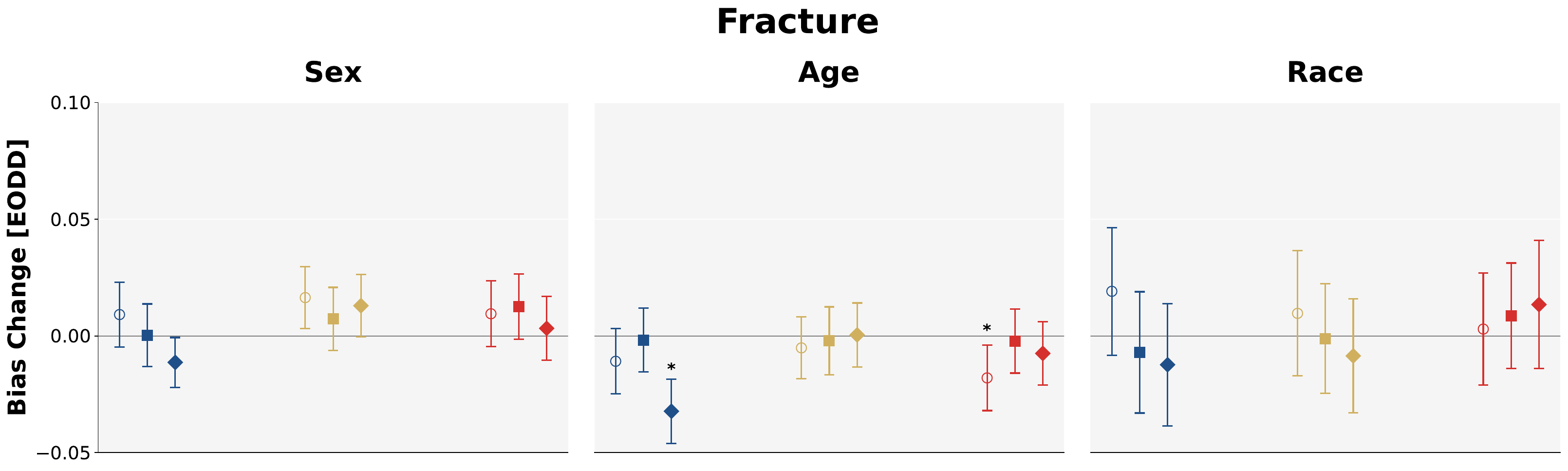}
\end{minipage}
}
\\[0.3cm]

\subfigure[]{%
\begin{minipage}[t]{0.49\textwidth}
\centering
\includegraphics[width=\linewidth]{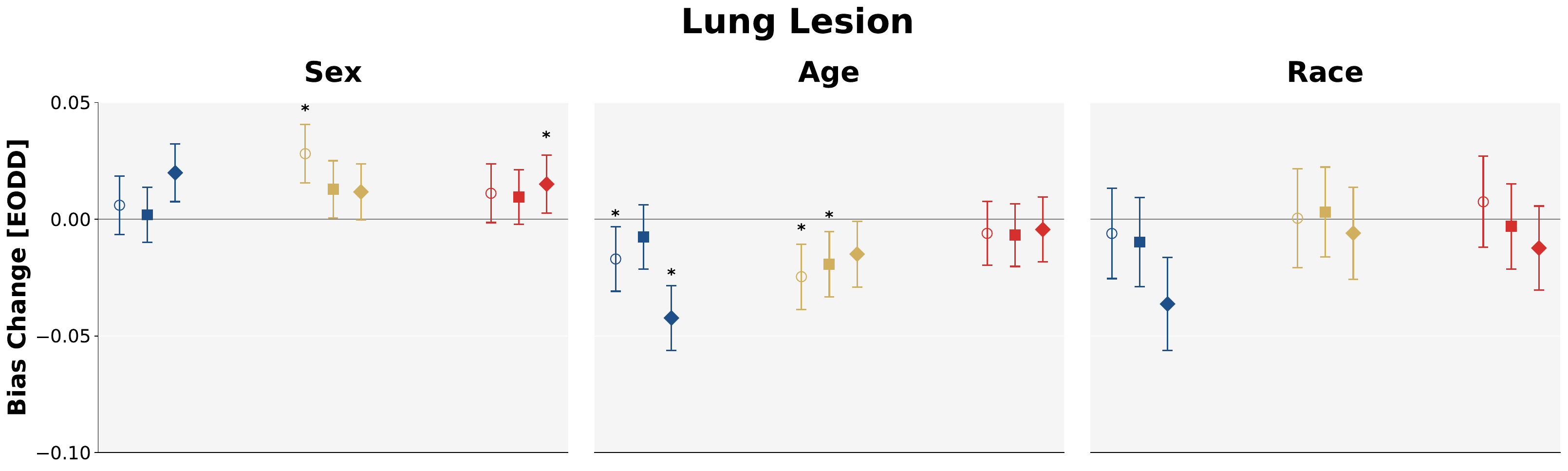}
\end{minipage}
}
&
\subfigure[]{%
\begin{minipage}[t]{0.49\textwidth}
\centering
\includegraphics[width=\linewidth]{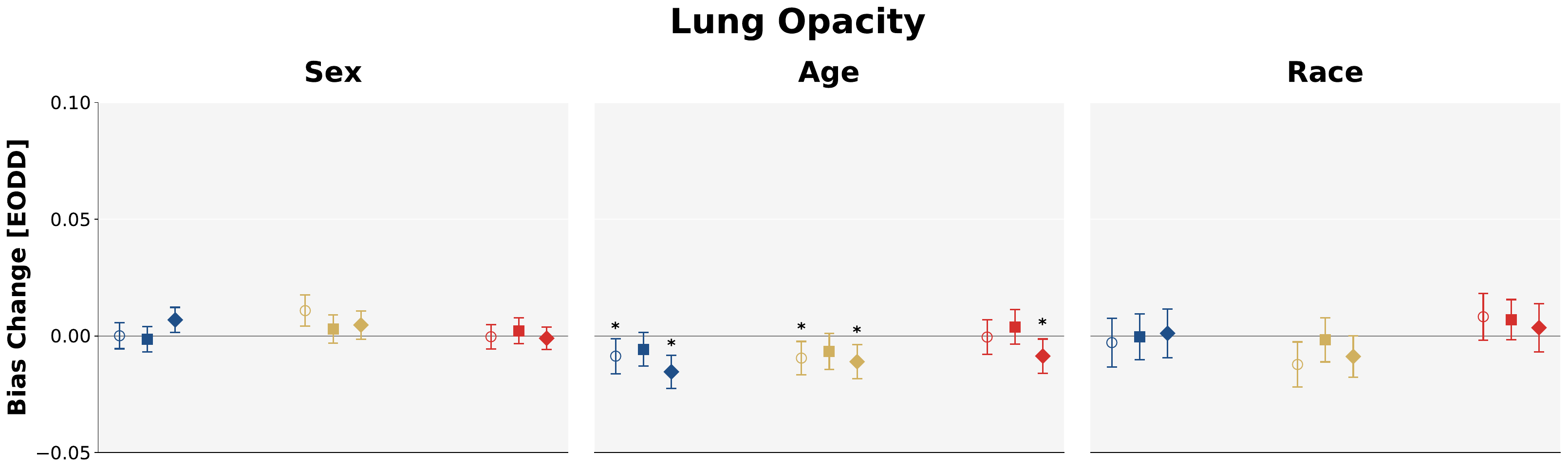}
\end{minipage}
}
\\[0.3cm]

\subfigure[]{%
\begin{minipage}[t]{0.49\textwidth}
\centering
\includegraphics[width=\linewidth]{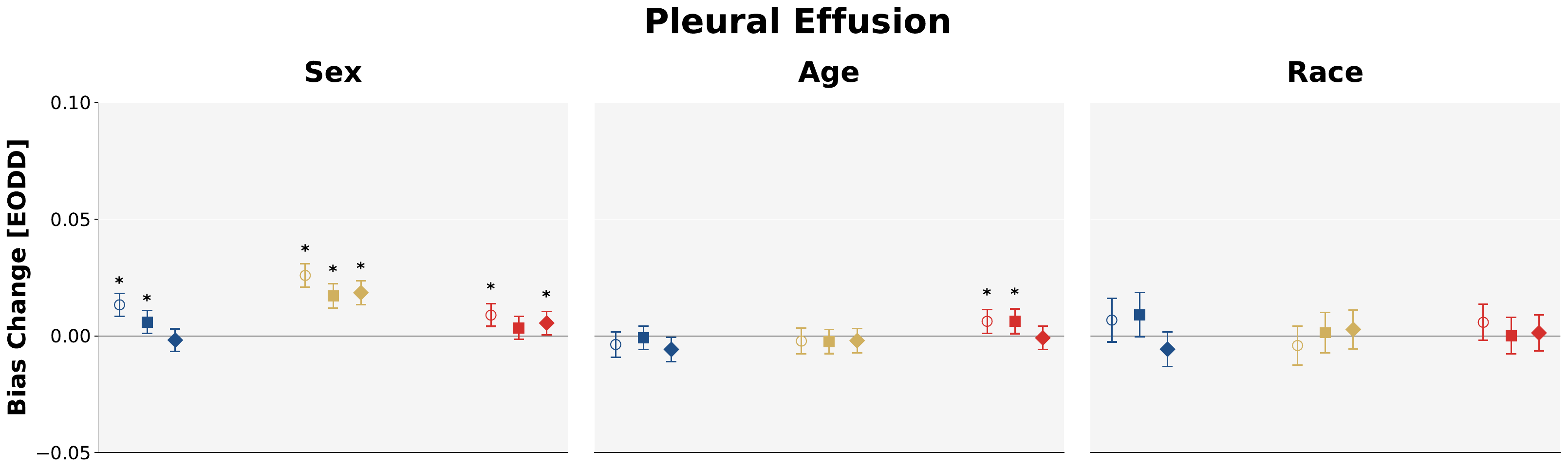}
\end{minipage}
}
&
\subfigure[]{%
\begin{minipage}[t]{0.49\textwidth}
\centering
\includegraphics[width=\linewidth]{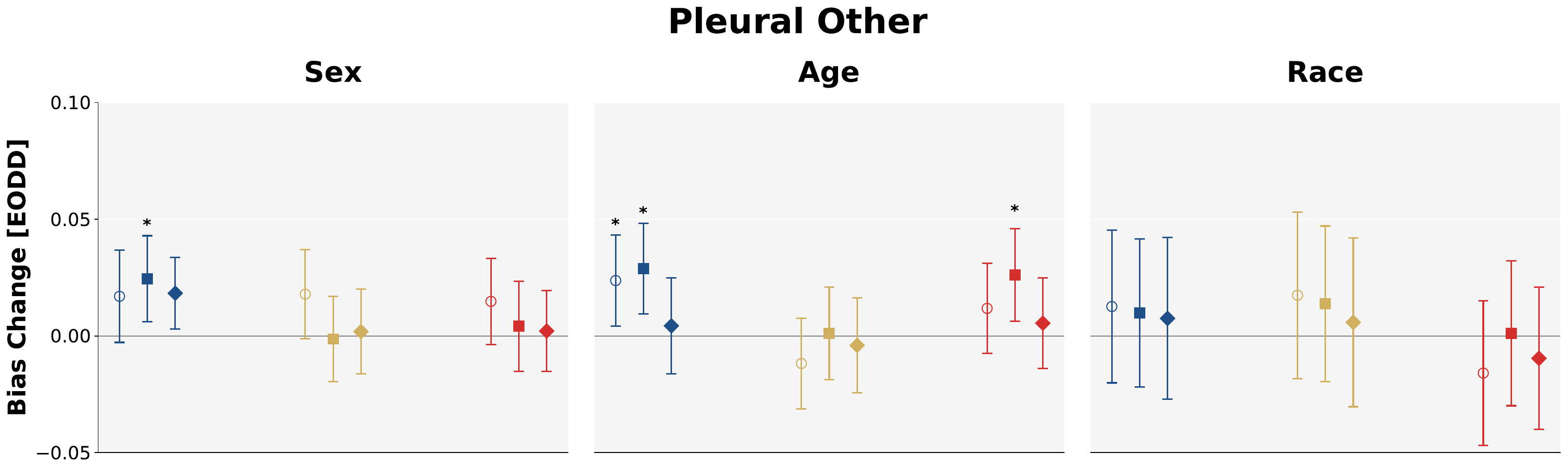}
\end{minipage}
}
\\[0.3cm]

\subfigure[]{%
\begin{minipage}[t]{0.49\textwidth}
\centering
\includegraphics[width=\linewidth]{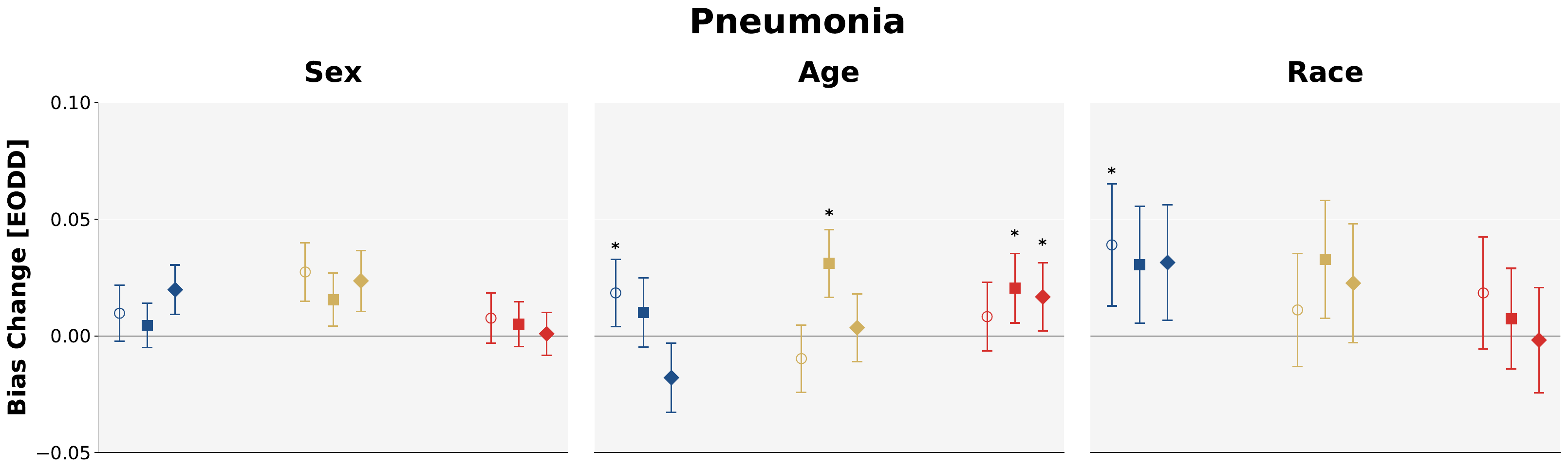}
\end{minipage}
}
&
\subfigure[]{%
\begin{minipage}[t]{0.49\textwidth}
\centering
\includegraphics[width=\linewidth]{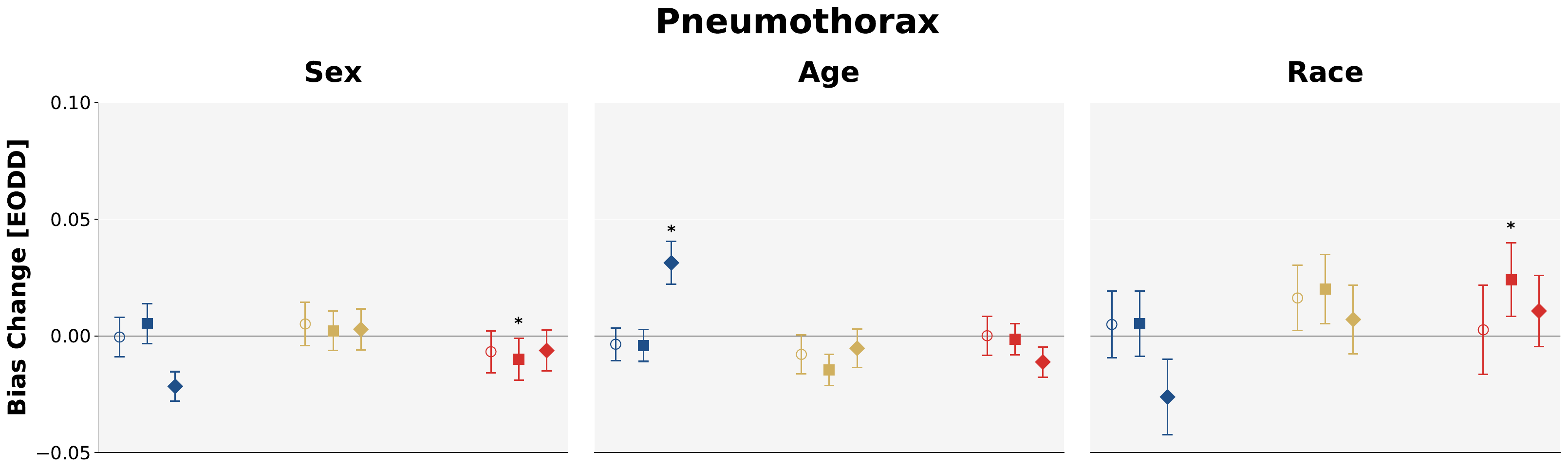}
\end{minipage}
}

\end{tabular}

\caption{Equalized odds bias change pre- and post-mitigation compared to predictions on original images for CheXpert when using alternative race subgroups.}
\label{fig:bias_chexpert_race}

\end{figure*}

Additionally, Figures \ref{fig:histogram_race} and \ref{fig:bias_chexpert_race} contain results using different race subgroups for CheXpert. Our original evaluations considered each of the original subgroups listed within the dataset (Table \ref{tab:chex_dataset}) when computing the fairness metrics. Given the small counts for the American Indian or Alaska Native and Native Hawaiian or Other Pacific Islander subgroups, leading to large error bars, we also computed these metrics when including these subgroups within the Other subgroup. 

\begin{figure}[ht]
    \centering

    \includegraphics[width=0.95\linewidth]{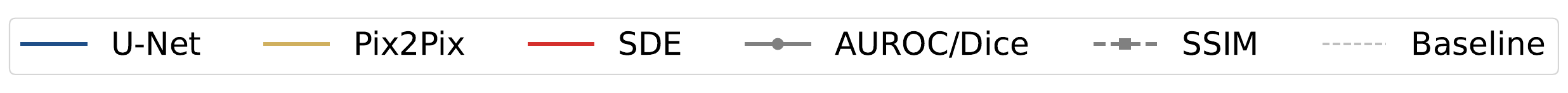}
    \vspace{1em}

    \begin{tabular}{c@{\hspace{0.5cm}}c}
        \subfigure[]{\includegraphics[width=0.49\linewidth]{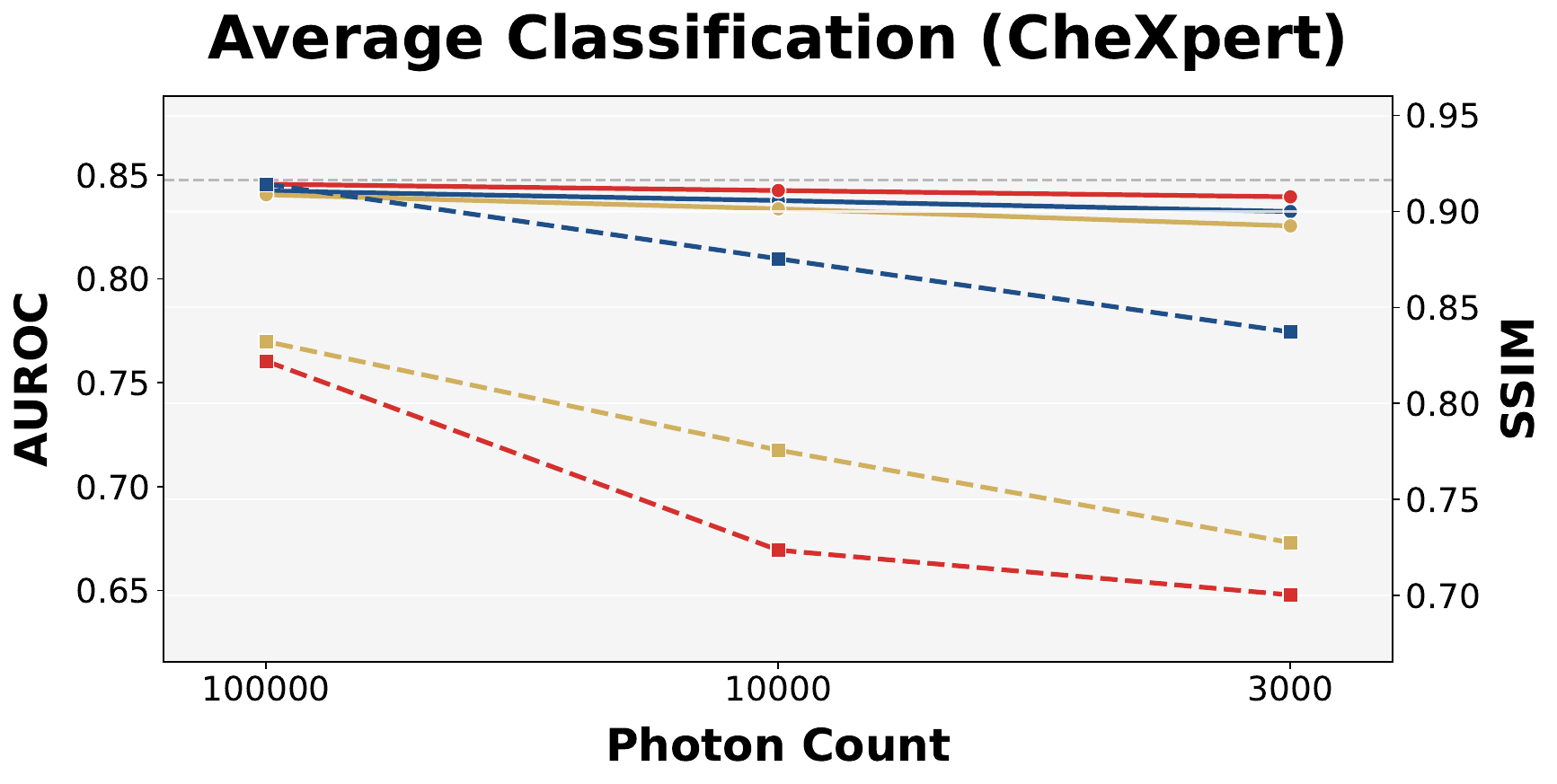}} &
        \subfigure[]{\includegraphics[width=0.49\linewidth]{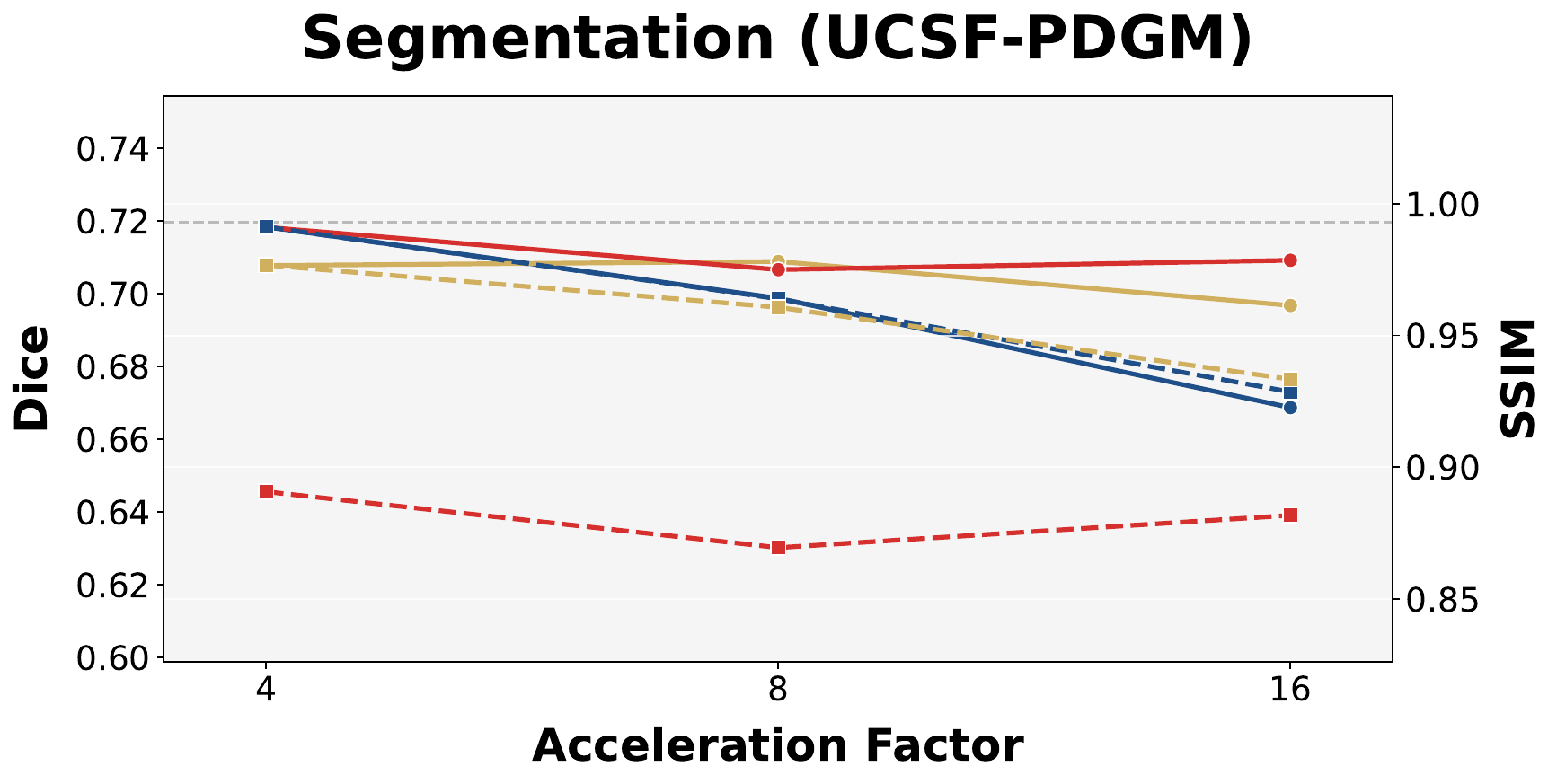}}
    \end{tabular}
    \vspace{-10pt}
    \caption{Downstream performance and SSIM at varying noise levels. Axes for SSIM and task performance are scaled to comparable percentage ranges. Baseline indicates performance on original images.}
    \label{fig:performance_ssim1}
\end{figure}

\begin{figure}[ht]
    \centering

    \includegraphics[width=0.95\linewidth]{plots/rebuttal/performance/ucsf-evaluation_performance_legend_ssim.pdf}
    \vspace{1em}

    \begin{tabular}{c@{\hspace{0.5cm}}c}
        \subfigure[]{\includegraphics[width=0.49\linewidth]{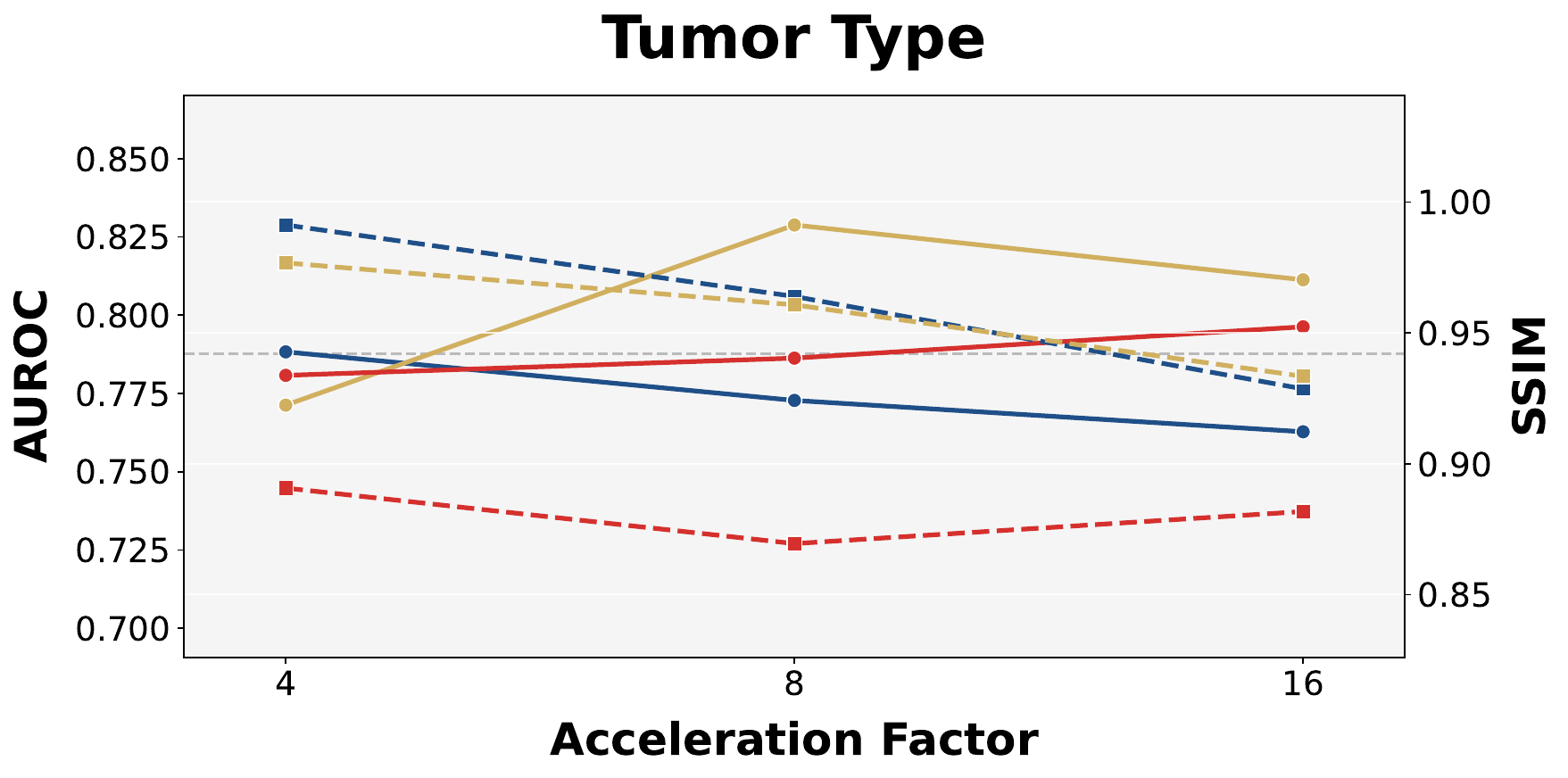}} &
        \subfigure[]{\includegraphics[width=0.49\linewidth]{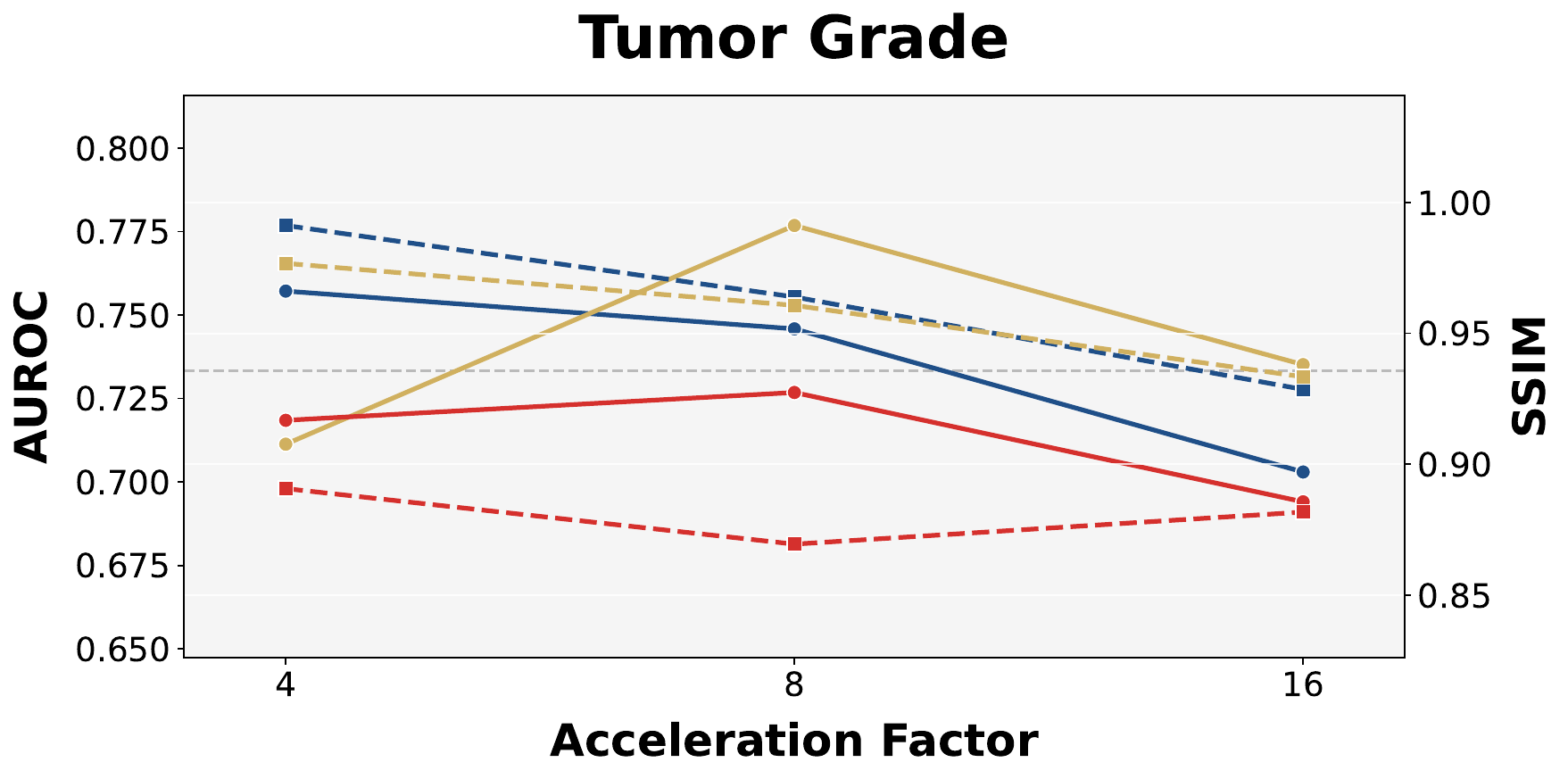}}
    \end{tabular}
    \vspace{-10pt}
    \caption{Downstream performance and SSIM at varying noise levels on classification tasks in UCSF-PDGM. Axes for SSIM and task performance are scaled to comparable percentage ranges. Baseline indicates performance on original images.}
    \label{fig:performance_ssim2}
\end{figure}

\begin{figure}[ht]
    \centering


    \begin{tabular}{c}
        \subfigure[]{\includegraphics[width=0.5\linewidth]{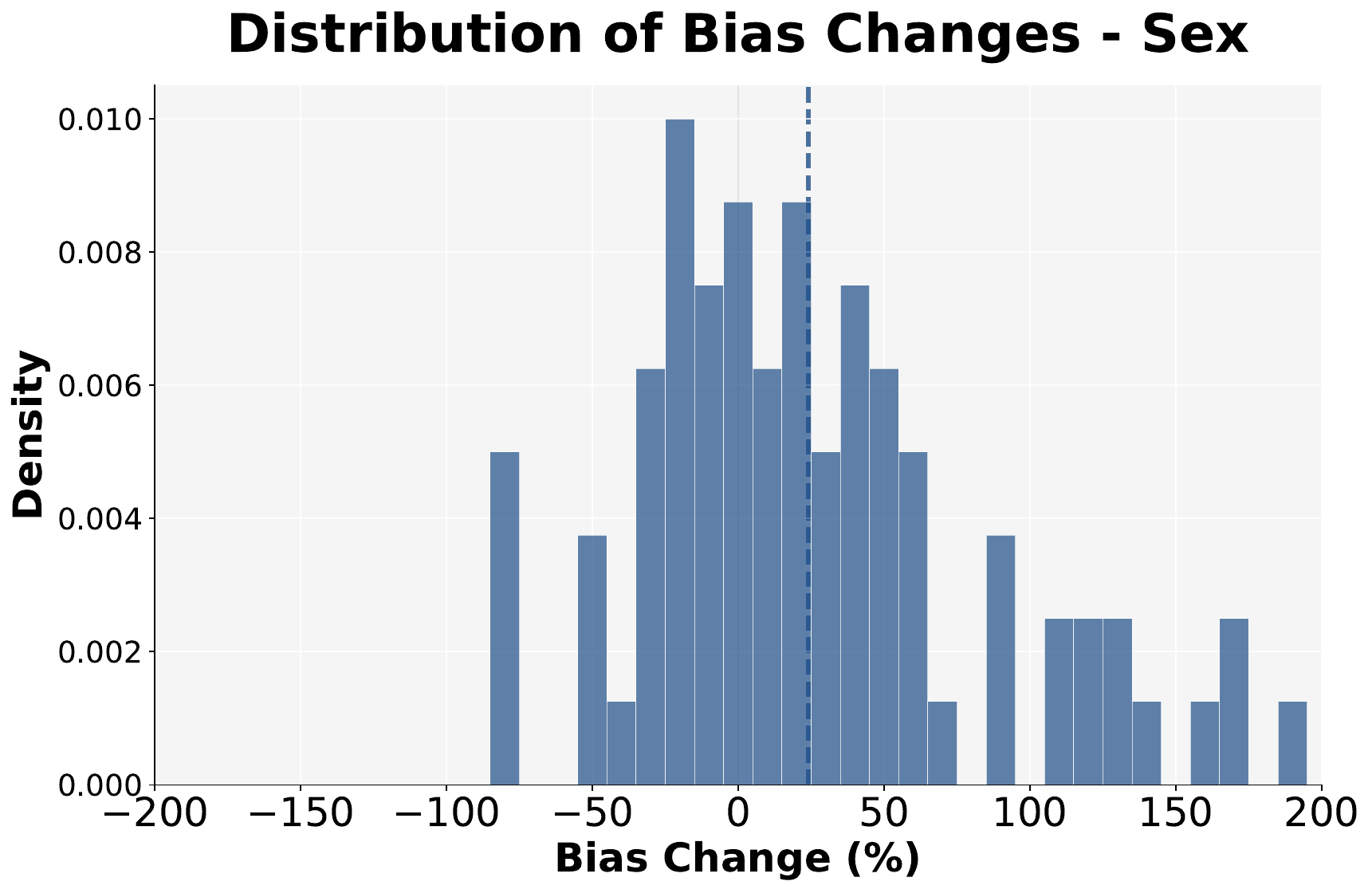}} \\[0.5cm]
        \subfigure[]{\includegraphics[width=0.5\linewidth]{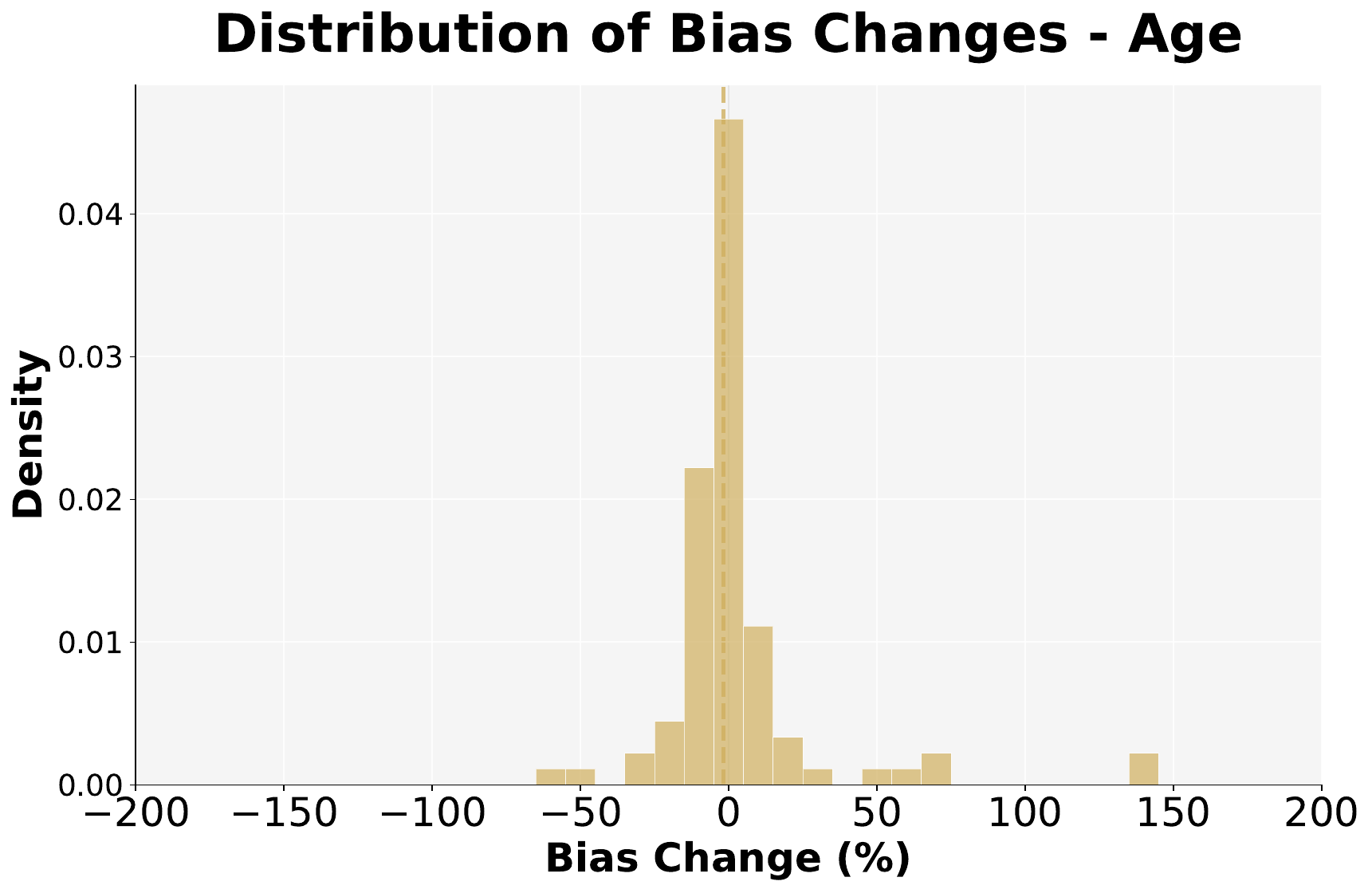}} \\[0.5cm]
        \subfigure[]{\includegraphics[width=0.5\linewidth]{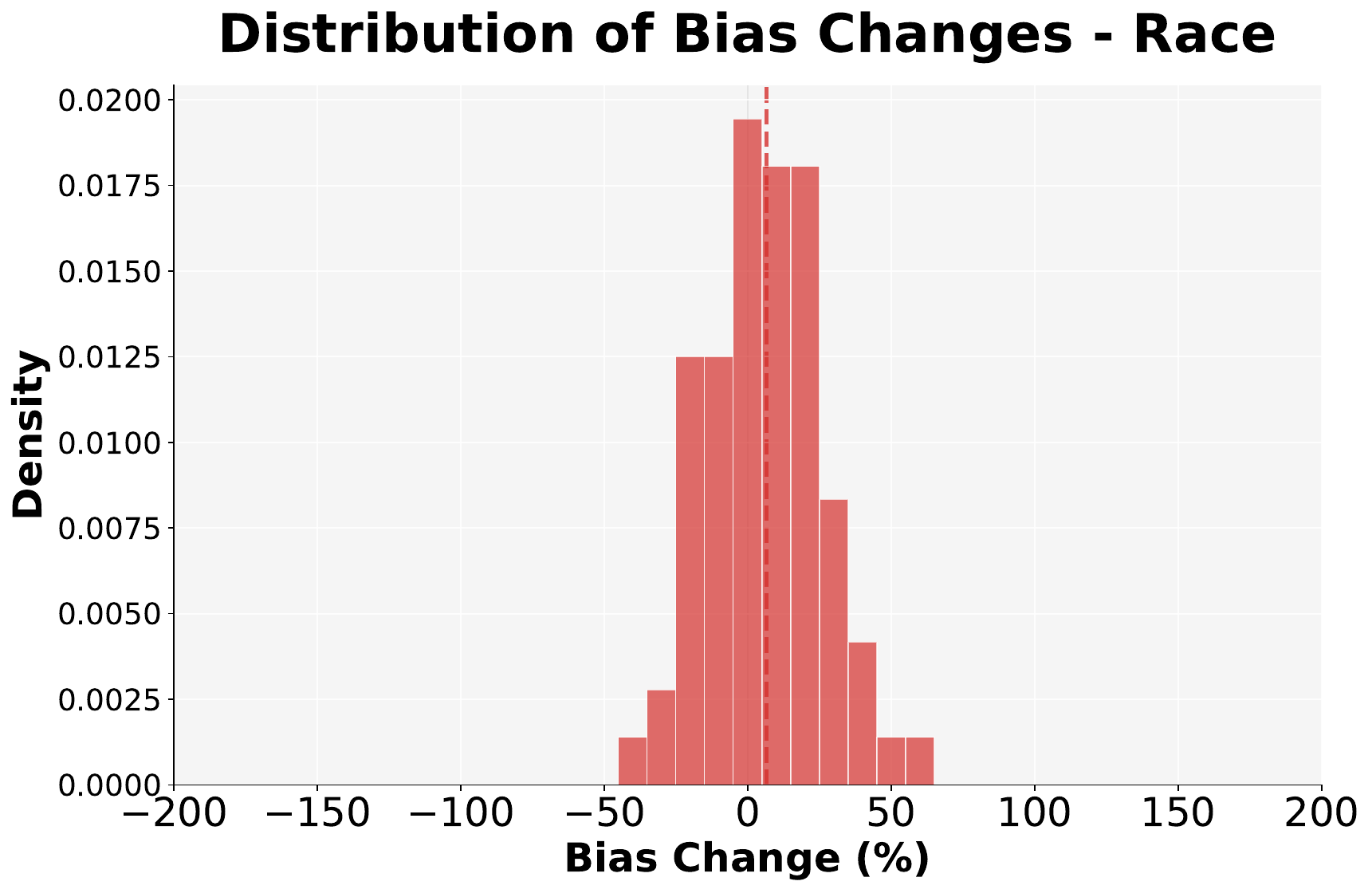}}
    \end{tabular}

    \vspace{-10pt}
    \caption{Distribution of bias changes separated by sensitive attribute.}
    \label{fig:alt_bias_histogram}
\end{figure}

\end{document}